\documentclass{article}

\PassOptionsToPackage{numbers, compress}{natbib}

\usepackage[preprint]{neurips_2023}

\usepackage{wrapfig}
\usepackage{subfigure}
\usepackage{setspace}
\usepackage[utf8]{inputenc} 
\usepackage[T1]{fontenc}    
\usepackage[pagebackref,breaklinks,colorlinks]{hyperref}
\usepackage{url}            
\usepackage{booktabs}       
\usepackage{amsfonts}       
\usepackage{nicefrac}       
\usepackage{microtype}      
\usepackage[table,xcdraw,dvipsnames]{xcolor}         
\usepackage{graphicx} 
\usepackage{tabularx}
\usepackage{amsmath}
\usepackage{xspace}
\usepackage{graphbox}
\usepackage{float}
\usepackage{multirow}
\usepackage{booktabs}
\usepackage{adjustbox}
\usepackage{caption,setspace}

\definecolor{blue-violet}{rgb}{0.54, 0.17, 0.89}

\newcommand{\ie}{\textit{i.e., }}
\newcommand{\eg}{\textit{e.g., }}
\newcommand{\statement}[1]{\noindent\textbf{#1}}
\newcommand{\showwidth}{0.245\linewidth}

\newcommand{\cprompt}[1]{{\scriptsize\textit{#1}}}

\newcommand{\tabletitle}[1]{{\scriptsize\textbf{#1}}}

\newcommand{\wcprompt}[1]{\parbox[c]{0.16\linewidth}{\centering\setstretch{0.5}\cprompt{#1}}}

\title{Guide3D: Create 3D Avatars from Text and Image Guidance}

\author{Yukang Cao\textsuperscript{1}
\quad
Yan-Pei Cao\textsuperscript{2}
\quad
Kai Han\textsuperscript{1$\dagger$}
\quad
Ying Shan\textsuperscript{2}
\quad
Kwan-Yee K. Wong\textsuperscript{1} 
\vspace{0.3em} 
\\
{\normalsize \textsuperscript{1}The University of Hong Kong} \qquad  
{\normalsize \textsuperscript{2}ARC Lab, Tencent PCG}
}

\begin{document}

\maketitle

\begin{abstract}

Recently, text-to-image generation has exhibited remarkable advancements, with the ability to produce visually impressive results. In contrast, text-to-3D generation has not yet reached a comparable level of quality. Existing methods primarily rely on text-guided score distillation sampling (SDS), and they encounter difficulties in transferring 2D attributes of the generated images to 3D content. In this work, we aim to develop an effective 3D generative model capable of synthesizing high-resolution textured meshes by leveraging both textual and image information. To this end, we introduce {\em Guide3D}, a zero-shot text-and-image-guided generative model for 3D avatar generation based on diffusion models. Our model involves (1) generating sparse-view images of a text-consistent character using diffusion models, and (2) jointly optimizing multi-resolution differentiable marching tetrahedral grids with pixel-aligned image features.
We further propose a similarity-aware feature fusion strategy for efficiently integrating features from different views. Moreover, we introduce two novel training objectives as an alternative to calculating SDS, significantly enhancing the optimization process. We thoroughly evaluate the performance and components of our framework, which outperforms the current state-of-the-art in producing topologically and structurally correct geometry and high-resolution textures. Guide3D enables the direct transfer of 2D-generated images to the 3D space. Our code will be made publicly available.

\end{abstract} 
\renewcommand{\thefootnote}{}
\footnote{$\dagger$ Corresponding author}
\vspace{-2.5em}
\section{Introduction}

\begin{figure}[t]
    \centering 
    \setlength{\tabcolsep}{0.245pt}
    \begin{tabular}{cccccc} 
        \includegraphics[align=c,width=\showwidth]{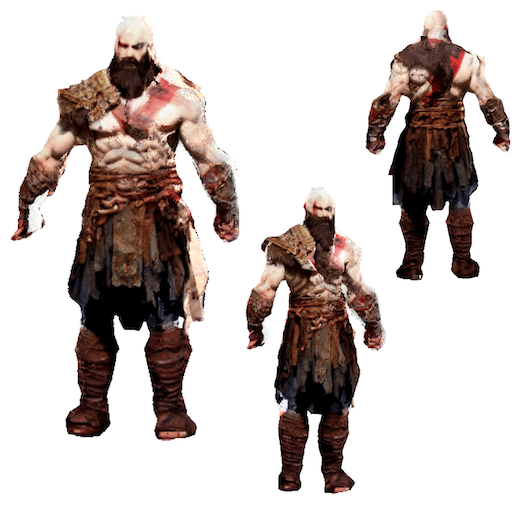}&
        \includegraphics[align=c,width=\showwidth]{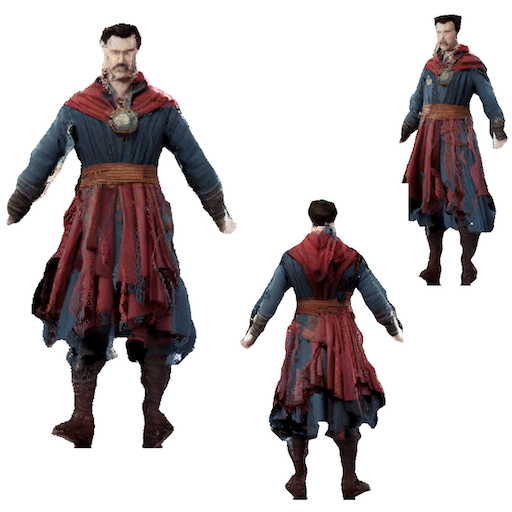}&
        \includegraphics[align=c,width=\showwidth]{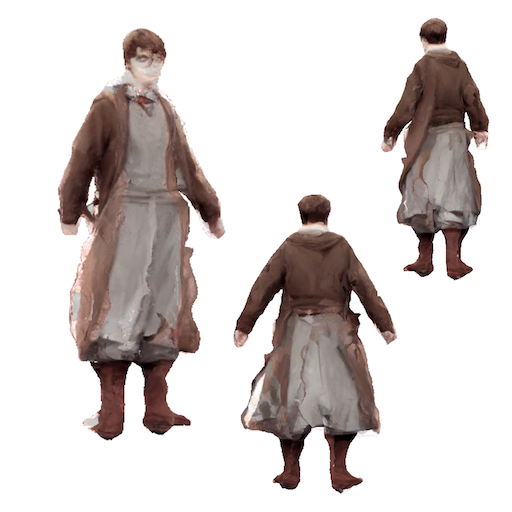}&
        \includegraphics[align=c,width=\showwidth]{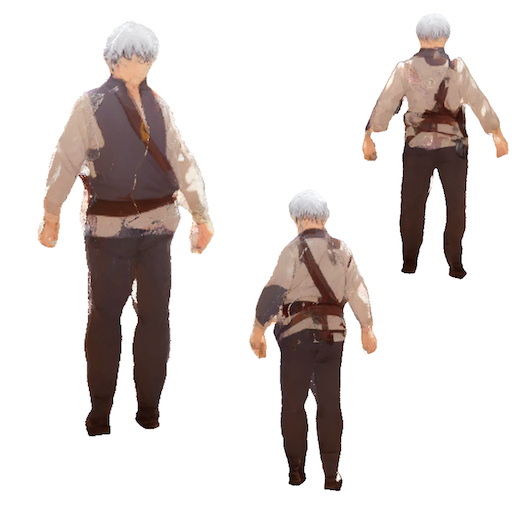}\\
        {\wcprompt{Kratos in God of War}} & 
        {\wcprompt{Doctor Strange, marvel chracter}} &
        {\wcprompt{Harry Potter}} &
        {\wcprompt{Gintoki}} \\
        
        \includegraphics[align=c,width=\showwidth]{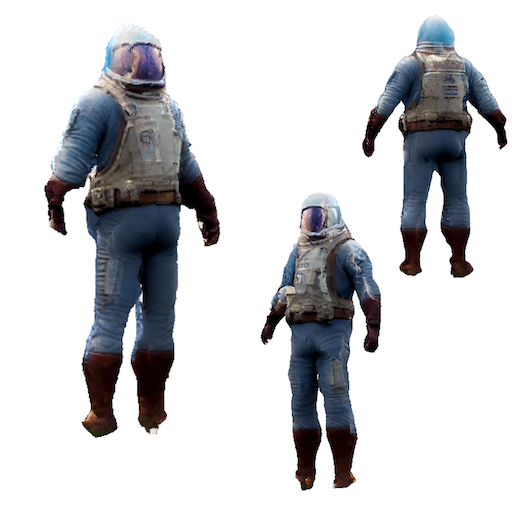}&
        \includegraphics[align=c,width=\showwidth]{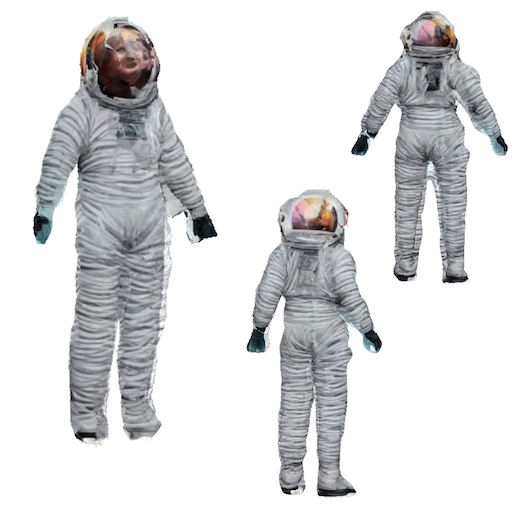}&
        \includegraphics[align=c,width=\showwidth]{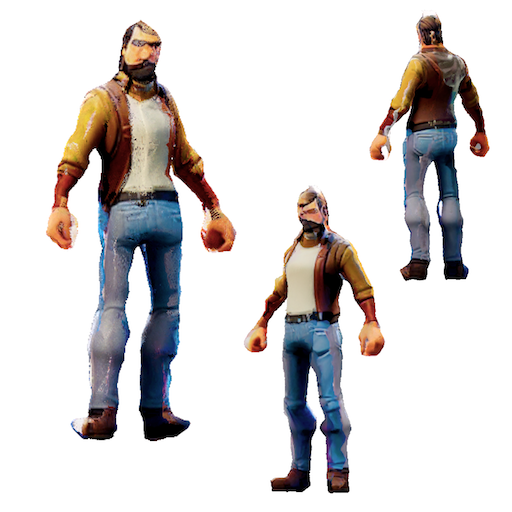}&
        \includegraphics[align=c,width=\showwidth]{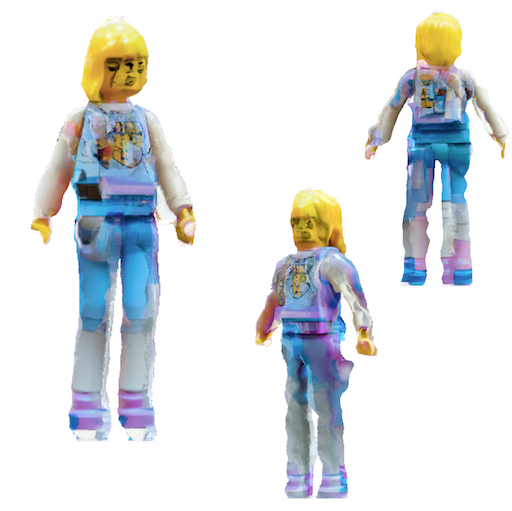} \\
        {\wcprompt{Spacesuit}} & 
        {\wcprompt{Woman Astronaut}} &
        {\wcprompt{Steve Jobs as the Fortnite character}} &
        {\wcprompt{A lego friend figurine}} \\
        
        \includegraphics[align=c,width=\showwidth]{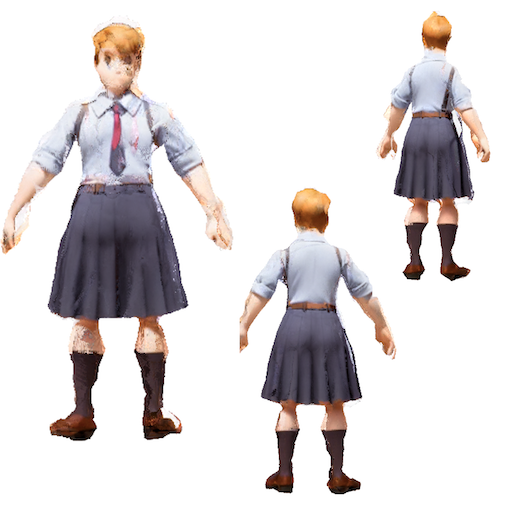}&
        \includegraphics[align=c,width=\showwidth]{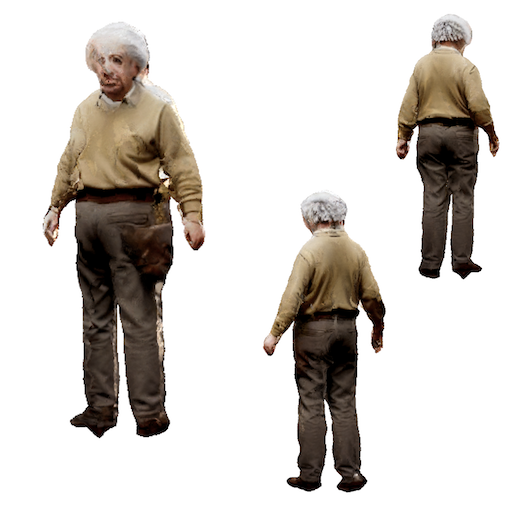}&
        \includegraphics[align=c,width=\showwidth]{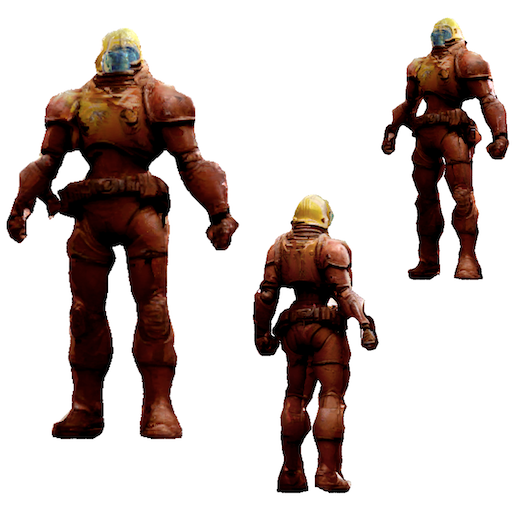}&
        \includegraphics[align=c,width=\showwidth]{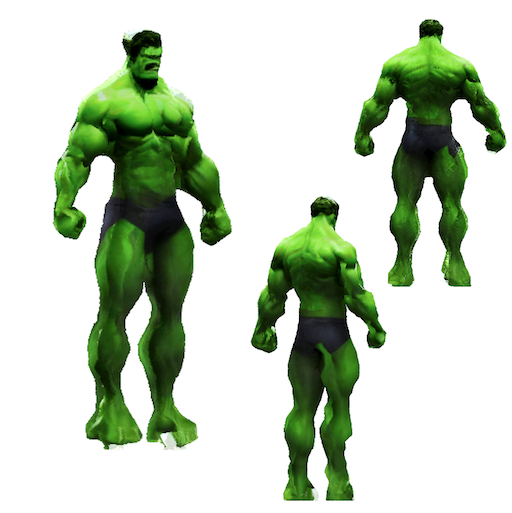} \\
        {\wcprompt{Girl wearing school uniform}} & 
        {\wcprompt{Albert Einstein}} &
        {\wcprompt{Fallout armor figurine}} &
        {\wcprompt{Hulk}} \\
    \end{tabular}
    \caption{Generation results of Guide3D. Our method can generate high-quality geometry and texture for any type of avatars, effectively transferring 2D image attributes to the 3D content.
    }
    \vspace{-1.5em}
    \label{fig:main_results}
\end{figure}

3D avatar modeling, driven by growing public curiosity and demand across diverse fields, including virtual and augmented reality~\cite{cipresso2018past}, game character modeling, and film production, has emerged as an area of significant interest. 
Traditionally, creating 3D contents with intricate details is known to be a time-consuming task that requires expertise in computer-aided design (CAD) and fine arts. Recent developments in deep learning have enabled researchers to incorporate prior information from various sources, such as images~\cite{saito2019pifu, saito2020pifuhd, he2020geopifu, he2021arch++, Huang_2020_CVPR, xiu2023econ, xiu2022icon, cao2022jiff, zheng2021pamir, cao2023sesdf}, voxel data~\cite{gilbert2018volumetric, varol2018bodynet, Zheng2019DeepHuman, Trumble_2018_ECCV, jackson20183d, Zheng2019DeepHuman}, and point clouds~\cite{zheng2022pointavatar, zhang2022differentiable, xu2022point}, to generate high-fidelity 3D human models. Despite their promising results, these methods face challenges in reconstructing unseen attributes outside their training data, and are limited by the scarcity of large and diverse 3D human datasets.

In recent years, the remarkable success of text-to-image generative models~\cite{radford2021learning, stable-diffusion, rombach2022ldm, saharia2022imagen} has spurred the advancement of methods that utilize  languages~\cite{poole2022dreamfusion, lin2023magic3d, cao2023dreamavatar} to guide the creation of high-fidelity 3D content. DreamFusion~\cite{poole2022dreamfusion} was the first to propose score distillation sampling (SDS) based on a pre-trained image diffusion model for generating 3D content in a self-optimization manner. Follow-up works~\cite{metzer2022latent-nerf, lin2023magic3d, seo20233dfuse} have attempted to improve the 3D consistency and performance of the networks; however, they often face challenges in generating detailed textures for 3D avatars and providing essential constraints to enforce consistent and correct 3D human structures. Recently, DreamAvatar~\cite{cao2023dreamavatar} utilizes the SMPL prior to robustly generate 3D avatars under controllable shapes and poses. Despite their impressive results, recent text-to-3D generative frameworks exhibit four major limitations:  (1) \textit{View inconsistency and geometric distortion.} These stem from the absence of 3D awareness in 2D diffusion models and inconsistent camera poses, leading to multi-face "Janus" problems and incorrect topological and structural geometry.
(2) \textit{High failure rate.} They struggle to generate good results consistently.
(3) \textit{Low-quality geometry.} The extracted geometry from optimized NeRF or DMTet often exhibits artifacts and imperfections. 
(4) \textit{Limitations in image processing.} The current methods primarily rely on text, which poses challenges when it comes to effectively handling images. This issue is compounded by the extensive use of images in various practical applications.

To address these challenges, we propose Guide3D, which minimizes the view inconsistency to text-generated images and leverages differentiable marching tetrahedra (\textsc{DMTet}) to effectively transfer high-quality generated 2D image attributes to the 3D content with topologically and structurally correct geometry.
Specifically, given a textual description as input, Guide3D begins by generating multi-view images of the text-described character. We then jointly optimize the multi-resolution \textsc{DMTet} grids, ultimately yielding a highly detailed textured mesh. Our approach distinguishes itself from existing techniques by integrating the comprehensive priors found in 3D human reconstruction methods and the imagination capability of diffusion models. 

The main contributions of Guide3D can be attributed to three principal components. \emph{First}, we devise a framework that utilizes pixel-aligned image features obtained from the generated multi-view images as priors and jointly optimizes multi-resolution \textsc{DMTet} grids. This multi-resolution optimization significantly improves both the stability and quality of geometry by seamlessly and mutually transferring information between low- and high-resolution grids.
\emph{Second}, we propose a similarity-aware feature fusion strategy that efficiently incorporates multi-view image features. \emph{Third}, we introduce two novel training objectives aimed at refining the optimization process: (1) To mitigate noticeable inconsistency among different views of text-generated images, rather than calculating SDS, we minimize the image-level difference between generated multi-view images and the renderings denoised by the pre-trained image diffusion model. (2) We further propose a training guidance designed to minimize the difference between the HED (holistically-nested edge detection~\cite{xie2015holistically}) boundaries detected from the generated multi-view images and denoised renderings. The specially designed supervision reduces the impact of inconsistency among the generated images and further improves the geometry quality.

We perform comprehensive evaluations of Guide3D and compare it with prior methods. Guide3D significantly outperforms existing methods, demonstrating the ability to create topologically and structurally correct geometry and transfer detailed 2D image attributes to 3D geometry.
\vspace{-1em}
\section{Related work}
\vspace{-1em}
\statement{Text-guided 3D generation.}
The advancement of large vision-language models (\eg CLIP~\cite{radford2021learning}) and diffusion models~\cite{saharia2022imagen, stable-diffusion} has sparked a surge of interest in text-guided 3D generation. Notable efforts in this domain include CLIP-mesh~\cite{khalid2022clipmesh}, CLIP-Forge~\cite{sanghi2022clip-forge}, DreamField~\cite{jain2022dreamfeild}, Dream3D~\cite{xu2022dream3d}, Text2mesh~\cite{michel2022text2mesh}, and AvatarCLIP~\cite{hong2022avatarclip}. They employ CLIP as their learning prior to self-optimize the underlying 3D representations (\eg NeRF or vertex-based mesh). These methods represent a significant leap forward in the realm of 3D content generation, enabling the creation of 3D models without the need for 3D training dataset. Nonetheless, owing to the limited ability of CLIP to fully grasp human languages, they fall short of producing convincing 3D meshes.

Recently, diffusion-based text-guided 2D image generation models have emerged, exhibiting improved language comprehension. DreamFusion~\cite{poole2022dreamfusion} is the first to introduce a Score Distillation Sampling (SDS) strategy based on a pre-trained 2D diffusion model~\cite{stable-diffusion,saharia2022imagen,balaji2022ediffi} to accomplish self-optimization of a NeRF-based 3D content. Follow-up studies~\cite{wang2022sjc,metzer2022latent-nerf} have improved its performance and efficiency, and further extended SDS to various 3D representations, including mesh~\cite{lin2023magic3d,chen2023fantasia3d}, texture~\cite{richardson2023texture,chen2023text2tex}, and point cloud~\cite{nichol2022point-e}.
Notably, Magic3D~\cite{lin2023magic3d} adopts a coarse-to-fine pipeline to generate high-resolution 3D textured mesh based on \textsc{DMTet}~\cite{shen2021dmtet}. HeadSculpt~\cite{han2023headsculpt} proposes a novel IESD loss for identity-aware 3D editing. Fantasia3D~\cite{chen2023fantasia3d} leverages \textsc{DMTet} and introduces a disentangled pipeline for independent learning of geometry and texture. 3DFuse~\cite{seo20233dfuse} incorporates depth information into diffusion models to enhance the stability of the generation process. DreamAvatar~\cite{cao2023dreamavatar} employs SMPL~\cite{bogo2016smpl} and proposes a dual space design with a density-residual setup to achieve robust control over the generated 3D avatars. ProlificDreamer~\cite{wang2023prolificdreamer} proposes Variational Score Distillation for better results and diversity. Despite their efforts, text-guided 3D avatar generation methods still confront obstacles in producing fine-grained meshes.

\statement{Text-guided image to 3D reconstruction.}
Text-guided image to 3D reconstruction methods endeavor to utilize both textual and image information for 3D reconstruction by taking both image and text as input. RealFusion~\cite{melas2023realfusion} proposes to learn a textual inversion embedding from the input image, followed by supervising novel view rendering with SDS. Make-it-3D~\cite{tang2023make} presents a two-stage pipeline utilizing point cloud as an intermediate to elevate the quality of 3D reconstruction from a single image. Zero-1-to-3~\cite{liu2023zero} trains a text-guided 2D image diffusion model to integrate camera rotation and translation information, thereby enhancing the quality of single-view 3D reconstruction with SDS. DreamBooth3D~\cite{raj2023dreambooth3d} extends the capability of DreamBooth~\cite{ruiz2022dreambooth} by introducing a 3-stage optimization strategy that improves the 3D consistency and personalization potential of the model, making it a promising solution for various applications. While these methods have shown promising results, there is still a need for an efficient approach to integrate information from both text and image for 3D avatar generation.

\statement{Image-guided 3D human reconstruction.}
The advance in neural networks has enabled implicit function based methods to demonstrate promising outcomes in image-based 3D human reconstruction. These methods predict either an occupancy field~\cite{mescheder2019occupancy, chibane20ifnet} or signed distance field~\cite{park2019deepsdf, chabra2020deep, jiang2020sdfdiff} as their intermediates. PIFu~\cite{saito2019pifu} is the pioneering method that proposes pixel-aligned image features for 3D clothed human reconstruction. Subsequent studies have exploited 3D features obtained from 3D parametric human/face models to address the depth-ambiguity issue~\cite{zheng2021pamir}, animate 3D clothed human model~\cite{Huang_2020_CVPR, he2021arch++}, or enhance the level of details~\cite{saito2020pifuhd,cao2022jiff, alldieck2022phorhum}. Multi-stage methods, such as PIFuHD~\cite{saito2020pifuhd} and FITE~\cite{lin2022fite}, aim to improve the clothing topology in the reconstruction. Geo-PIFu~\cite{he2020geopifu} introduces a novel approach to compute 3D features directly from images. Other methods~\cite{xiu2022icon,xiu2023econ,cao2023sesdf} leverage signed distance fields based on SMPL as a guide or investigate explicit representation by using bi-linear normal-depth integration~\cite{cao2022bilateral}. Despite their spectacular results, these methods rely heavily on expensive and hard-to-obtain 3D datasets for training. Furthermore, they face challenges in reconstructing attributes that are not present in the 3D datasets.
\section{Methodologies}
\label{sec:methodologies}
\begin{figure}[t]
  \centering
  \includegraphics[width=0.98\linewidth]{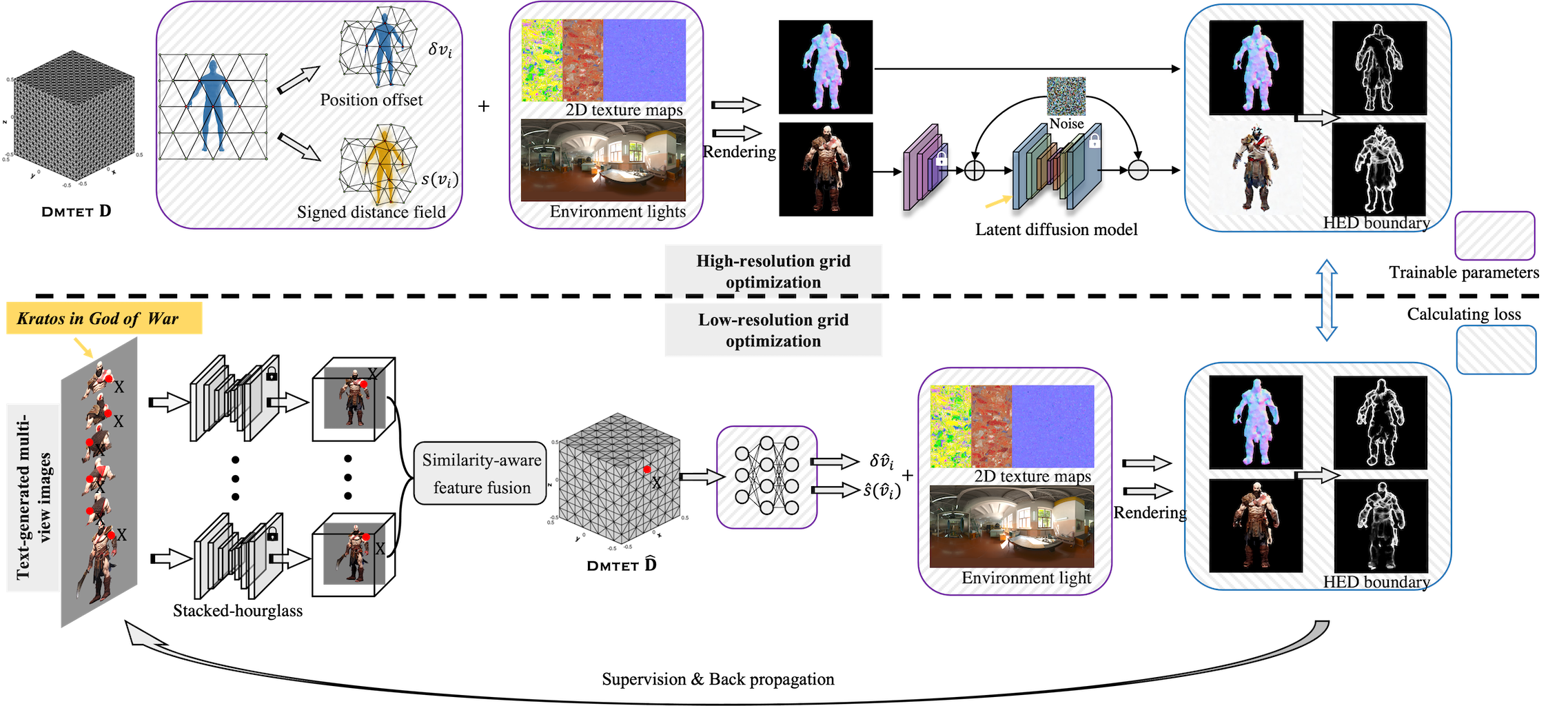}
  \vspace{-2pt}
  \caption{Overall architecture of Guide3D. By taking a text prompt as input, our network sequentially generates multi-view 2D images and transfers the image attributes to the 3D avatar. At the core of our networks are two multi-resolution \textsc{DMTet} grids that are jointly optimized by the use of pixel-aligned image features and image diffusion models.}
  \label{fig:pipeline}
  \vspace{-10pt}
\end{figure}
In this paper, we introduce Guide3D, a novel text-to-3D approach that aims to generate 3D human avatars with topologically and structurally correct geometry, in the presentation of \textsc{DMTet}, based on character-oriented multi-view images. 
As presented in Fig.~\ref{fig:pipeline}, the process of transferring the image attributes to 3D space involves the joint optimization of multi-resolution \textsc{DMTet} grids which incorporates the guidance from pre-trained pixel-aligned image features and pre-trained diffusion models. In the subsequent sections, we will describe the details of Guide3D, focusing on the following key aspects: (1) preliminary knowledge in Sec.~\ref{sec:preliminaries}; (2) controllable multi-view image generation in Sec.~\ref{sec:images_generation}; (3) joint optimization of multi-resolution \textsc{DMTet} grids in Sec.~\ref{sec:joint_optimization}; and (4) formulation of the training objectives in Sec.~\ref{sec:training_objectives}. 

\subsection{Preliminaries}
\label{sec:preliminaries}
\statement{Diffusion models.}
To achieve high-quality image generation, the text-to-image diffusion model is trained to learn the noise $\hat{\epsilon}_{\phi}(\mathbf{z}_t;t, y)$ at level $t$ conditioned on the text embedding $y$~\cite{stable-diffusion, rombach2022ldm} with classifier-free guidance (CFG)~\cite{ho2022classifier}
\begin{equation}
\hat{\epsilon}_{\phi}(\mathbf{z}_t;y, t) = (1 + \omega) \epsilon_{\phi}(\mathbf{z}_t;t, y) - \omega \epsilon_{\phi}(\mathbf{z}_t;t),
\end{equation}
Where $\mathbf{z}_t$ represents the encoded features of the image, $\epsilon{\phi}(\mathbf{z}_t;t)$ and $\epsilon{\phi}(\mathbf{z}t;y, t)$ refer to the learned noise conditioned on the image features and text, respectively.

While DreamFusion relies solely on textual prompts with a guidance scale $\omega = 100$, \emph{our model prioritizes image attributes} and applies pixel-level losses on denoised renderings (see Sec.~\ref{sec:training_objectives}). This results in high-quality 3D models that faithfully reflect input image attributes and are devoid of unwanted inconsistency.

\statement{Pixel-aligned implicit function.}
To extract faithful priors from the generated images, we propose utilizing pixel-aligned image features introduced in PIFu~\cite{saito2019pifu}.
Specifically, PIFu learns a pixel-aligned implicit function $F_{occ}$ that takes the the indexed pixel-aligned image feature $f(x)$ for 3D point $x$ as input, predicting per-point occupancy value:
\begin{equation}\label{implicitFunction}
    F_{\text{occ}}(x, f(x)) \mapsto [0, 1] \in \mathbb{R},
\end{equation}

\statement{\textsc{DMTet}.}
\textsc{DMTet}~\cite{shen2021dmtet} is a 3D geometry representation that comprises two key components: (1) A deformable tetrahedral grid $(V_T, T)$ defined by the vertices $V_T$ and tetrahedra $T$. Each vertex $v_i \in V_T$ is assigned with a signed distance value $s(v_i)$ and a position offset value $\delta v_i$:
\begin{equation}
    (s(v_i), \delta v_i) = \Psi(v_i; \phi),
\end{equation}
where $\Psi$ denotes a neural network (\eg MLP) with training parameters $\phi$ to predict the values. A mesh can be subsequently extracted using the differentiable marching tetrahedra algorithm.
(2) A differentiable renderer~\cite{munkberg2022extracting, laine2020modular} is employed to render the extracted mesh using the physically-based (PBR) material model from Disney~\cite{mcauley2012practical}, which integrates a diffuse term with an isotropic, specular GGX lobe~\cite{walter2007microfacet}. Guide3D prioritizes \textsc{DMTet} over the neural radiance field (NeRF) due to its enhanced computational efficiency and high-resolution geometry and textures.

\vspace{-0.5em}
\subsection{Controllable multi-view images generation.}
\label{sec:images_generation}
\vspace{-0.5em}
\begin{figure}[t]
    \centering 
    \includegraphics[width=\linewidth]{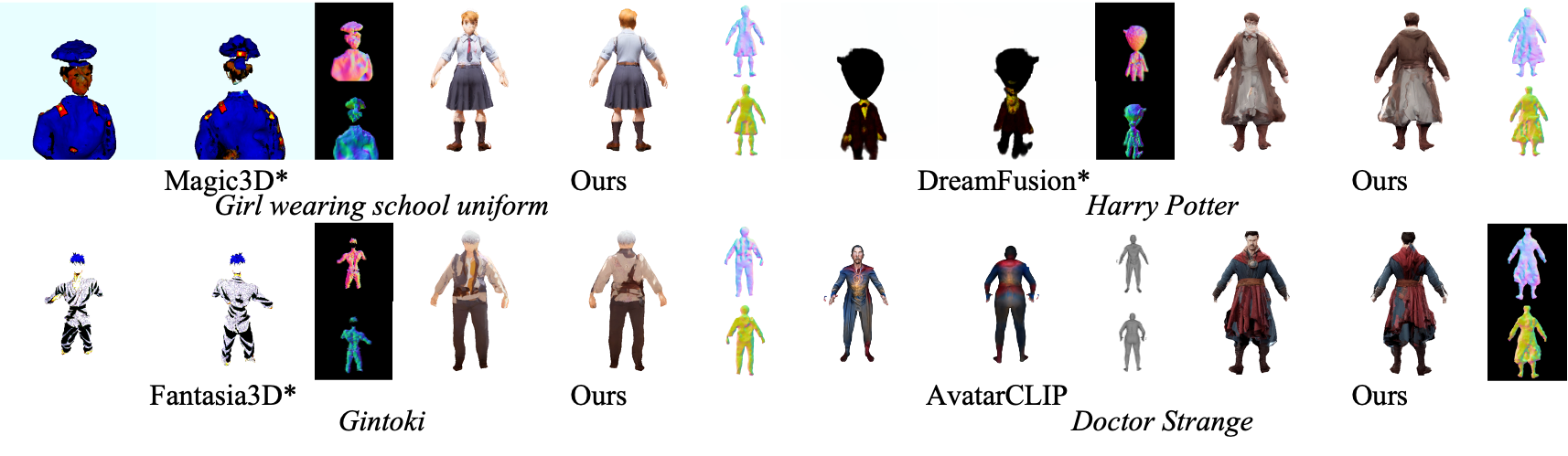}
    \vspace{-2em}
    \caption{Comparisons with Magic3D~\cite{lin2023magic3d}, DreamFusion~\cite{poole2022dreamfusion}, Fantasia3D~\cite{chen2023fantasia3d}, and AvatarCLIP~\cite{hong2022avatarclip}. *Non-official implementation.}
    \label{fig:comparison_sota}
\end{figure}

\begin{figure}[t]
    \centering 
    \setlength{\tabcolsep}{0.2pt}
    \includegraphics[align=c,width=0.98\linewidth]{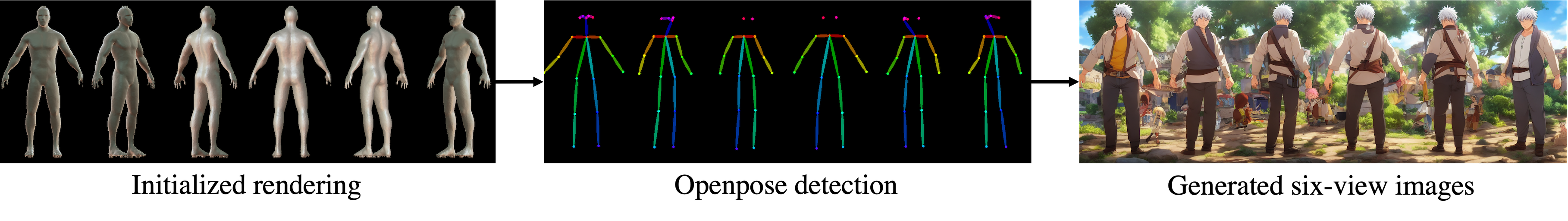}
    \vspace{-5pt}
    \caption{Examples of text-to-2D image generations. Our network first renders multi-view images from the initialized model, providing the OpenPose guidance to the ControlNet image diffusion model to generate coherent characters. However, the images obtained from image generation may exhibit inconsistency in terms of body pose, texture, and orientation.}
    \label{fig:example}
\end{figure}

Existing methods follow the Score Distillation Sampling (SDS) strategy proposed in DreamFusion~\cite{poole2022dreamfusion}, which involves sampling a camera viewpoint, defining its orientation (i.e., top, bottom, left, right, front, back) at each iteration, and afterward optimizing the 3D model via diffusion models with renderings as intermediate. SDS, therefore, overlooks the consistency among different viewpoints, resulting in the multi-face "Janus" problems and difficulty in generating accurate geometry and texture (see Fig.~\ref{fig:comparison_sota} and \ref{fig:comparison_dreamavatar}). To handle such problems, we propose to utilize the generated multi-view 2D images as intermediates, transferring the image attributes to 3D human avatars with topologically and structurally correct geometry.

Specifically, we employ ControlNet~\cite{zhang2023controlnet} and textual inversion (CharTurnerV2~\cite{CharTurnerV2}) to generate $K$-view images of a text-coherent character for subsequent processing, where we set $K=6$ in this paper (see supplementary for analysis on the number of views).
Unfortunately, these multi-view images may exhibit inconsistency in terms of body pose and appearance. Furthermore, their orientations may not align with the provided guidance, \emph{presenting a non-trivial challenge in reconstructing a 3D avatar model directly from such images (see Fig.~\ref{fig:example})}.

\vspace{-0.5em}
\subsection{Joint optimization of multi-resolution grids}
\label{sec:joint_optimization}
\begin{figure}[t]
    \centering 
    \begin{minipage}{0.48\linewidth}
    \includegraphics[width=\columnwidth]{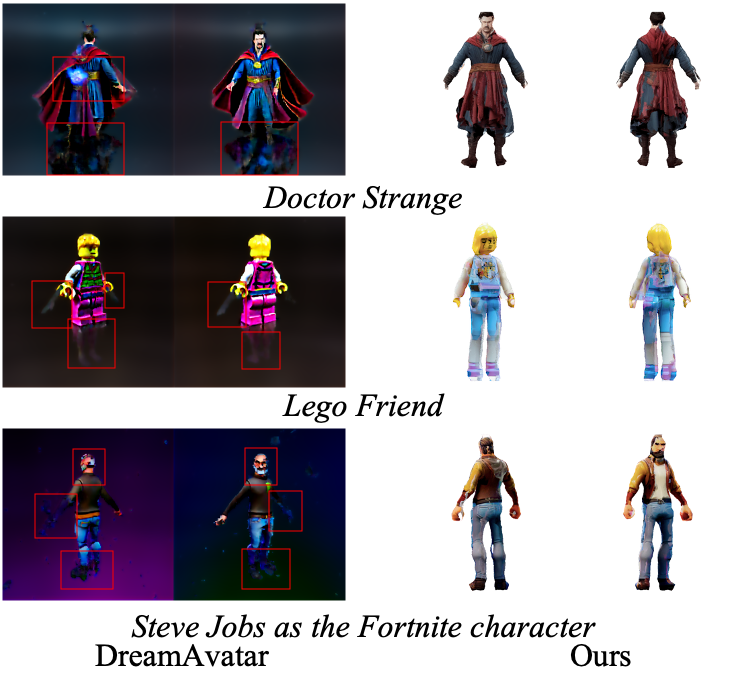}
    \vspace{-2em}
    \caption{Comparisons with DreamAvatar~\cite{cao2023dreamavatar}.}
    \label{fig:comparison_dreamavatar}
    \end{minipage}
    \vspace{-0.5em}
    \hfill
    \begin{minipage}{0.48\linewidth}
    \includegraphics[width=\linewidth]{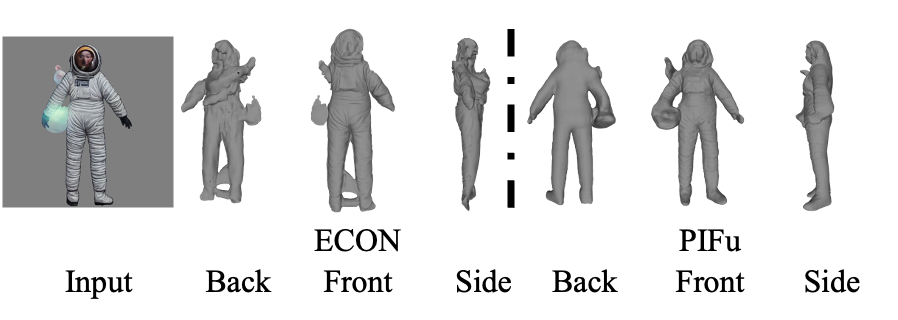}
    \caption{Single-view human reconstruction.}
    \label{fig:reconstruction_single}
    \includegraphics[width=\linewidth]{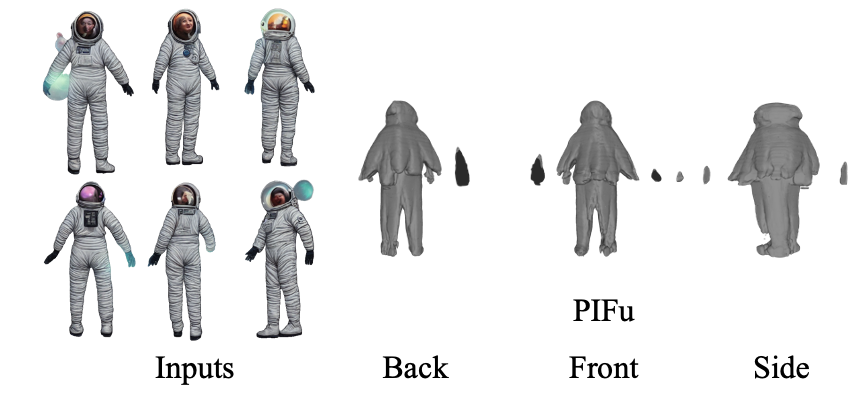}
    \vspace{-2em}
    \caption{Multi-view human reconstruction.}
    \label{fig:reconstruction_multi}
    \end{minipage}
    \vspace{-0.5em}
\end{figure}

\vspace{-0.5em}
As discussed above, transferring the generated multi-view images to 3D models poses a notable challenge. One straightforward approach is 3D human reconstruction, as demonstrated in previous works such as PIFu~\cite{saito2019pifu} and ECON~\cite{xiu2023econ}. However, Fig.~\ref{fig:reconstruction_single} and \ref{fig:reconstruction_multi} showcase the limitations of both single-view and multi-view implicit human reconstruction methods in capturing the pose, orientation, and intricate details of human subjects, particularly when dealing with out-of-distribution subjects such as characters from movies and cartoons. Furthermore, the absence of a comprehensive 3D dataset containing diverse and novel characters has impeded the ability to address these limitations through more effective network training. To overcome this challenge, we propose a self-optimization network with the joint optimization of multi-resolution grids. 

\paragraph{Grids initialization.}
For subsequent illustration, we denote our multi-resolution grids as $\mathbf{\hat{D}}$ and $\mathbf{D}$, with resolutions of $64^3$ and $256^3$, respectively. 
We first employ an optimization procedure, inspired by Fantasia3D~\cite{chen2023fantasia3d}, to initialize our multi-resolution grids to a default 3D human model in ``A-pose'': during each iteration, we randomly select 10,000 points, $\hat{v}_i$ and $v_i$, respectively from each grid and infer their signed distance values $\mathbf{s}(\hat{v}_i)$ and $s(v_i)$. We then minimize the difference between these values and the pseudo-ground-truth value $\mathtt{SDF}(\cdot)$ based on the default 3D human shape. This initialization procedure runs for 10,000 iterations to ensure convergence and completion:
\begin{equation}
    \mathcal{L}_{\text{init-low}} = \sum_{v_i \in \mathbf{D}} ||s(v_i; \phi) - \mathtt{SDF}(v_i)||_2^2, \quad \mathcal{L}_{\text{init-high}} = \sum_{\hat{v}_i \in \mathbf{\hat{D}}} ||\mathbf{s}(\hat{v}_i) - \mathtt{SDF}(\hat{v}_i)||_2^2,
\end{equation}
where $s(v_i; \phi)$ is predicted via an MLP, and $\gamma(\cdot)$ denotes hash grid position encoding. The above trainable parameters $\phi$ and the learnable variable $\mathbf{s}(\hat{v}_i)$ serve as starting points for the subsequent optimization.
\paragraph{Pixel-aligned image feature.}
Despite the limitations of 3D human reconstruction pipelines when dealing with novel characters from movies or cartoons, we have observed that their pixel-aligned image features can serve as a reliable prior (see Sec.~\ref{sec:ablation}). 
To leverage this, we take the publicly released PIFu model and apply its image encoder~\cite{newell2016stacked, saito2019pifu} to each of the $K$ generated images $G_k \in \mathbb{R}^{512\times512\times3}$ (where $k \in [1, K]$) to extract the image feature map $F_k \in \mathbb{R}^{128\times128\times256}$. 
However, single-view image features have inherent limitations in providing comprehensive information, e.g., depth, and often encounters difficulties with self-occlusion, particularly for the back of the human (see Fig.~\ref{fig:reconstruction_single}).
To overcome these limitations, We leverage multi-view image features to predict the signed distance and deformation for points in the low-resolution grid. For each point, we project it onto each image and extract pixel-aligned image features via bi-linear interpolation.

\paragraph{Similarity-aware feature fusion.}
Nevertheless, considering the self-occlusion in 3D space and inconsistencies among multi-view images, it is crucial to account for the distinct impacts of features extracted from different views on the final results. Unlike average pooling in PIFu~\cite{saito2019pifu}, we propose a similarity-aware feature fusion strategy that prioritizes point-level similarity in corresponding areas across views.
In each training iteration, we randomly designate a ``reference viewpoint'' denoted as $r$ from the $K$ pre-defined views for image rendering and supervision.
Technically, we first compute the cosine similarity between feature $f_k(x)$ and the ``reference feature'' $f_r(x)$ by
\begin{equation}
    w_k(\hat{v}_i) = \frac{f_k(\hat{v}_i) \cdot f_r(\hat{v}_i)}{\text{max}(||f_k(\hat{v}_i)||_2 \cdot ||f_r(\hat{v}_i)||_2,~\alpha)},
\end{equation}
where $\alpha$ is a hyperparameter (default to $1e^{-8}$ in our experiments). We then fuse all $K$ image features by weighted average, \ie
\begin{equation}
    f(\hat{v}_i) = \frac{\sum_{k=1}^{K} w_k(\hat{v}_i)f_k(\hat{v}_i)}{\sum_{k=1}^{K} w_k(\hat{v}_i)},
\end{equation}
Our final fused features $f(\hat{v}_i)$ are then input to the $\hat{\mathtt{MLP}}$ to optimize the low-resolution grid.
\vspace{-0.5em}
\paragraph{Diffusion denoising process.}
By incorporating our similarity-aware feature fusion strategy, we effectively utilize robust and consistent information from multi-view images. However, inconsistencies in the images lead to noticeable vacancies that need to be addressed to achieve high-resolution geometry and texture in the final output. Unlike SDS which employs pre-trained diffusion models for noise-level supervision and 3D optimization based on text prompts, our method focuses on minimizing image-level differences between the generated multi-view images and denoised renderings obtained from the diffusion model. This process, leveraging the imagination of diffusion models, effectively fills the vacancies at the image level and enables the generation of high-resolution textures that remain consistent across multiple views. For more details, refer to Sec.~\ref{sec:training_objectives}.
\vspace{-0.5em}
\paragraph{Joint optimization}
While both pixel-aligned image features and the diffusion denoising process offer informative priors (refer to Sec.~\ref{sec:ablation}), optimizing a single high-resolution grid (e.g., $256^3$) with both priors presents two challenges: (1) strong gradients from the diffusion model lead to noise and structural incorrectness (see Figure 6 in the submitted paper), and (2) the high point density in the high-resolution grid significantly increases training time and required GPU memory for pixel-aligned image features. Conversely, we have observed that renderings from a low-resolution grid, despite its coarse geometry, perform comparably to the high-resolution grid. To leverage this and overcome the aforementioned challenges, we propose a joint optimization approach for multi-resolution \textsc{DMTet} grids. This enables the mutual transfers of diffusion-optimized features to the low-resolution grid and robust information from pixel-aligned image features to the high-resolution grids.

Specifically, our multi-resolution grids are optimized via two separate MLPs:
\begin{equation}
\begin{array}{cc}
    \mathbf{D}: & (s(v_i; \phi), \delta v_i) = \mathtt{MLP}(\gamma(v_i)), \\
    \mathbf{\hat{D}}: & (\hat{s}(\hat{v}_i), \delta\hat{v}_i) = \hat{\mathtt{MLP}}(f(\hat{v}_i)) + (\mathbf{s}(\hat{v}_i), 0), 
\end{array}
\end{equation}
where $\hat{v}_i$ and $v_i$ are points respectively from low-resolution and high-resolution grids. $f(\hat{v}_i)$ represents the fused multi-view image features used for optimizing the low-resolution grid. Our joint optimization is then achieved by enforcing both the novel-view and known-view agreement. More details about the optimization process can be found in Sec.~\ref{sec:training_objectives}.

\vspace{-0.5em}
\paragraph{Appearance optimization.}
Please refer to the supplementary materials.

\vspace{-0.5em}
\begin{figure}[t]
    \centering 
    \setlength{\tabcolsep}{0.195pt}
    \begin{tabular}{ccccc} 
         \includegraphics[align=c,width=0.195\linewidth]{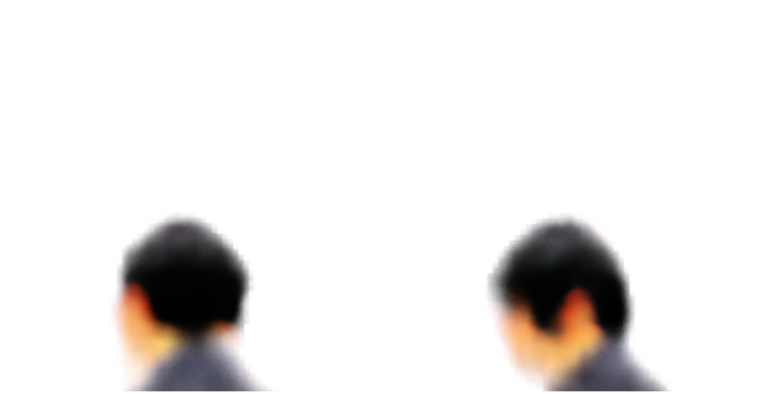}&
        \includegraphics[align=c,width=0.195\linewidth]{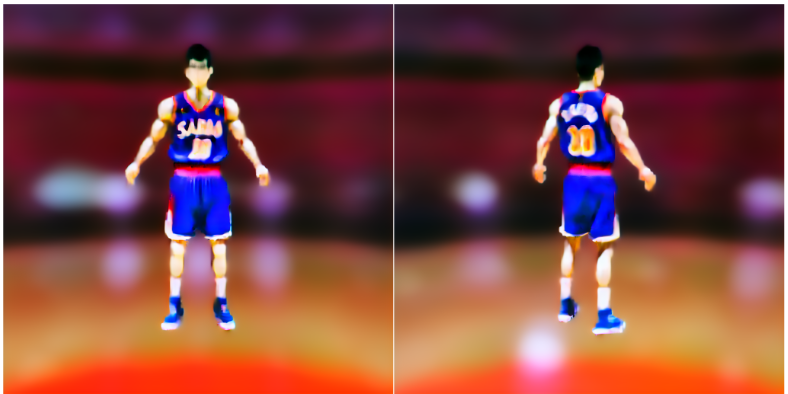} &
        \includegraphics[align=c,width=0.195\linewidth]{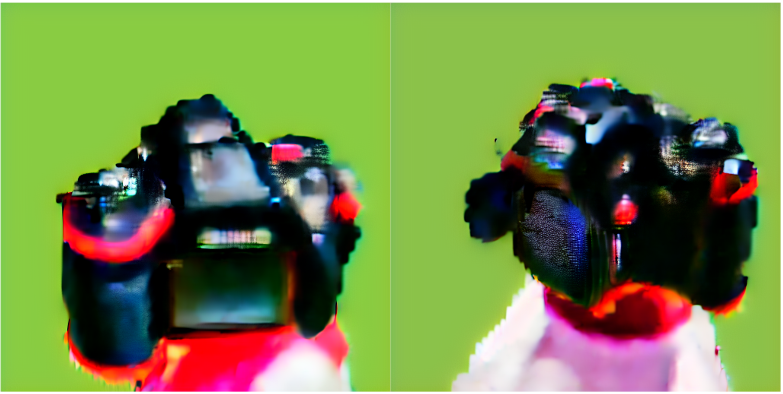}&
        \includegraphics[align=c,width=0.195\linewidth]{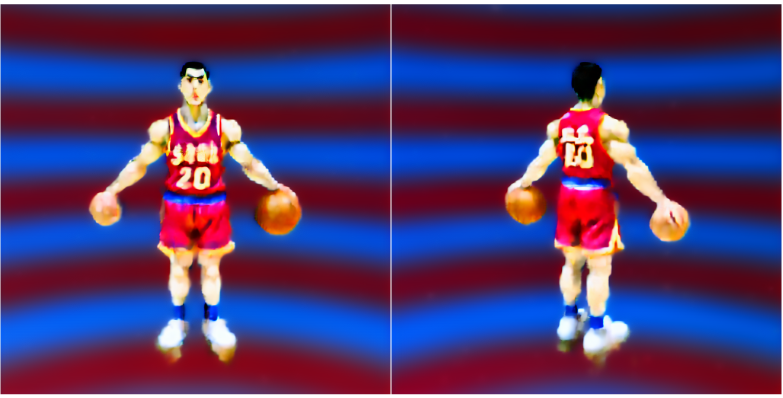}
        &
        \includegraphics[align=c,width=0.195\linewidth]{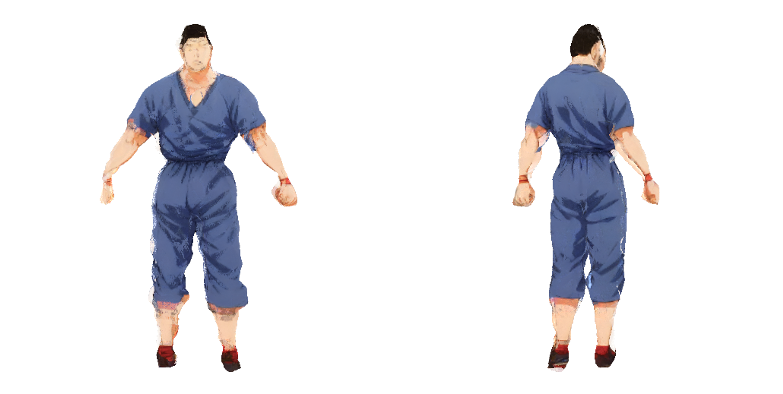}\\
        \multicolumn{5}{c}{\cprompt{a DSLR photo of Sakuragi}} 
        \\
        \includegraphics[align=c,width=0.195\linewidth]{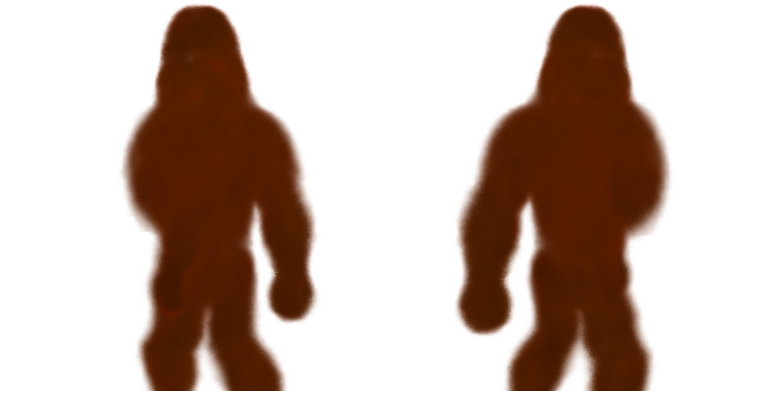}&
        \includegraphics[align=c,width=0.195\linewidth]{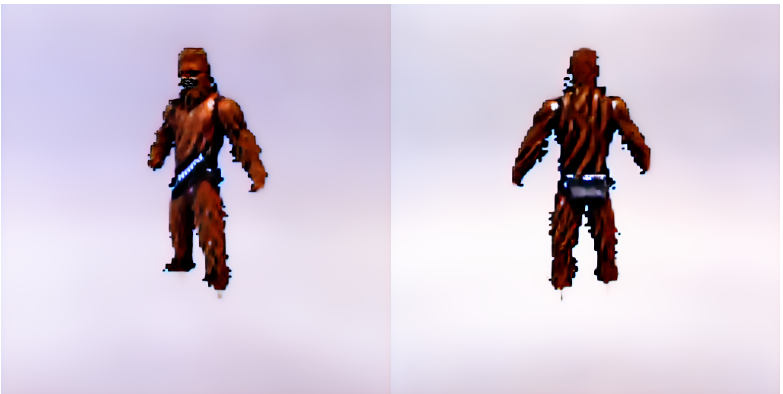} &
        \includegraphics[align=c,width=0.195\linewidth]{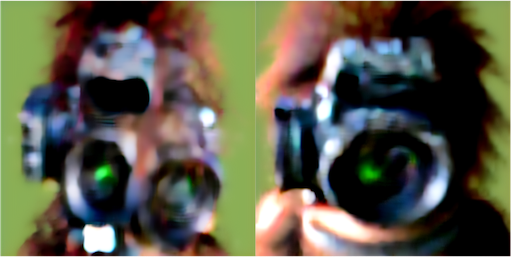}&
        \includegraphics[align=c,width=0.195\linewidth]{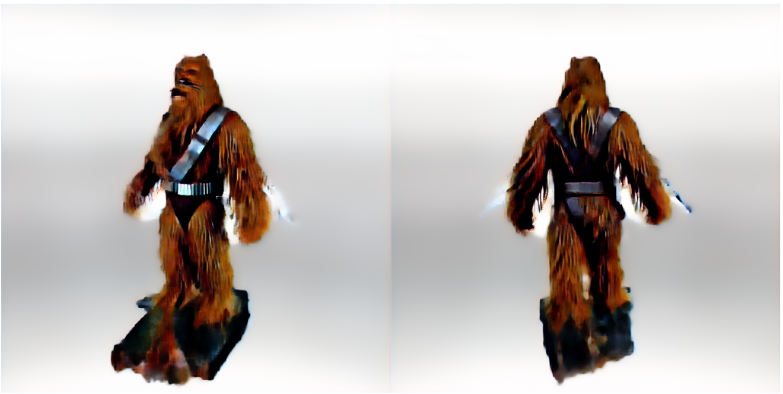}
        &
        \includegraphics[align=c,width=0.195\linewidth]{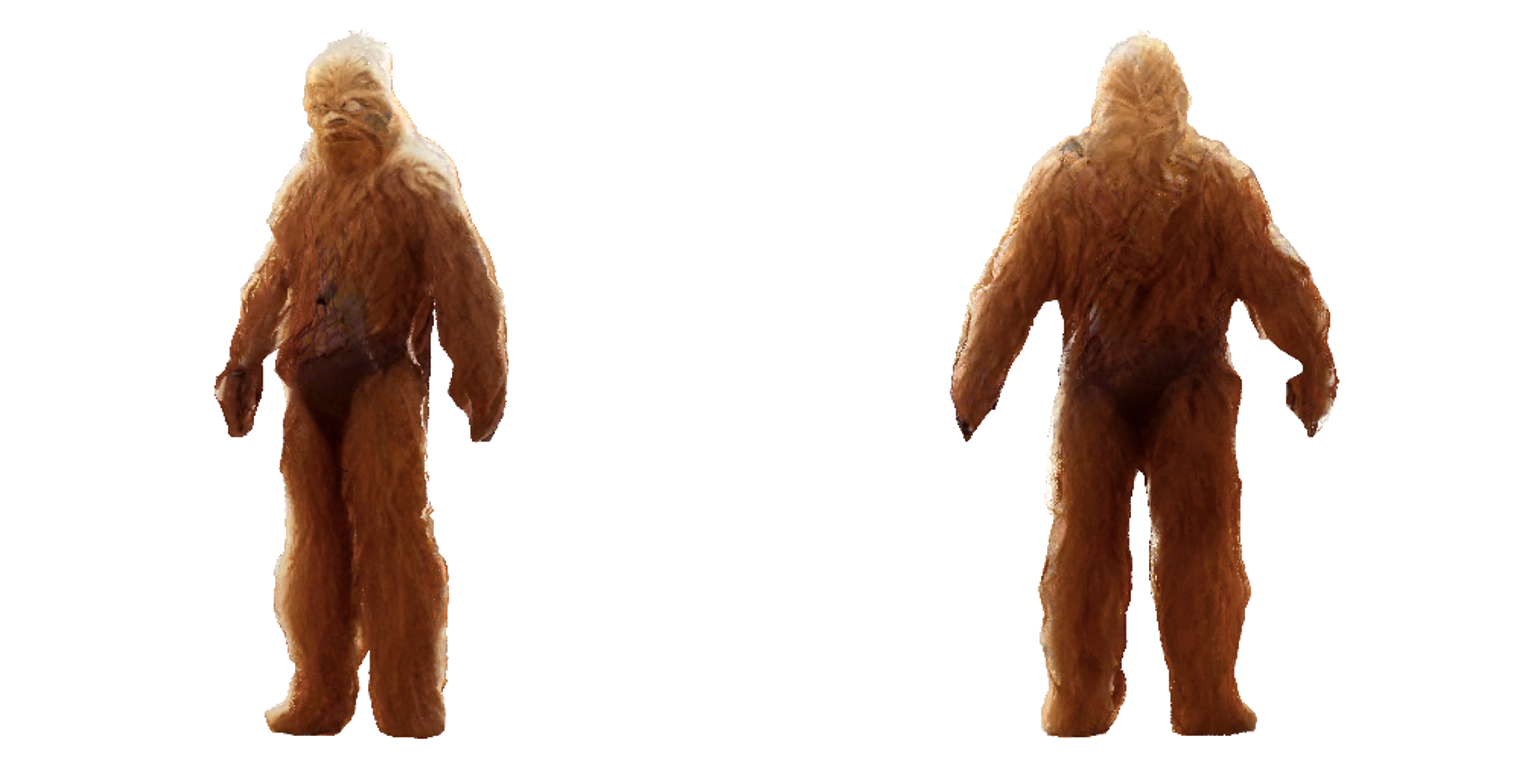}\\
        \multicolumn{5}{c}{\cprompt{a DSLR photo of Buff Chewbacca}}
        \\\includegraphics[align=c,width=0.195\linewidth]{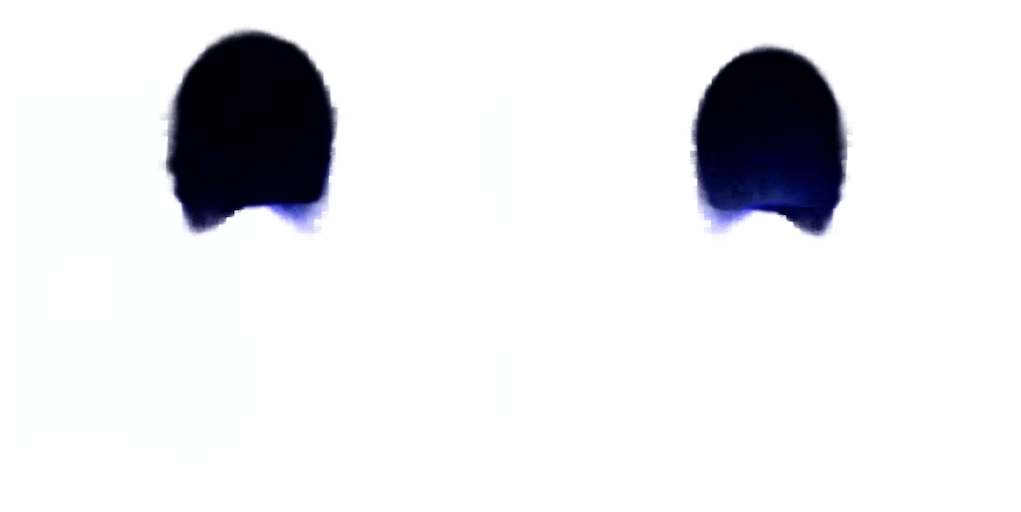}&
        \includegraphics[align=c,width=0.195\linewidth]{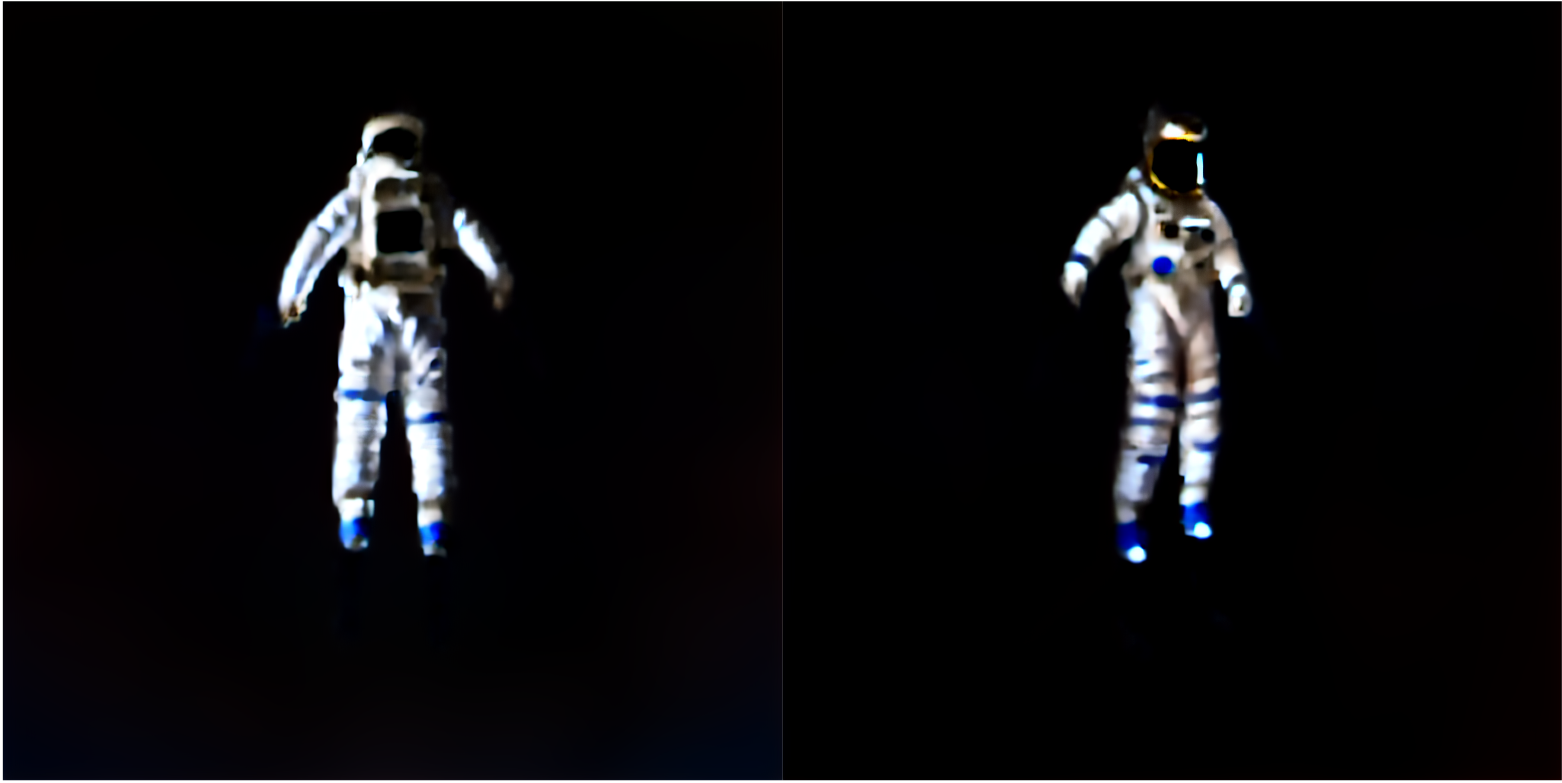} &
        \includegraphics[align=c,width=0.195\linewidth]{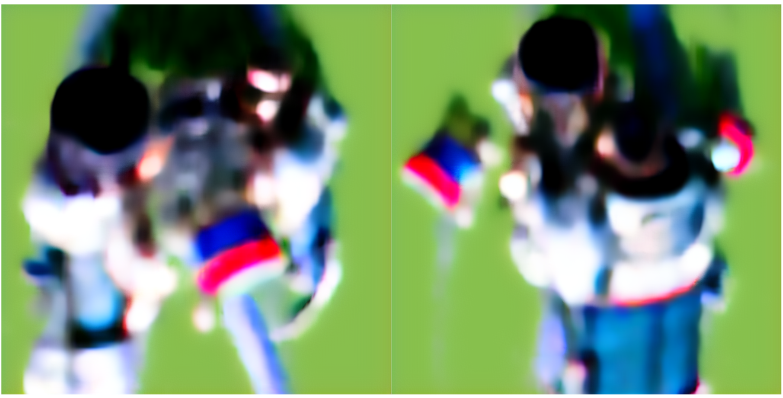}&
        \includegraphics[align=c,width=0.195\linewidth]{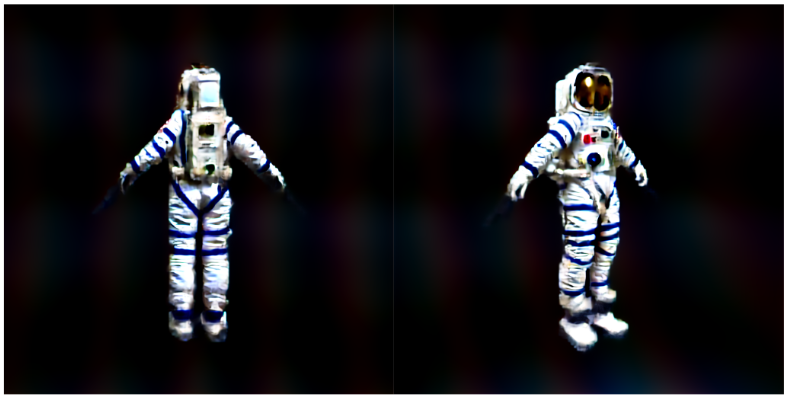}&
        \includegraphics[align=c,width=0.195\linewidth]{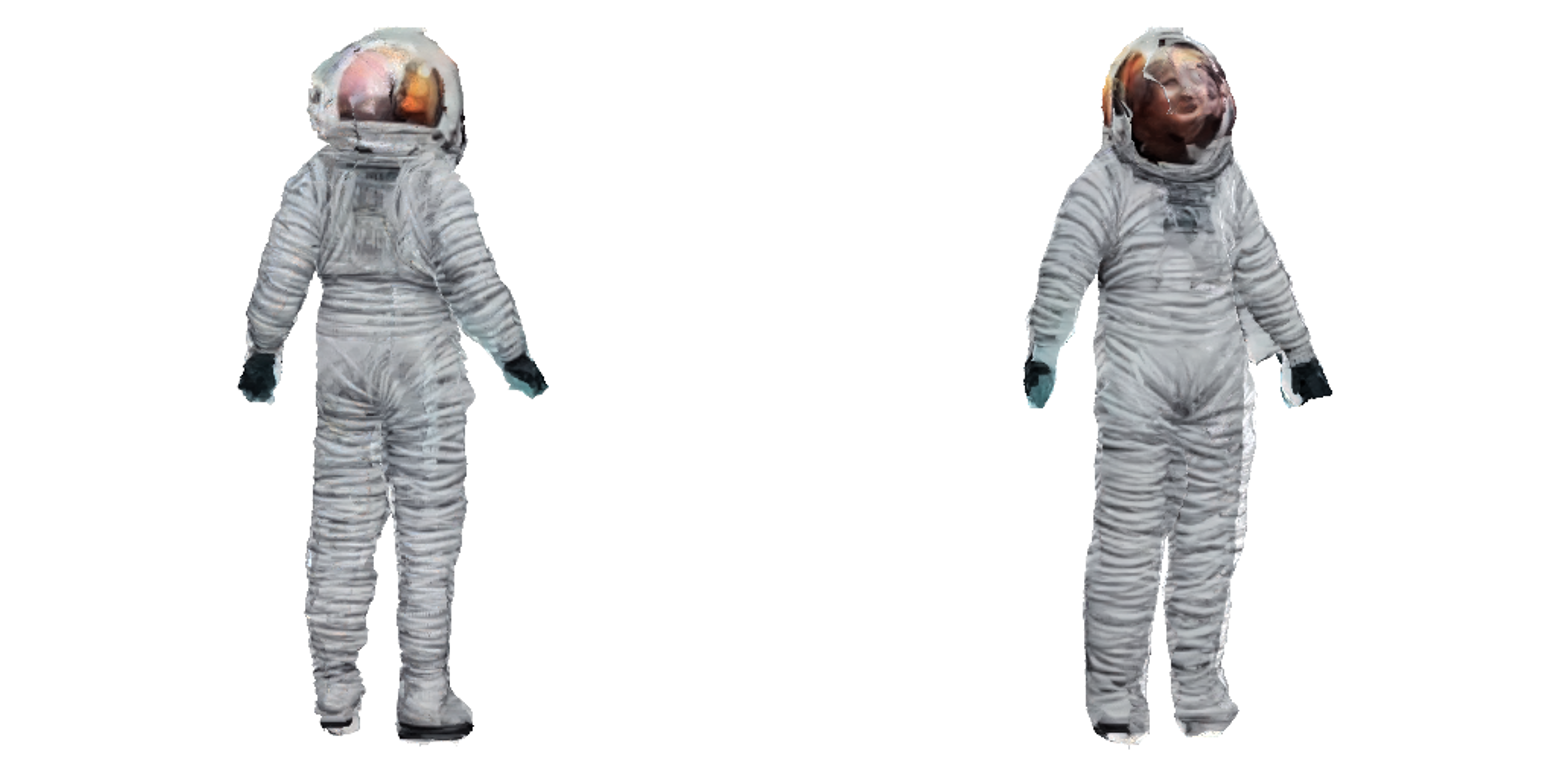}\\
        \multicolumn{5}{c}{\cprompt{a DSLR photo of a woman astronaut}} \\
        \includegraphics[align=c,width=0.195\linewidth]{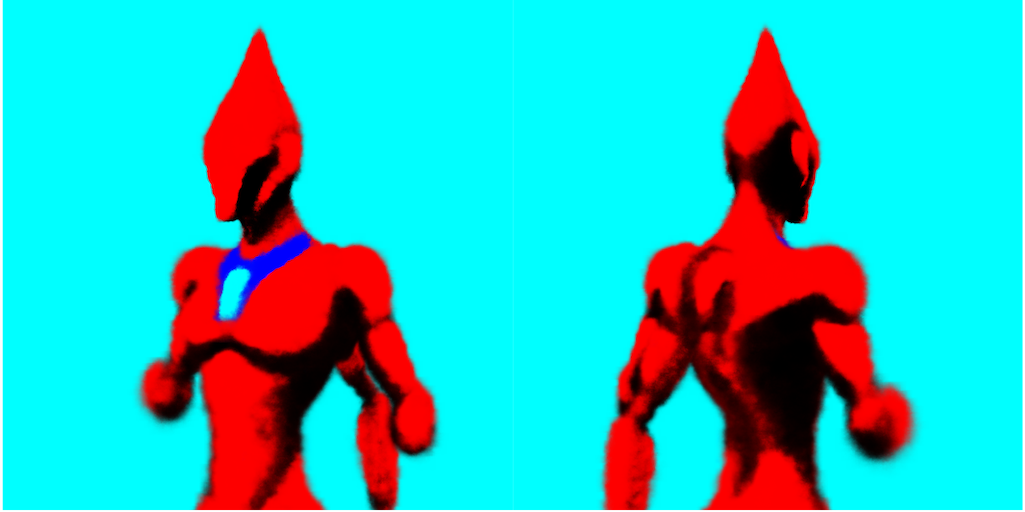}&
        \includegraphics[align=c,width=0.195\linewidth]{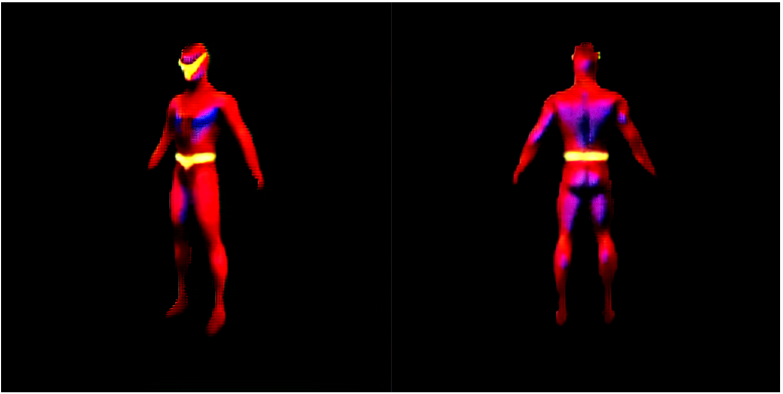} &
        \includegraphics[align=c,width=0.195\linewidth]{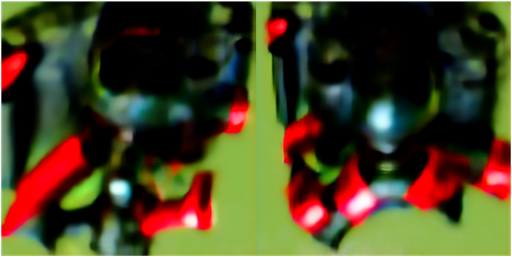}&
        \includegraphics[align=c,width=0.195\linewidth]{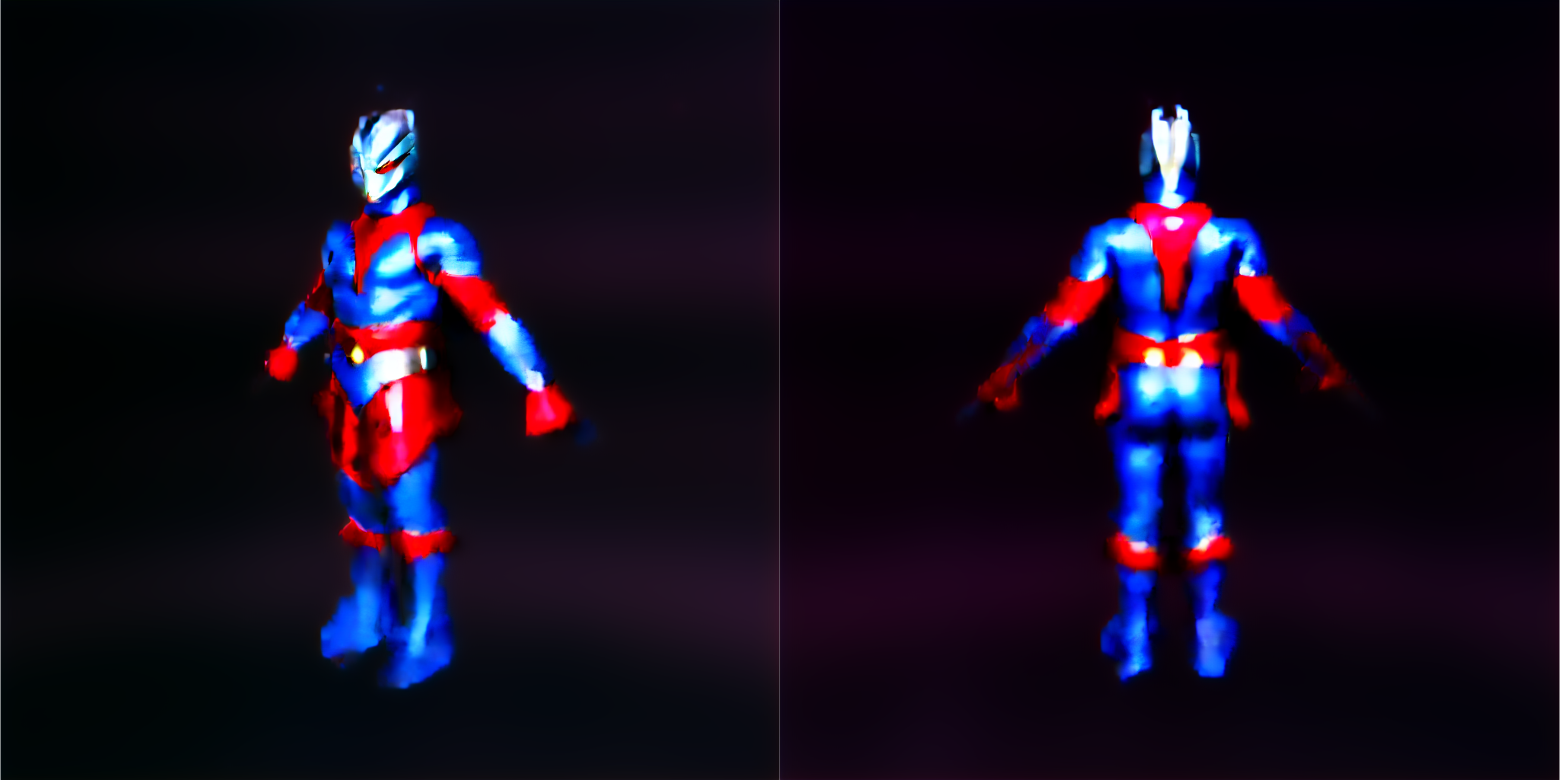}&
        \includegraphics[align=c,width=0.195\linewidth]{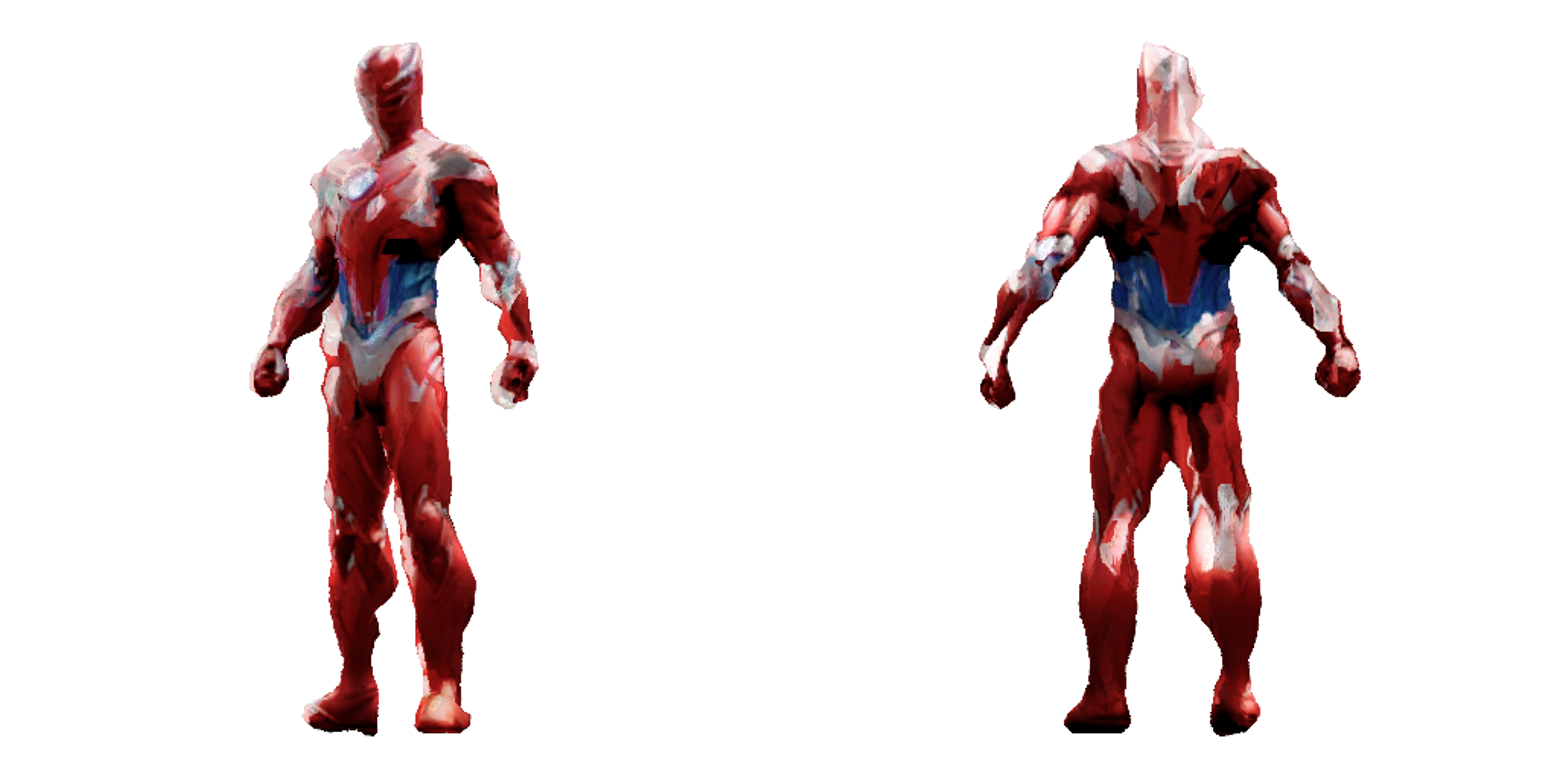}\\
        \multicolumn{5}{c}{\cprompt{a DSLR photo of an Ultraman}} \\
        \tabletitle{DreamFusion*~\cite{stable-dreamfusion}} & \tabletitle{Latent-NeRF~\cite{metzer2022latent-nerf}} & 
        \tabletitle{3DFuse~\cite{seo20233dfuse}} & \tabletitle{DreamAvatar~\cite{cao2023dreamavatar}} &      
        \tabletitle{Ours}    \\
    \end{tabular}
    \caption{Comparison with existing text-to-3D methods. Our approach distinguishes itself from other methods that struggle or fail to generate reasonable results with high-resolution texture and fine-grained geometry. Our method consistently delivers superior results and has the capability to create content based on text-to-2D image generation. *Non-official implementation.}
    \vspace{-1em}
    \label{fig:compare_sota}
\end{figure}
\subsection{Training objectives}
\label{sec:training_objectives}
To jointly optimize our multi-resolution \textsc{DMTet} grids, we randomly sample two viewpoints at each iteration, namely a known camera viewpoint $c_k$ chosen from the $K$ pre-defined viewpoints and a novel camera viewpoint $c_{\text{nv}}$. We use $G_k$ to denote the image that is previously generated at viewpoint $c_k$ (where $k \in [1, K]$), along with its alpha map $A_k$ detected by Carvekit~\cite{carvekit} and HED map $\tilde{H}_k$ by \cite{xie2015holistically} to supervise the training.

For the known viewpoint $c_k$, we obtain the RGB renderings ($I_k$, $\hat{I}_k$), normal renderings ($N_k$, $\hat{N}_k$), and mask images ($M_k$, $\hat{M}_k$) from both grids ($\mathbf{D}$, $\mathbf{\hat{D}}$) respectively. Similarly, we obtain ($I_\text{nv}$, $\hat{I}_\text{nv}$), ($N_\text{nv}$, $\hat{N}_\text{nv}$), and ($M_\text{nv}$, $\hat{M}_\text{nv}$) from ($\mathbf{D}$, $\mathbf{\hat{D}}$) respectively for the novel viewpoint $c_\text{nv}$.

\statement{Image supervision.}
As previously discussed, DreamFusion~\cite{poole2022dreamfusion} relies entirely on textual prompts with a guidance scale of 100, resulting in outputs that are fully dependent on the text.
Instead of calculating SDS loss, we propose exploiting image-level supervision on denoised renderings:
we feed the RGB rendering $I_k$ from the high-resolution grid, along with the text embedding $y$, to the pre-trained diffusion model to produce the ``denoised'' RGB image $\mathcal{I}_k$. 
We then define the image-level loss as
\begin{equation}
    \mathcal{L}_k = \mathtt{MSE}(\mathcal{I}_k, G_k) + \mathtt{SMAPE}(\mathcal{I}_k \cdot A_k, G_k \cdot M_k) + \mathtt{MSE}(\hat{I}_k, G_k) + \mathtt{SMAPE}(\hat{I}_k \cdot A_k, G_k \cdot \hat{M}_k),
\end{equation}
where $\mathtt{SMAPE}(a, b) = {|a - b|}/\{|a| + |b| + 0.01\}$ (see supplementary for analysis on loss functions).
Additionally, for novel viewpoints $c_\text{nv}$, we define the image-level loss as 
\begin{equation}
    \mathcal{L}_\text{nv} = \mathtt{MSE}(I_\text{nv}, \hat{I}_\text{nv}) + \mathtt{SMAPE}(I_\text{nv} \cdot \hat{M}_\text{nv}, \hat{I}_\text{nv} \cdot M_\text{nv})
\end{equation}
$\mathcal{L}_\text{nv}$ then helps facilitate the mutual transfer between low- and high-resolution grids.

\statement{HED boundary supervision.}
To enhance the geometry quality and further address the image inconsistency issue, we introduce additional supervision based on the HED boundary. We use a pre-trained HED boundary model~\cite{xie2015holistically} to detect the HED images $H_k$, $H_k^N$, $\hat{H}_k$, $\hat{H}_k^N$ from $\mathcal{I}_k$, $N_k$, $\hat{I}_k$, $\hat{N}_k$ respectively. We then define the HED boundary loss as 
\begin{equation}
    \mathcal{L}_h = \sum_{\chi \in \{H_k, H_k^N, \hat{H}_k, \hat{H}_k^N\}} \mathtt{MSE} (\chi, \tilde{H}_k),
\end{equation}
Our motivation behind this supervision is two-fold: (1) while using predicted normal from the image is a common supervision~\cite{xiu2023econ}, current normal prediction methods face challenges in predicting reasonable results for images with attributes outside the distribution of training data (\eg robots with wings);
(2) by utilizing the concise information in the sketch-based HED, we can easily identify consistent patterns across different views, thereby enhancing the final geometry.

We also apply eikonal loss for the high-resolution tetrahedral grid to smooth the surface:
\begin{equation}
    \mathcal{L}_{\text{eik}} = (||\nabla_{v_i} s(v_i; \phi)|| - 1)^2.
\end{equation}
The overall training loss function is given by
\begin{equation}
    \mathcal{L} = \lambda_k \mathcal{L}_k + \lambda_\text{nv} \mathcal{L}_\text{nv} + \lambda_h \mathcal{L}_h + \lambda_{\text{eik}} \mathcal{L}_{\text{eik}},
\end{equation}
where $\lambda_k$, $\lambda_\text{nv}$, $\lambda_h$, and $\lambda_{\text{eik}}$ are weights to balance the effect of the respective loss.
\vspace{-0.5em}
\section{Experiments}
\vspace{-0.5em}
In this section, we present experimental results to demonstrate the performance and analyze the effectiveness of our proposed method, comparing it with current state-of-the-art approaches.

\statement{Implementation details.}
Guide3D is built upon the official implementation of \textsc{DMTet}~\cite{munkberg2022extracting, shen2021dmtet} and adopts the denoising diffusion function from the open source project~\cite{stable-dreamfusion} of DreamFusion. 
We use the Ver.1-5 model of Stable Diffusion~\cite{stable-diffusion} and obtain the pre-trained CharTurnerV2~\cite{CharTurnerV2} from the Civitai website.
We utilize the officially released PIFu network which is trained on RenderPeople~\cite{RenderPeople} datasets, and the HED boundary detection model from Xie~{\em et~al.}~\cite{xie2015holistically}. For each provided example, we train our network for 5,000 iterations, taking around 1.25 hours on an NVIDIA RTX 3090Ti GPU. Notably, good results can already be obtained within 1,000 iterations, approximately 15 minutes.

\statement{Baselines.}
We focus on comparisons with stable-DreamFusion~\cite{stable-dreamfusion}, the open source project of DreamFusion~\cite{poole2022dreamfusion}, Latent-NeRF~\cite{metzer2022latent-nerf}, 3DFuse~\cite{seo20233dfuse}, the upgraded version of SJC~\cite{wang2022sjc}, Fantasia3D~\cite{chen2023fantasia3d}, Magic3D~\cite{lin2023magic3d}, and DreamAvatar~\cite{cao2023dreamavatar}. Make-it-3D~\cite{tang2023make}, RealFusion~\cite{melas2023realfusion}, and Zero-1-to-3~\cite{liu2023zero} are excluded from the comparison, as they require images as inputs. 

\subsection{Qualitative evaluations}

\statement{Avatar generation.}
Fig.~\ref{fig:main_results} displays the diverse 3D avatars generated by Guide3D, showcasing topologically and structurally correct geometry and high-resolution texture from various viewpoints. Our method is capable of creating a wide range of avatars, spanning from ordinary people to superheroes and even non-existent characters. See supplementary for more visualizations.

\statement{Comparison with SOTAs.}
We provide qualitative comparisons with SOTA methods in Fig.~\ref{fig:comparison_sota}, \ref{fig:comparison_dreamavatar}, and \ref{fig:compare_sota}. We can observe that: (1) In contrast to other methods that exhibit instability and sensitivity to text prompts, our approach consistently generates 3D avatars with accurate and robust geometry, ensuring topological and structural correctness. (2) While Latent-NeRF and DreamAvatar, which employ the same diffusion model, often produce similar outcomes, our method stands out by generating diverse and versatile results. This is achieved through the text-to-image stage, allowing for a wider range of outputs with distinct characteristics. 

\subsection{Quantitaive evaluations.}
\begin{figure}[t]
    \centering 
    \begin{minipage}{0.4\linewidth}
    \includegraphics[width=\columnwidth]{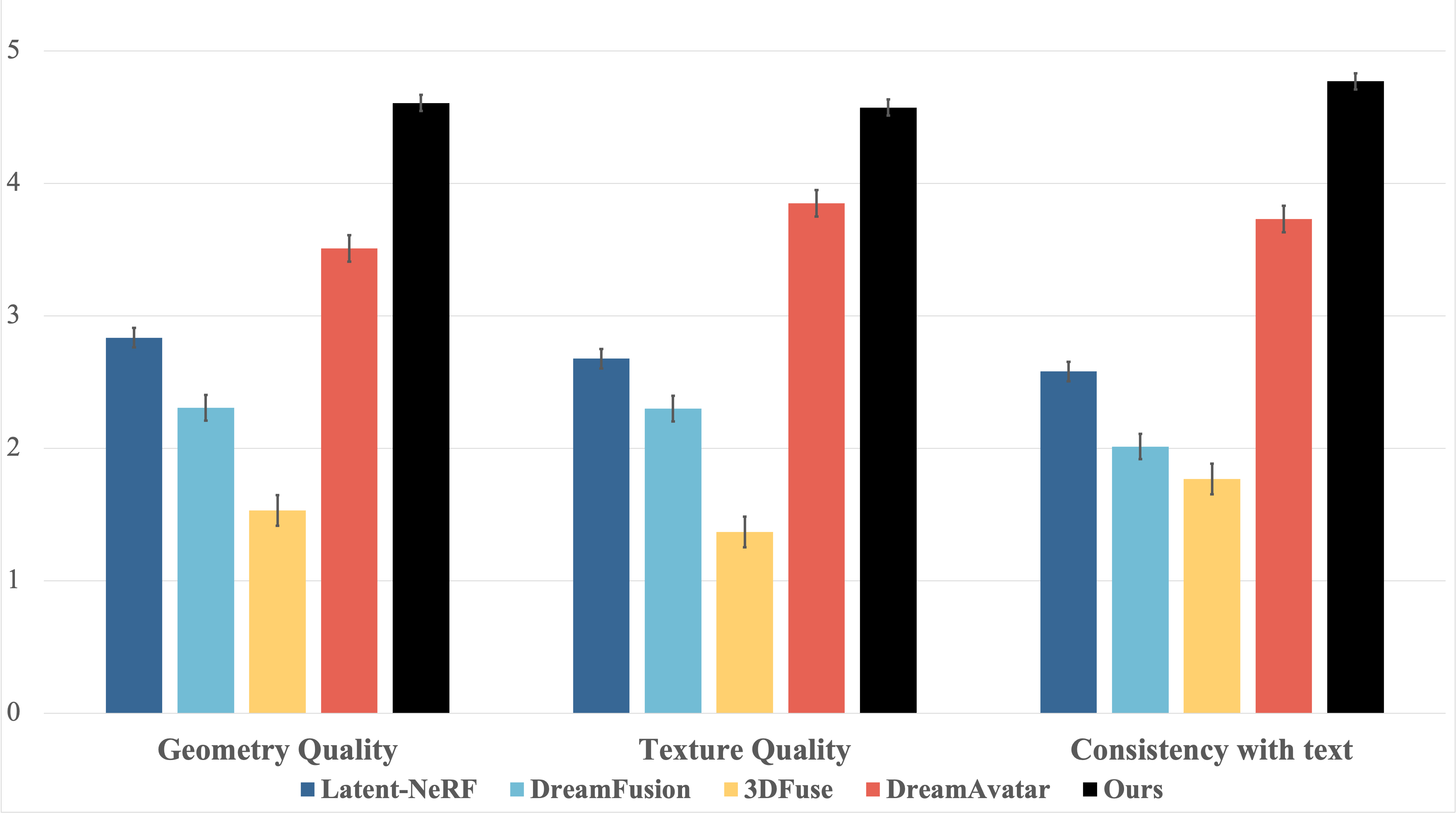}
    \caption{User study.}
    \label{fig:user_study}
    \vspace{-2em}
    \end{minipage}
    \hfill
    \begin{minipage}{0.58\linewidth}
    
        \begin{subtable}
            \centering
            \makebox[\linewidth]{\resizebox{0.4\columnwidth}{!}{
            \centering
            \begin{tabular}{l||cc}
                \toprule
                \multirow{2}*{Metric} & \multicolumn{2}{c}{Image Similarity} \\
                {} & {Ours} & {Ground-Truth} \\
                \midrule
                LPIPS  & 0.107 & 0.000\\         
                SSIM & 0.920 & 1.000\\ 
                \bottomrule
            \end{tabular}
            }}
            \captionsetup{type=table}
            \caption{Quantitative evaluation with image similarity.}
            \label{tab:image_similarity}
        \end{subtable}
        \vspace{0.4em}
        \begin{subtable}
            \centering
            \makebox[\linewidth]{\resizebox{\columnwidth}{!}{
            \begin{tabular}{l||cccccc}
                \toprule
                \multirow{2}*{R-precision} & \multicolumn{6}{c}{Method} \\
                {} & {DreamFusion} & {Latent-NeRF} & {3DFuse} & {DreamAvatar} &{AvatarCLIP} & {Ours} \\
                \midrule
                CLIP B/32   & 87.18 & 88.46 & 53.34 & 93.53 & 64.76 & \textbf{98.96}\\         
                CLIP B/16  & 87.18 & 88.46 & 53.34 & 93.53 & 64.76 & \textbf{98.96}\\ 
                \bottomrule
            \end{tabular}
            }}
            \captionsetup{type=table}
            \caption{Quantitative evaluation with CLIP-based metrics.}
            \label{tab:clip_r}
        \end{subtable}
    \vspace{-2em}
    \end{minipage}
\end{figure}

\paragraph{User studies.}
We conducted user studies comparing with DreamFusion, Latent-NeRF, 3DFuse, and DreamAvatar using rotated videos. 25 volunteers ranked them by geometry quality, texture quality, and consistency with the text, with scores ranging from 1 to 5 (higher means better). Results in Fig.~\ref{fig:user_study} indicate that Guide3D consistently achieves the highest ranks.

\vspace{-0.5em}
\paragraph{Image similarity.}
We further evaluated image-level similarity (LPIPS and SSIM) between generated multi-view images and renderings from our created 3D models. Tab.~\ref{tab:image_similarity} demonstrates that we successfully address the inconsistencies and transfer the image attributes to 3D space.

\vspace{-0.5em}
\paragraph{CLIP R-precision.}
We follow DreamFusion~\cite{poole2022dreamfusion} and performed CLIP R-Precision evaluations with the above methods. As shown in Tab.~\ref{tab:clip_r}, Guide3D outperforms baseline methods.

\subsection{Ablation studies}
\label{sec:ablation}
\begin{figure}[t]
    \centering 
    \setlength{\tabcolsep}{0.195pt}
    \begin{tabular}{ccccc} 
         \includegraphics[align=c,width=0.195\linewidth]{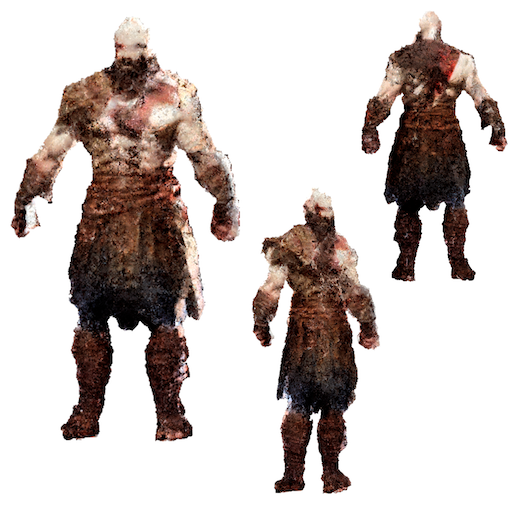}&
        \includegraphics[align=c,width=0.195\linewidth]{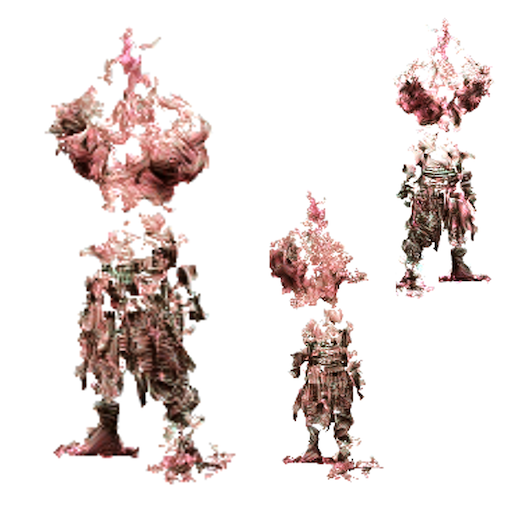} &
        \includegraphics[align=c,width=0.195\linewidth]{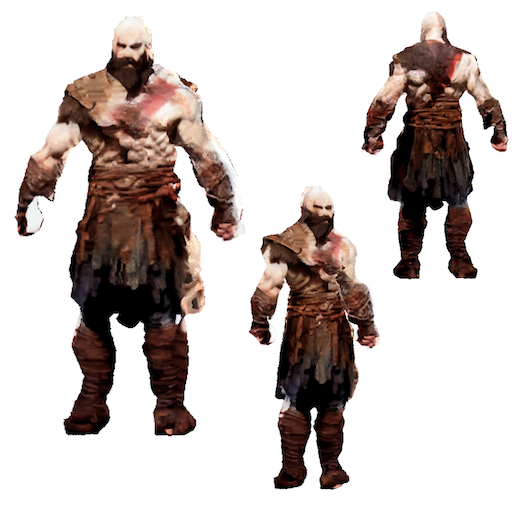}&
        \includegraphics[align=c,width=0.195\linewidth]{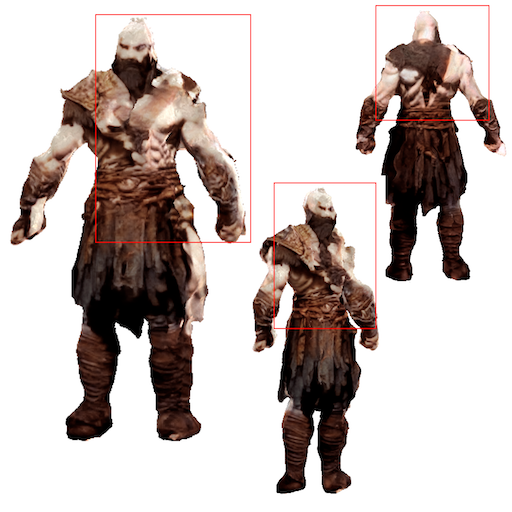}&
        \includegraphics[align=c,width=0.195\linewidth]{Figures/main_results/kratos.png}  \\
        \includegraphics[align=c,width=0.195\linewidth]{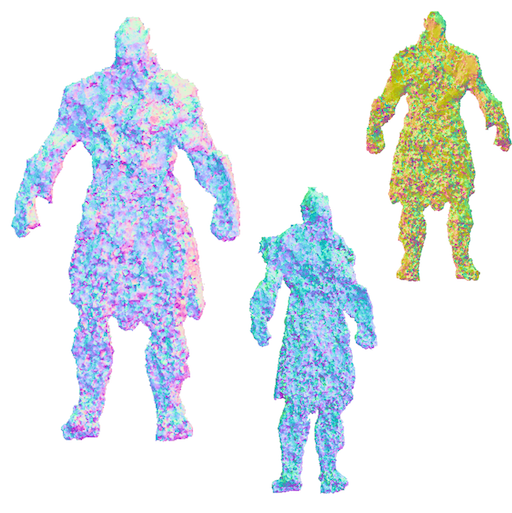}& 
        \includegraphics[align=c,width=0.195\linewidth]{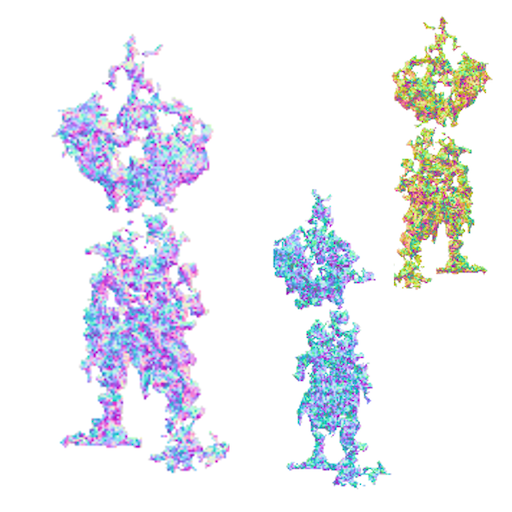} &
        \includegraphics[align=c,width=0.195\linewidth]{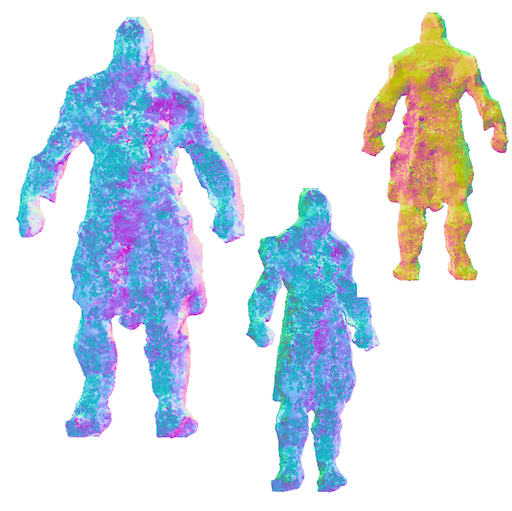}& 
        \includegraphics[align=c,width=0.195\linewidth]{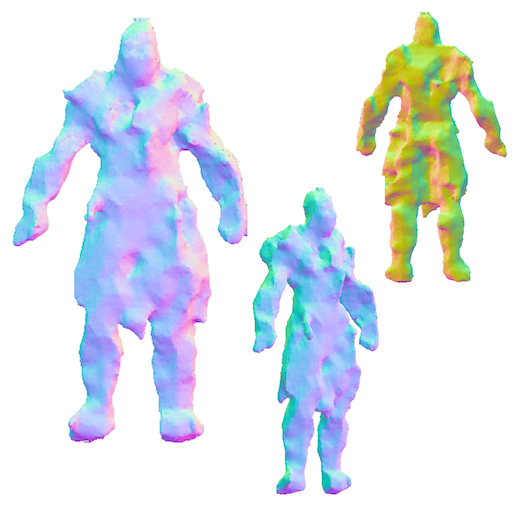}& 
        \includegraphics[align=c,width=0.195\linewidth]{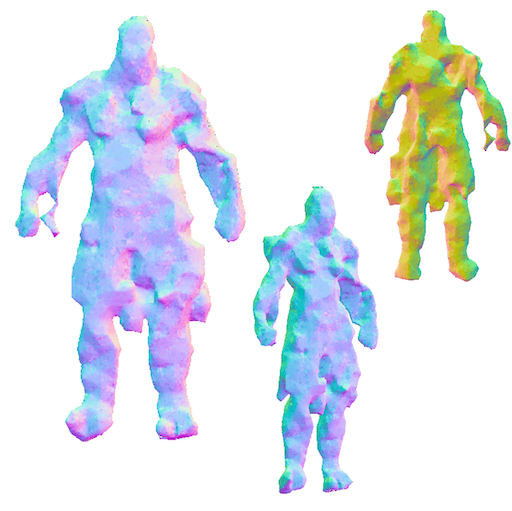}  \\
        \multicolumn{5}{c}{\cprompt{A DSLR photo of Kratos in God of War}} \\
        \tabletitle{\textit{w/o} HED loss} & \tabletitle{\textit{w/} SDS} & \tabletitle{\textit{w/} Normal loss} & \tabletitle{\textit{w/o} Denoising} &\tabletitle{Ours} \\
    \end{tabular}
    \caption{\textbf{Analysis of supervision strategy.} Best viewed in PDF with zoom.}
    \vspace{-1em}
    \label{fig:ablation_supervision}
\end{figure}

\statement{Analysis of supervision strategy.}
In Fig.~\ref{fig:ablation_supervision}, we present an ablation study on proposed training objectives, considering four variants: (1) our model without HED boundary supervision, (2) our model supervised by SDS instead of the proposed denoising setup, (3) the inclusion of normal predicted by ECON~\cite{xiu2023econ}, the current best method to predict normal maps from human-oriented images, as pseudo-ground-truth for supervision, and (4) direct supervision on renderings instead of denoised renderings.
From the comparisons, we draw the following conclusions: (i) The use of SDS can lead the model towards textual guidance, which may conflict with the generated images and result in corrupted 3D results. (ii) Our HED boundary supervision significantly improves smoothness, while normal supervision can harm the geometry. (iii) Our approach to supervising denoised renderings enhances texture quality and addresses inconsistency exhibited by the generated images. 

\statement{Analysis of pipeline components.}
We conducted further experiments to validate the design of our multi-resolution optimization strategy. Specifically, we experimented with four variants of our framework: (1) single high-resolution grid optimization, (2) single low-resolution grid optimization, (3) our multi-resolution pipeline but optimizing the geometry and texture sequentially, and (4) utilizing average pooling to integrate multi-view image features.
Our findings indicate that none of the variants performed as well as our proposed framework. 

Please refer to the supplementary for detailed illustrations and more analysis.
\begin{figure}[t]
    \vspace{-0.5em}
    \centering 
    \setlength{\tabcolsep}{0.195pt}
    \begin{tabular}{ccccc} 
         \includegraphics[align=c,width=0.195\linewidth]{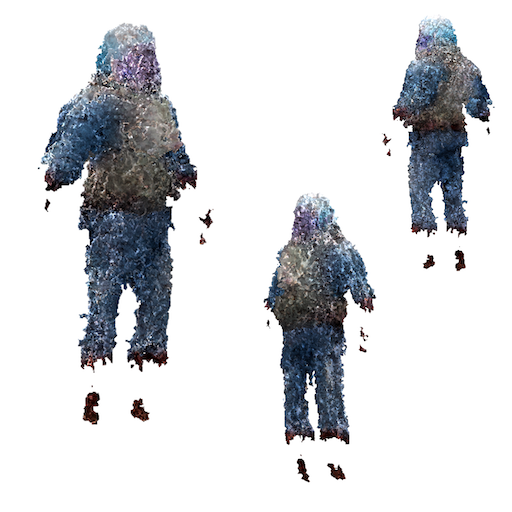}&
        \includegraphics[align=c,width=0.195\linewidth]{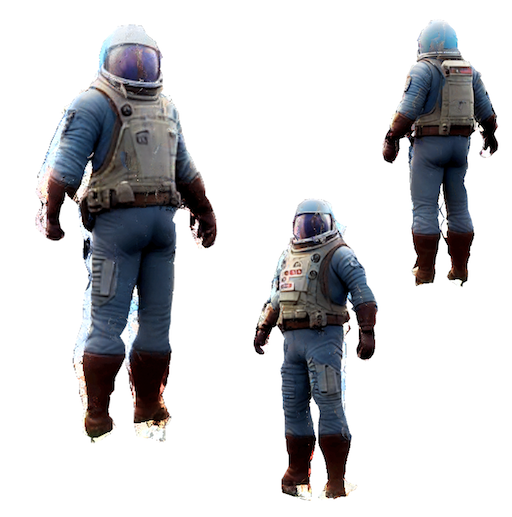} &
        \includegraphics[align=c,width=0.195\linewidth]{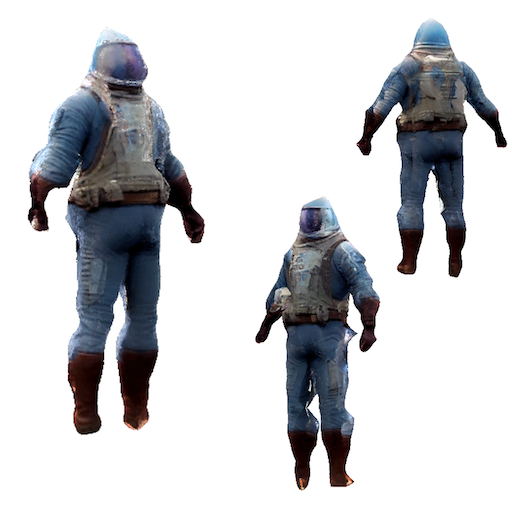}&
        \includegraphics[align=c,width=0.195\linewidth]{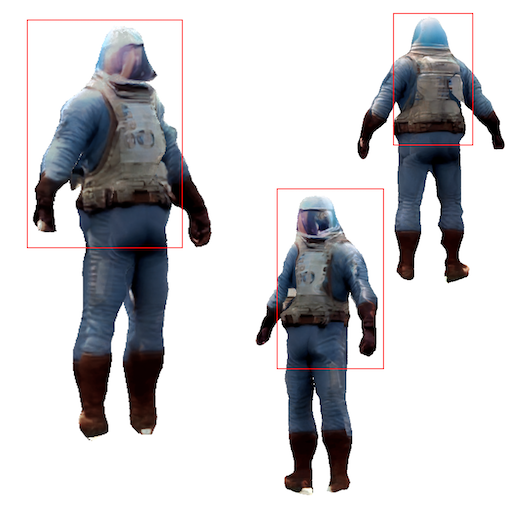}&
        \includegraphics[align=c,width=0.195\linewidth]{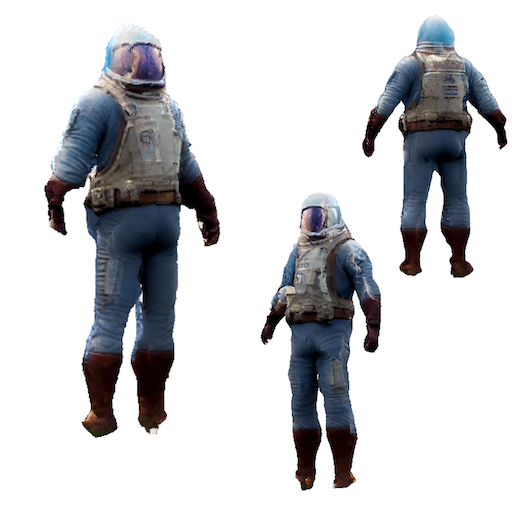}  \\
        \includegraphics[align=c,width=0.195\linewidth]{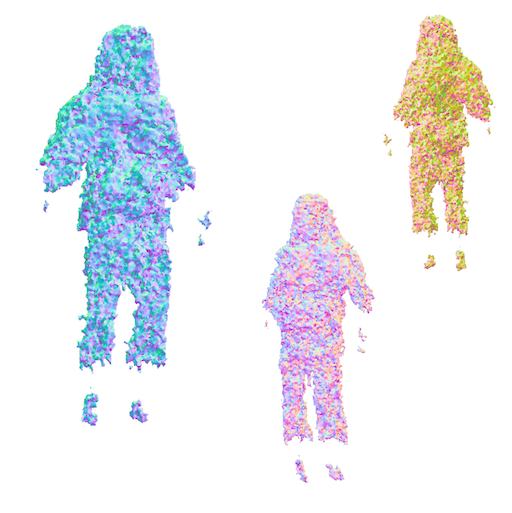}& 
        \includegraphics[align=c,width=0.195\linewidth]{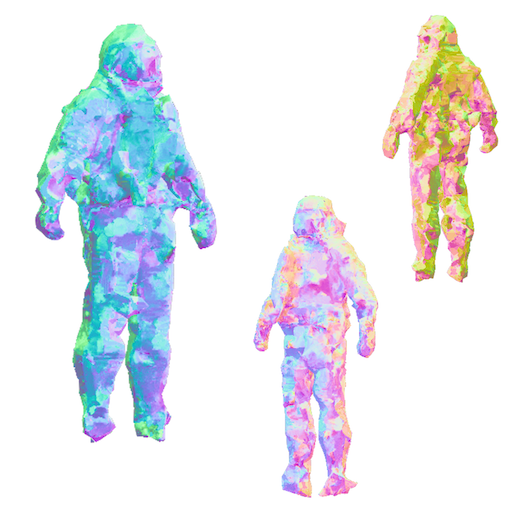} &
        \includegraphics[align=c,width=0.195\linewidth]{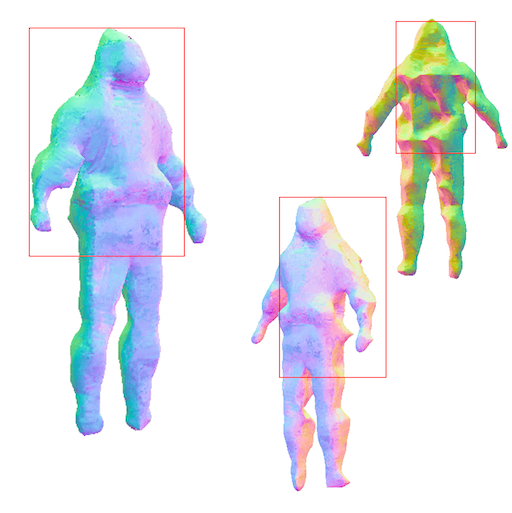}& 
        \includegraphics[align=c,width=0.195\linewidth]{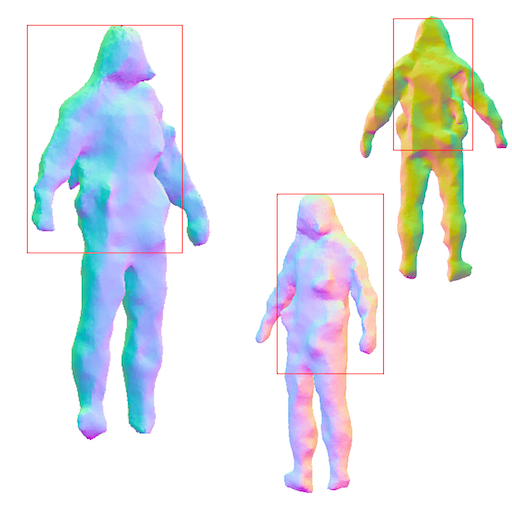}& 
        \includegraphics[align=c,width=0.195\linewidth]{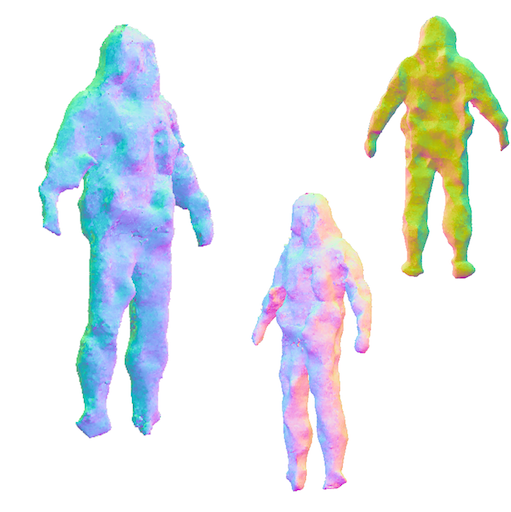}  \\
        \multicolumn{5}{c}{\cprompt{A DSLR photo of a spacesuit}} \\
        \tabletitle{Single high-resolution} & \tabletitle{Single low-resolution} & \tabletitle{Separate geometry \& texture} & \tabletitle{\textit{w/} Average pooling} &\tabletitle{Ours} \\
    \end{tabular}
    \caption{Analysis of pipeline components. Best viewed in PDF with zoom.}
    \vspace{-1em}
    \label{fig:ablation_components}
\end{figure}
\vspace{-1em}
\section{Conclusions}
\vspace{-1em}
Our paper introduces Guide3D, a generative network designed to create 3D avatars based on textual descriptions. We present a novel and robust approach, involving the joint optimization of multi-resolution DMTet grids, by integrating pixel-aligned image features and a new diffusion denoising process. Additionally, we propose a similarity-aware strategy to effectively incorporate consistent image features across different views, along with sketch-based HED training objectives that significantly enhance the geometry. The performance of our method surpasses that of state-of-the-art approaches in generating high-quality 3D avatars with topologically and structurally correct geometry, while also offering extensive research and industry possibilities through the utilization of multi-view images.

\statement{Limitations.}
While Guide3D undoubtedly enhances the performance and expands the possibilities of 3D avatar generation, it does have certain limitations. Firstly, the generated geometry lacks intricate details like clothing wrinkles. Secondly, while the textured mesh is not currently animatable. Thirdly, while Guide3D demonstrates robustness in creating 3D models from multi-view images, occasional failures can still be observed in the generation of multi-view images via diffusion models.

\statement{Societal impact.}
Our proposed generation of 3D avatars from multi-view images can largely help the development of metaverse but also raises associated concerns regarding potential malicious use, especially considering the relative ease of obtaining personal and informative multi-view images compared to text prompts.

\statement{Acknowledgement}
This work is partially supported by Hong Kong Research Grant Council - Early Career Scheme (Grant No. 27208022) and HKU Seed Fund for Basic Research. We also sincerely thank the reviewers for their constructive suggestions.

\appendix
\newpage
\section{Appearance modeling}
We follow nvdiffrec~\cite{munkberg2022extracting} to optimize the geometry, texture maps, and lighting concurrently. Our approach leverages volumetric texturing and uses the world space position $\mathbf{v}$ to index into the texture map, which includes $\mathbf{k}_s$, $\mathbf{k}_d$ and $\mathbf{k}_{orm}$. This ensures that the mapping varies smoothly with both vertex translation and changing topology. See more explanations on the texture maps in the supplementary.

\statement{Material model.}
In accordance with the approach delineated in nvdiffrec~\cite{munkberg2022extracting}, we adopt the Disney's physically-based (PBR) material model~\cite{mcauley2012practical} for advanced appearance rendering. It fuses the diffuse term $k_d$, the $k_{orm}$ term (encompassing occlusion, roughness, and metallic properties), and the normal variation term $k_n$. Among $k_{orm}$, the roughness $r$ operates as a reference to ascertain specular reflection magnitudes while functioning as a parameter for the GGX normal distribution~\cite{walter2007microfacet} within the rendering equation. On the other hand, the metalness term $m$ will be employed together with the diffuse term $k_d$ to obtain the specular term $k_s$, using the formula $k_s = (1 - m) \cdot 0.04 + m \cdot k_d$. In terms of the normal variation $k_n$, it is employed in tangent space to heighten surface illumination and introduce a heightened level of geometric intricacies.

Overall, in our implementation, we apply an MLP to calculate the $k_d$, $k_{orm}$, and $k_n$ for each 3D point within the differentiable tetrahedral grids:
\begin{equation}
    \Phi_{t}(v / \hat{v}) \rightarrow  (k_d, k_{orm}, k_n)
\end{equation}

Subsequently, building upon $k_d$, $k_{orm}$, $k_n$, as well as a learnable environment light that is represented by a high-resolution map, we utilize the image-based lighting model~\cite{munkberg2022extracting} to attain differentiable rendering. Particularly, the rendering equation for each image pixel at a specific viewing direction $\omega_o$ can be computed as follows:
\begin{equation}
L(\omega_o, v) = \int_\Omega L_i(\omega_i)f(\omega_i,\omega_o) (\omega_i \cdot \mathbf {n_v}) d\omega_i,
\end{equation}
where $L(\omega_o, v)$ represents the rendered image color from 3D point $v$ and at $\omega_o$ direction, $L_i(\omega_i)$ is the incident radiance from direction $\omega_i$, 
and $f(\omega_i, \omega_o)$ embodies the BSDF term influenced by the $k_d$, $k_{orm}$ and $k_n$ terms.
In our approach, we ascertain the BSDF term $f(\omega_i, \omega_o)$ as in the Cook-Torrance microfacet specular shading model~\cite{cook1982reflectance}:
\begin{equation}
    f(\omega_i,\omega_o) = \frac {G \ F \ D }{4 (\omega_o \cdot \mathbf {n_v}) (\omega_i \cdot \mathbf {n_v})},
\end{equation}
where $G$ corresponds to the geometric attenuation, $F$ signifies the Fresnel term, and $D$ represents the GGX~\cite{walter2007microfacet} normal distribution.

The rendered image color is subsequently determined by integrating within the hemisphere $\Omega$ around the surface intersection normal ($\mathbf{n_v}$). 

\section{Implementation details}
In addition to the specific network outlines provided in the main paper, here we provide more implementation details for better reproduction. Initially, we obtain our multi-resolution \textsc{DMTet} grids, \ie $\hat{\mathbf{D}}$ at $64^3$ resolution and $\mathbf{D}$ at $256^3$ resolution, from the open-source implementation~\cite{Tetrahedron}. 

Concerning the optimization pipeline, our low-resolution optimization pipeline receives the multi-view fused pixel-aligned image feature $F(\cdot) \in \mathbb{R}^{256}$ as input, followed by the application of a multi-layer perceptron (MLP) to predict the signed distance value $\hat{s}(\hat{v}_i)$ and position offset $\delta \hat{v}_i$ for low-resolution points. The MLP encompasses three layers with dimensions [256, 64, 64, 1+3], where `1' and `3' denote $\hat{s}(\hat{v}_i)$ and $\delta \hat{v}_i$, respectively.  In contrast, for high-resolution optimization, we directly utilize the 3D point $v_i$ as input and employ a hash-grid frequency encoder $\gamma(\cdot) \in \mathbb{R}^{32}$ to elevate the point to a higher dimension. Analogously, an MLP is applied to predict the signed distance value and position offset for high-resolution grids, which consists of layers with dimensions [32, 64, 64, 1+3].

\clearpage
\section{Further analysis}

\subsection{Effectiveness of multi-resolution optimization}
\begin{figure}[htbp]
    \centering 
    \setlength{\tabcolsep}{0.249pt}
    \begin{tabular}{cccc} 
         \includegraphics[align=c,width=0.249\linewidth]{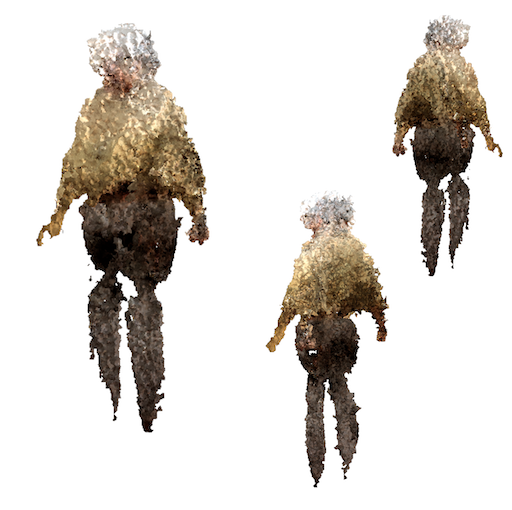}&
        \includegraphics[align=c,width=0.249\linewidth]{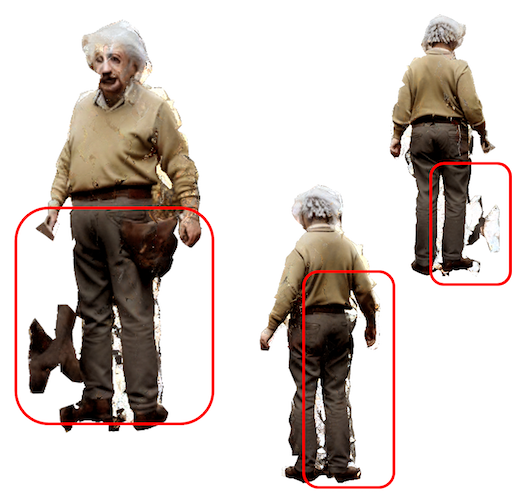} &
        \includegraphics[align=c,width=0.249\linewidth]{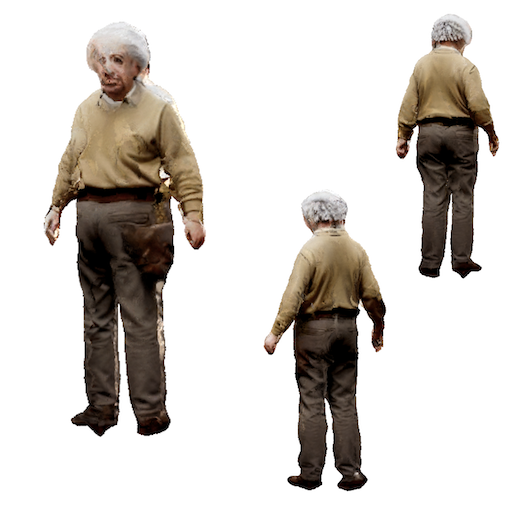}&
        \includegraphics[align=c,width=0.249\linewidth]{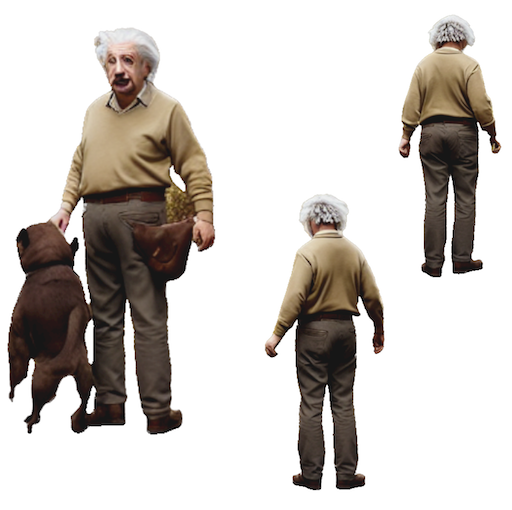}  \\
        \includegraphics[align=c,width=0.249\linewidth]{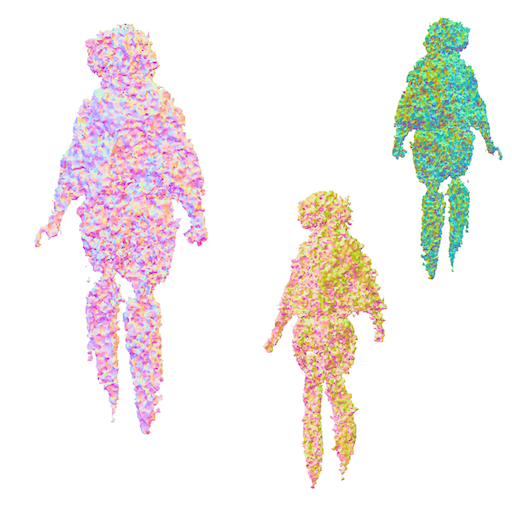}& 
        \includegraphics[align=c,width=0.249\linewidth]{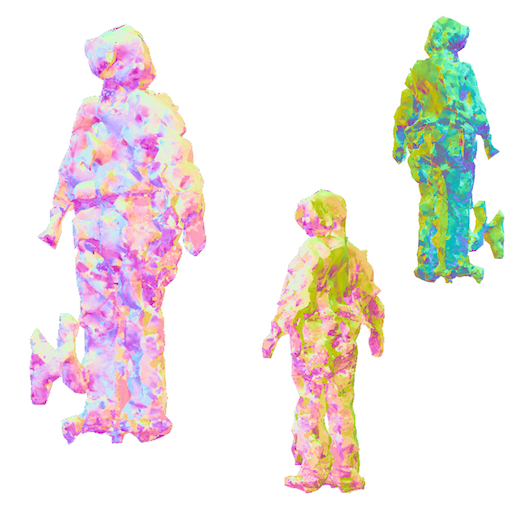} &
        \includegraphics[align=c,width=0.249\linewidth]{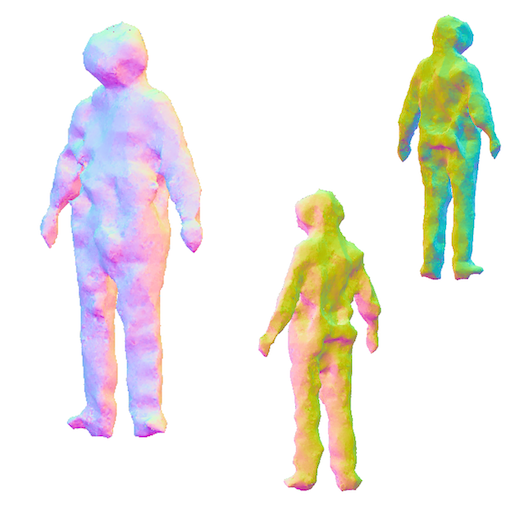}& 
        \includegraphics[align=c,width=0.249\linewidth]{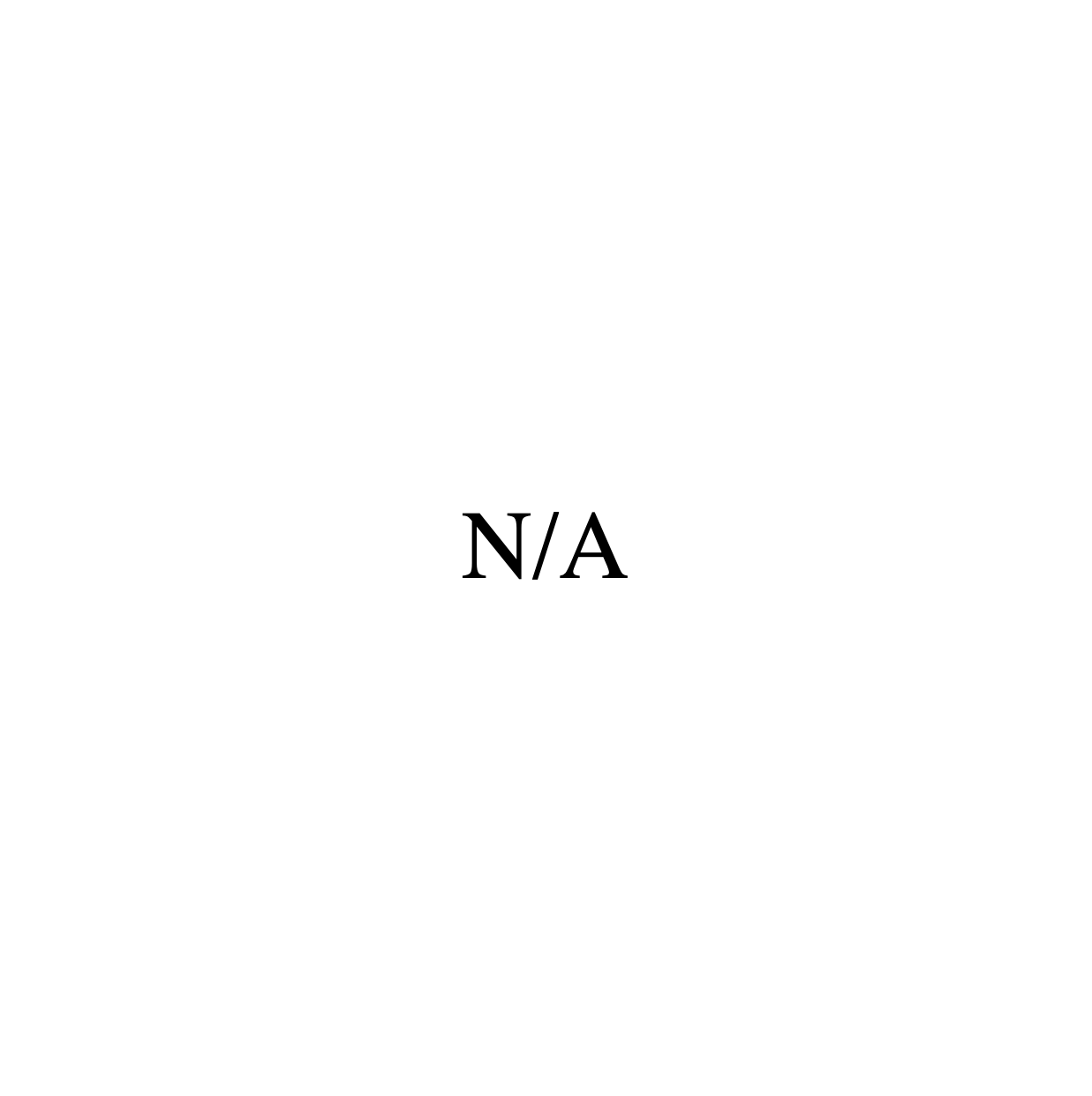}  \\
        \multicolumn{4}{c}{\cprompt{Albert Einstein}} \\
        
         \includegraphics[align=c,width=0.249\linewidth]{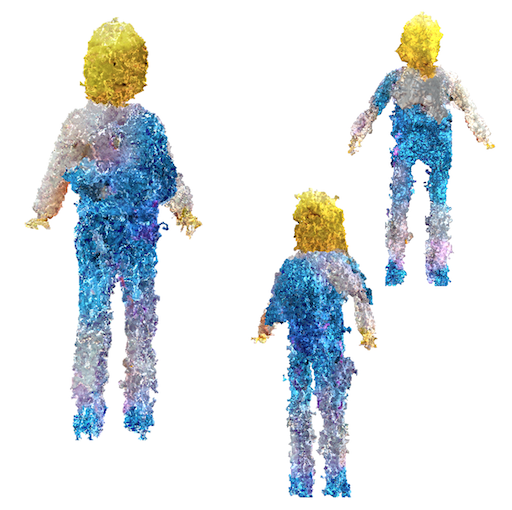}&
        \includegraphics[align=c,width=0.249\linewidth]{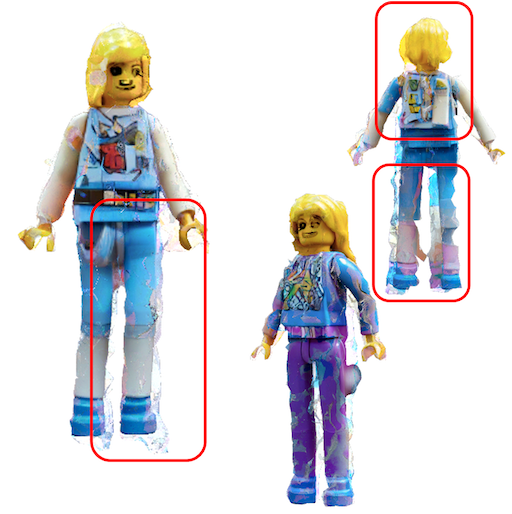} &
        \includegraphics[align=c,width=0.249\linewidth]{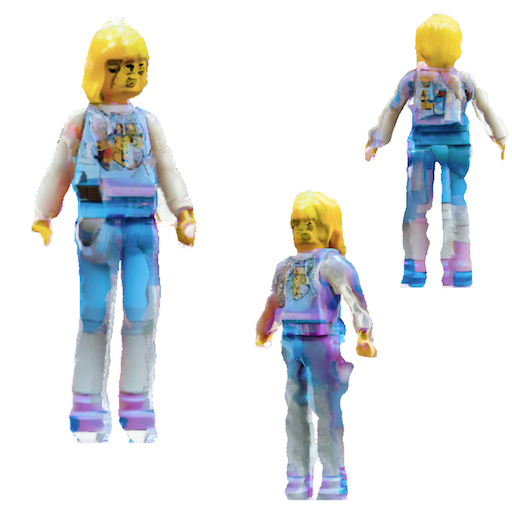}&
        \includegraphics[align=c,width=0.249\linewidth]{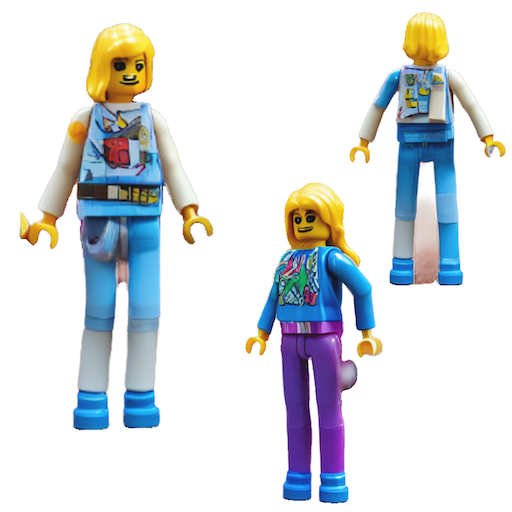}  \\
        \includegraphics[align=c,width=0.249\linewidth]{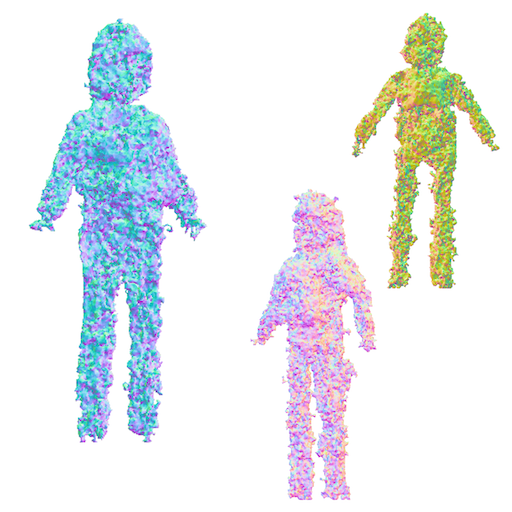}& 
        \includegraphics[align=c,width=0.249\linewidth]{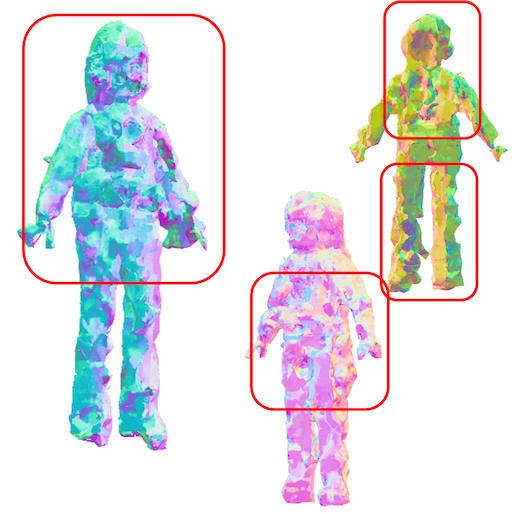} &
        \includegraphics[align=c,width=0.249\linewidth]{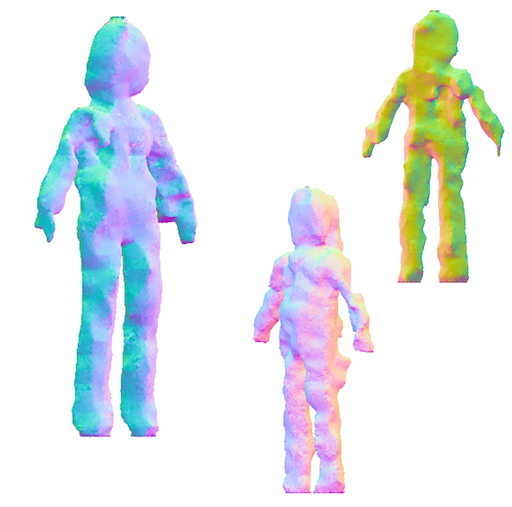}& 
        \includegraphics[align=c,width=0.249\linewidth]{Figures/supplementary/ablation-multi/na.png}  \\
        \multicolumn{4}{c}{\cprompt{Lego friend figurine}} \\
        \tabletitle{Single high-resolution} & \tabletitle{Single low-resolution} & \tabletitle{Ours (multi-resolution)} &\tabletitle{Generated images} \\
    \end{tabular}
    \caption{Analysis on multi-resolution tetrahedral grids.}
    \label{fig:ablation_multi}
\end{figure}

Beyond the ablation studies presented in the main paper, we further assess two discussed variations: (1) single high-resolution grid optimization, \ie the upper branch of Figure 3 in the main paper, and (2) single low-resolution grid optimization, \ie the lower branch of Figure 3 in the main paper. As evident from Fig.~\ref{fig:ablation_multi}, the use of the diffusion denoising function to alleviate inconsistency among multi-view generated images may inadvertently introduce conflicting gradients, adversely impacting the final results (see results from ``single high-resolution''). On the other hand, while the low-resolution grid optimization can produce pleasing texture for reference views, it struggles to (1) achieve exemplary quality in both novel view rendering and geometry, and (2) address the persisting inconsistency concerns that present in the generated multi-view images.

On the contrary, our multi-resolution optimization strategy effectively tackles these challenges by enabling seamless information exchange between low- and high-resolution grids, resulting in superior performance.

\clearpage

\subsection{Effectiveness of denoising-based supervision}
In order to address the persisting inconsistency among the generated images, our training process aims to minimize the difference between generated images and denoised renderings from the high-resolution tetrahedral grid. We proceed to deactivate this denoising configuration and present an ablation analysis in Fig.~\ref{fig:ablation_2}. As evident from the results, our proposed denoising setup fosters the creation of high-resolution textures that are better aligned with the generated images and exhibit reduced blurriness.

\subsection{Effectiveness of similarity-aware feature fusion}
\begin{figure}[htbp]
    \centering 
    \setlength{\tabcolsep}{0.249pt}
    \begin{tabular}{cccc} 
         \includegraphics[align=c,width=0.249\linewidth]{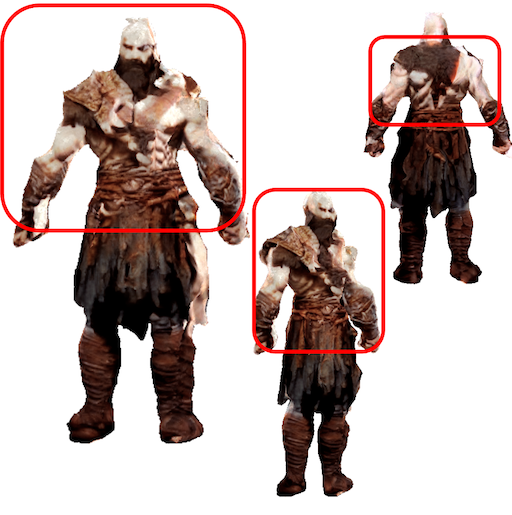}&
        \includegraphics[align=c,width=0.249\linewidth]{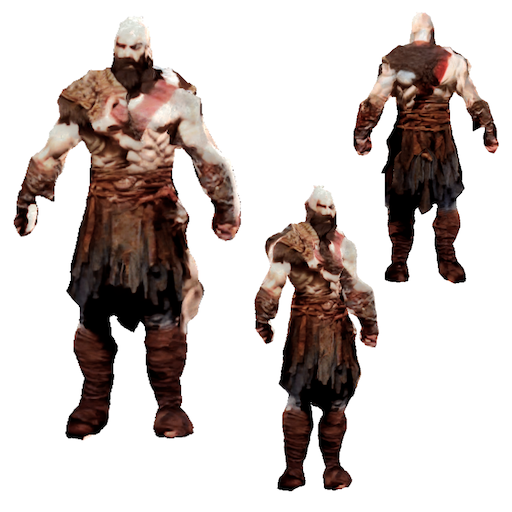} &
        \includegraphics[align=c,width=0.249\linewidth]{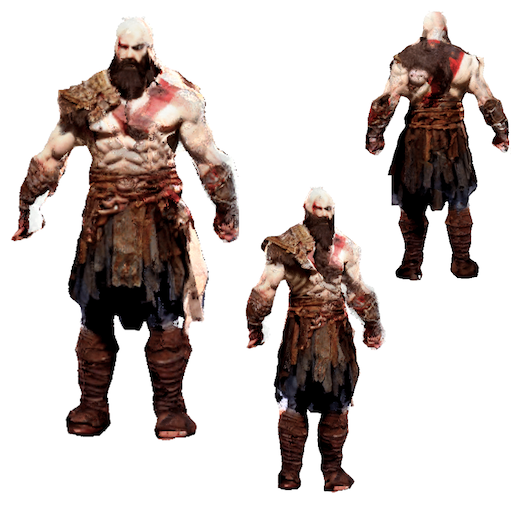}&
        \includegraphics[align=c,width=0.249\linewidth]{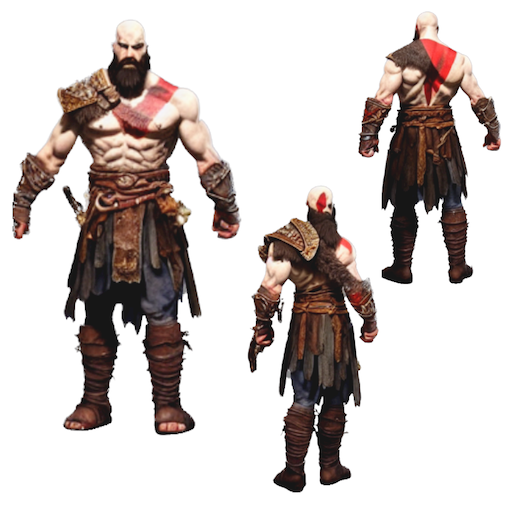}  \\
        \includegraphics[align=c,width=0.249\linewidth]{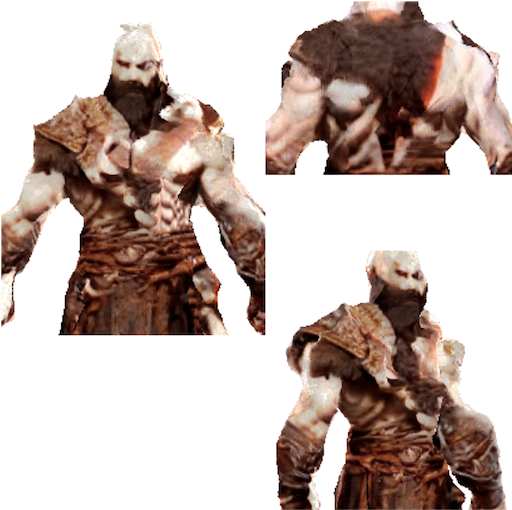}& 
        \includegraphics[align=c,width=0.249\linewidth]{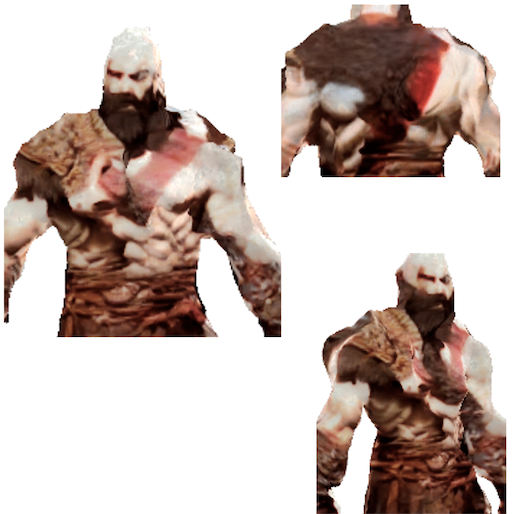} &
        \includegraphics[align=c,width=0.249\linewidth]{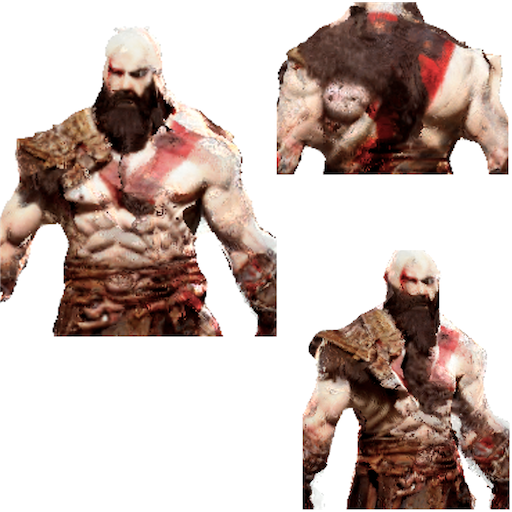}& 
        \includegraphics[align=c,width=0.249\linewidth]{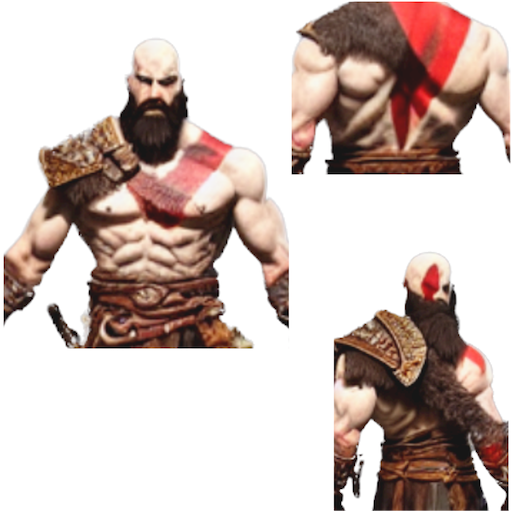}  \\
        \multicolumn{4}{c}{\cprompt{Kratos in God of War}} \\
        
         \includegraphics[align=c,width=0.249\linewidth]{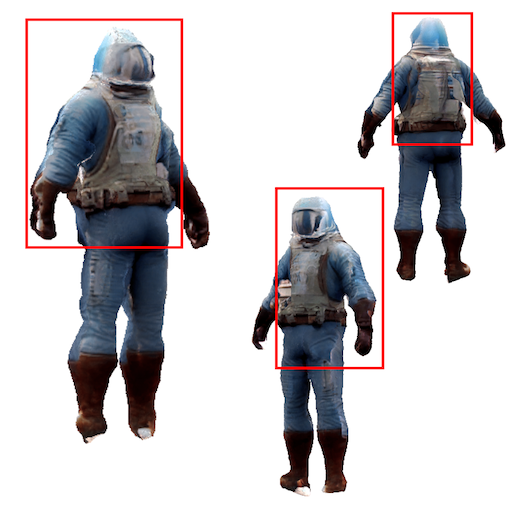}&
        \includegraphics[align=c,width=0.249\linewidth]{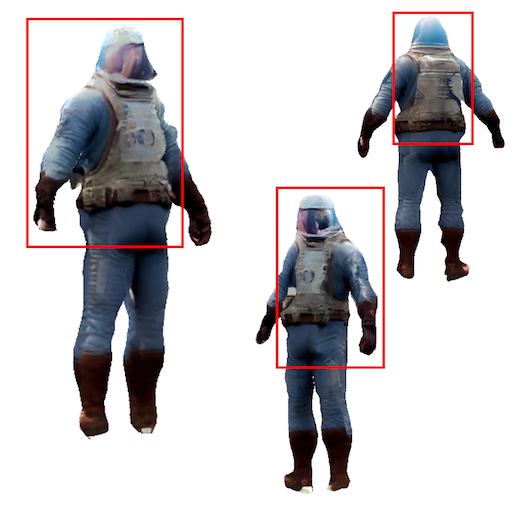} &
        \includegraphics[align=c,width=0.249\linewidth]{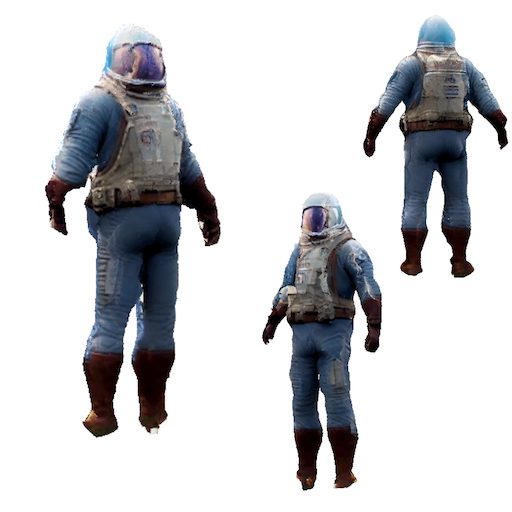}&
        \includegraphics[align=c,width=0.249\linewidth]{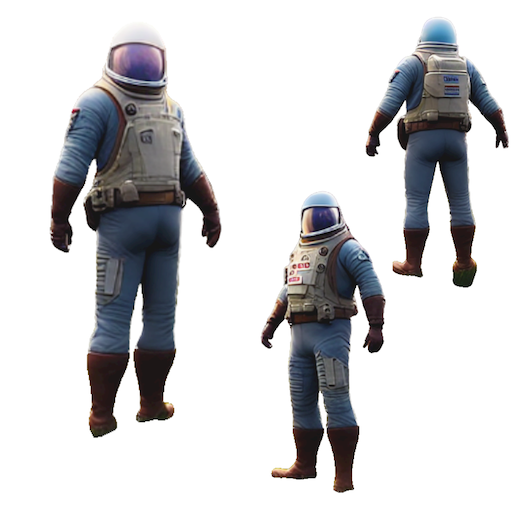}  \\
        \includegraphics[align=c,width=0.249\linewidth]{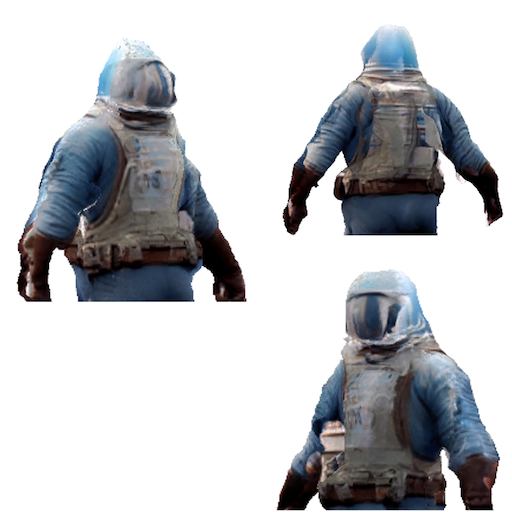}& 
        \includegraphics[align=c,width=0.249\linewidth]{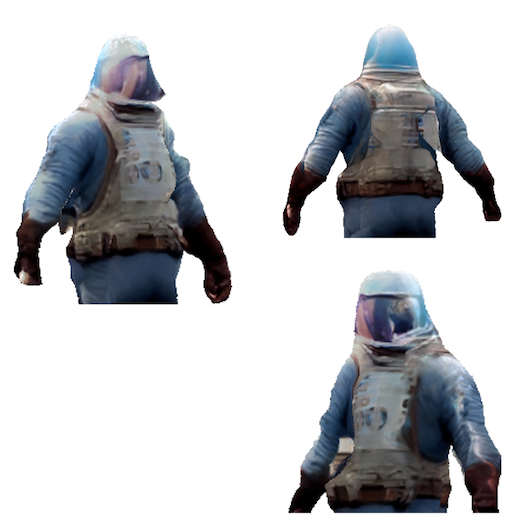} &
        \includegraphics[align=c,width=0.249\linewidth]{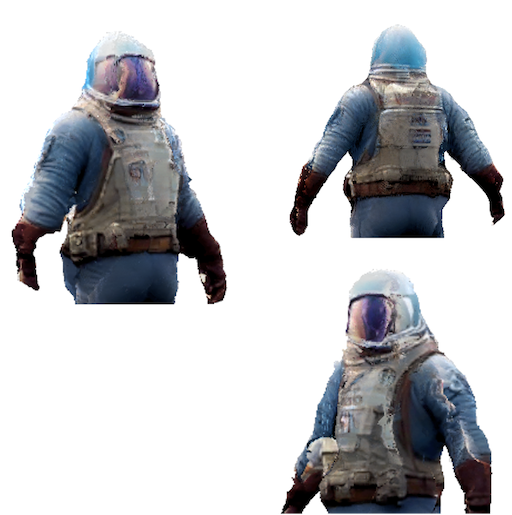}& 
        \includegraphics[align=c,width=0.249\linewidth]{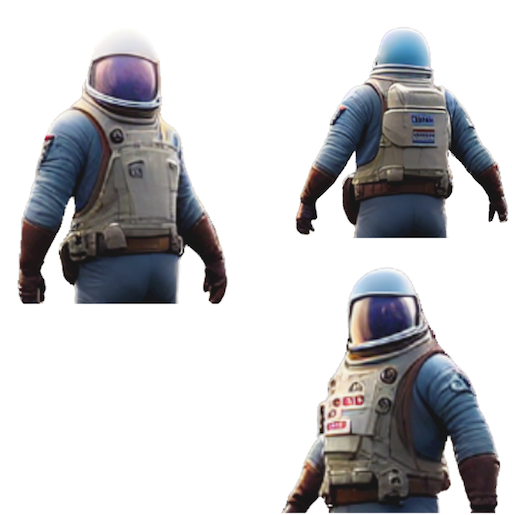}  \\
        \multicolumn{4}{c}{\cprompt{Spacesuit}} \\
        \tabletitle{\textit{w/o} Denoising} & \tabletitle{\textit{w/} Average} & \tabletitle{Ours} &\tabletitle{Generated images} \\
    \end{tabular}
    \caption{Analysis on similarity-aware feature fusion and denoising-based supervision.}
    \label{fig:ablation_2}
\end{figure}

In this section, we then provide further evaluations of the similarity-aware feature fusion strategy by comparing it with its variance, "Average". As depicted in Fig.~\ref{fig:ablation_2}, our proposed similarity-aware feature fusion strategy effectively discerns the most prevalent attributes within the generated images. In turn, our setup can help enhance the quality of geometry and texture, resulting in high-resolution and highly pertinent textures that complement the generated images.

\clearpage
\subsection{Denoising choice}
\begin{figure}[htbp]
    \centering 
    \setlength{\tabcolsep}{0.5pt}
    \begin{tabular}{cc} 
        \includegraphics[align=c,width=0.49\linewidth]{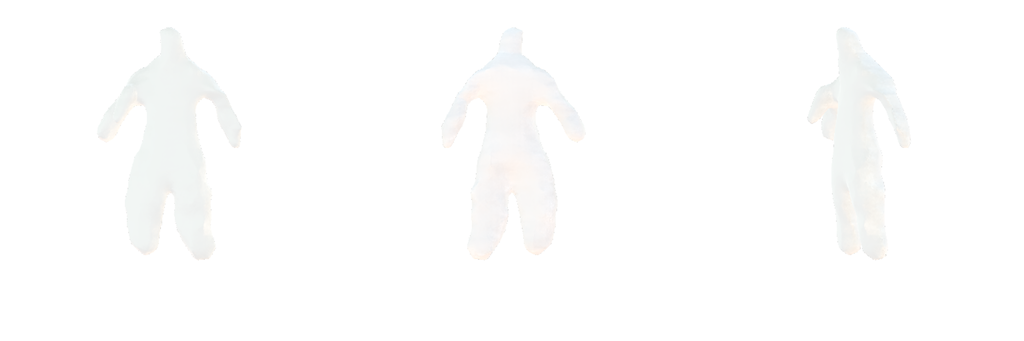}&
        \includegraphics[align=c,width=0.49\linewidth]{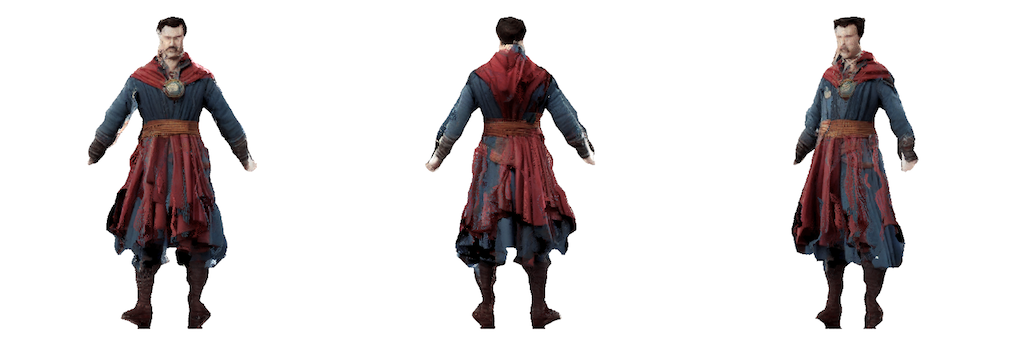} \\
        \multicolumn{2}{c}{\cprompt{Doctor Strange}} \\
        
        \includegraphics[align=c,width=0.49\linewidth]{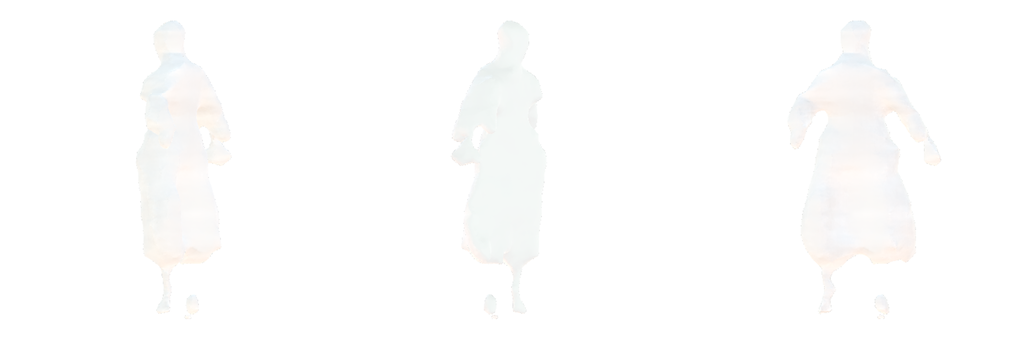}&
        \includegraphics[align=c,width=0.49\linewidth]{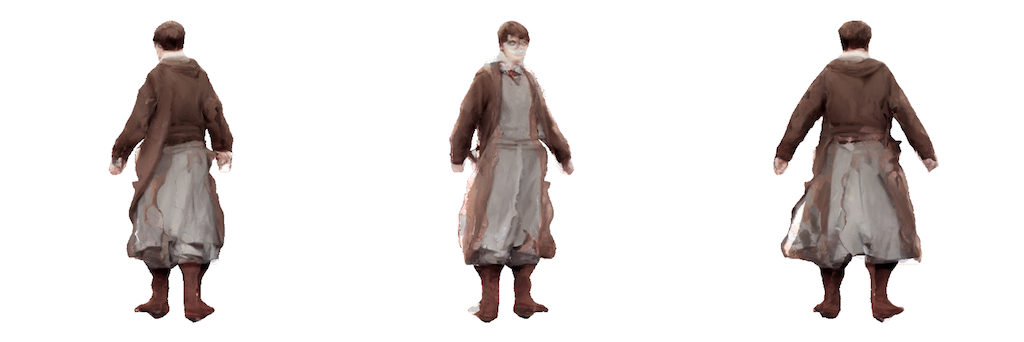} \\
        \multicolumn{2}{c}{\cprompt{Harry Potter}} \\ \tabletitle{(A)} &\tabletitle{(B)} \\
        
    \end{tabular}
    \caption{Ablation study on denoising setup. (A): denoise the renderings from both grids, (B): denoise the rendering from only the high-resolution grid (ours).
    }
    \label{fig:ablation-denoising}
\end{figure}

Beyond the preceding experiments, we further examine our approach by denoising the renderings originating from both low-resolution and high-resolution tetrahedral grids, showcasing visual comparisons in Fig.~\ref{fig:ablation-denoising}.  Our findings demonstrate that this configuration generates conflicting gradients respectively from low-resolution and high-resolution pipelines during the optimization, yielding undesirable results. In contrast, our design achieves good results with high-resolution texture and fine-grained geometry.

\subsection{Loss functions}
\begin{figure}[htbp]
    \centering 
    \setlength{\tabcolsep}{0.2pt}
    \includegraphics[align=c,width=0.98\linewidth]{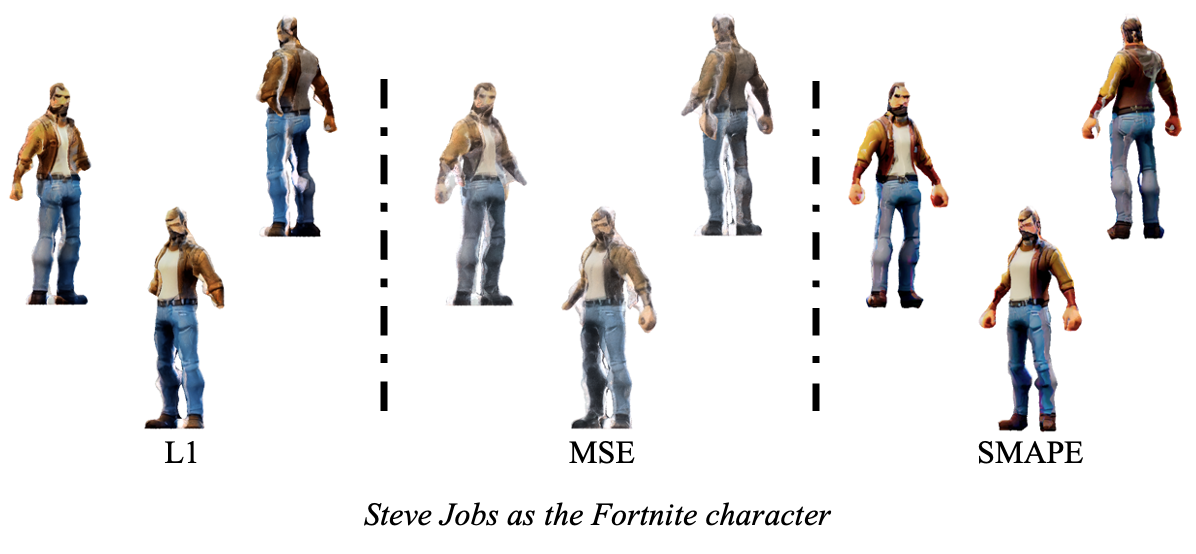}
    \caption{Evaluations on loss functions}
    \label{fig:loss_function}
\end{figure}

We evaluate the choice of loss function and compare with the results from L1 and MSE loss. Results provided in Fig.~\ref{fig:loss_function} demonstrate that our approach incorporating the SMAPE function outperforms the other loss functions.

\clearpage

\subsection{Number of views}
\begin{figure}[htbp]
    \centering 
    \setlength{\tabcolsep}{0.2pt}
    \includegraphics[align=c,width=0.8\linewidth]{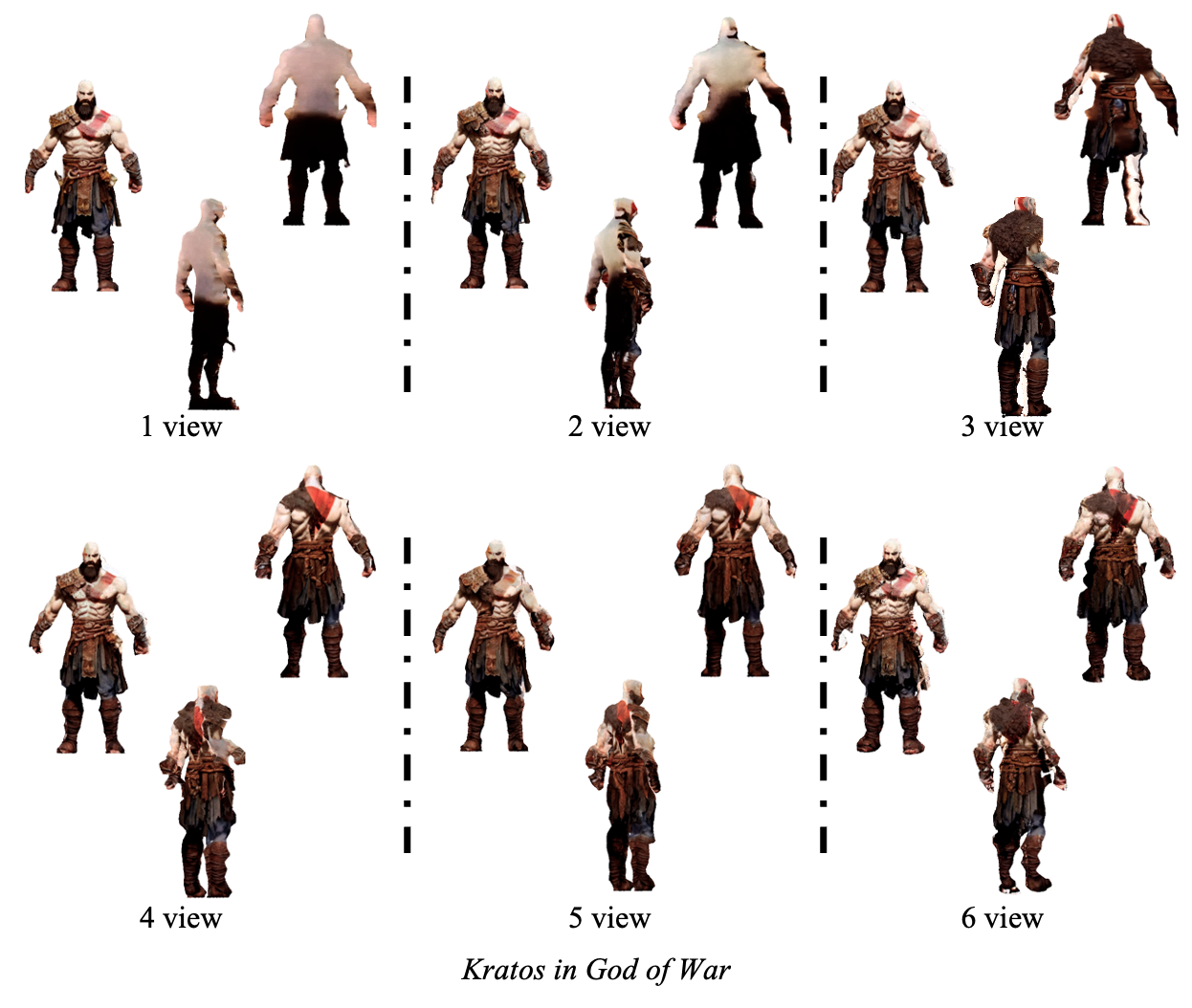}
    \caption{Evaluations on the number of views}
    \label{fig:number_of_views}
\end{figure}

We experimented with varying views from 1 to 6 and found that increasing the number of views will lead to better performance (see results provided in Fig.~\ref{fig:number_of_views} We conducted experiments using up to six views primarily because of the limitations posed by textual inversion and the restricted resolution of ControlNet. However, we are of the opinion that incorporating more views in the real-world applications would be advantageous.

\vspace{-1em}

\vspace{-2em}
\subsection{Pixel-aligned image features}
\begin{figure}[htbp]
    \centering 
    \setlength{\tabcolsep}{0.2pt}
    \includegraphics[align=c,width=0.8\linewidth]{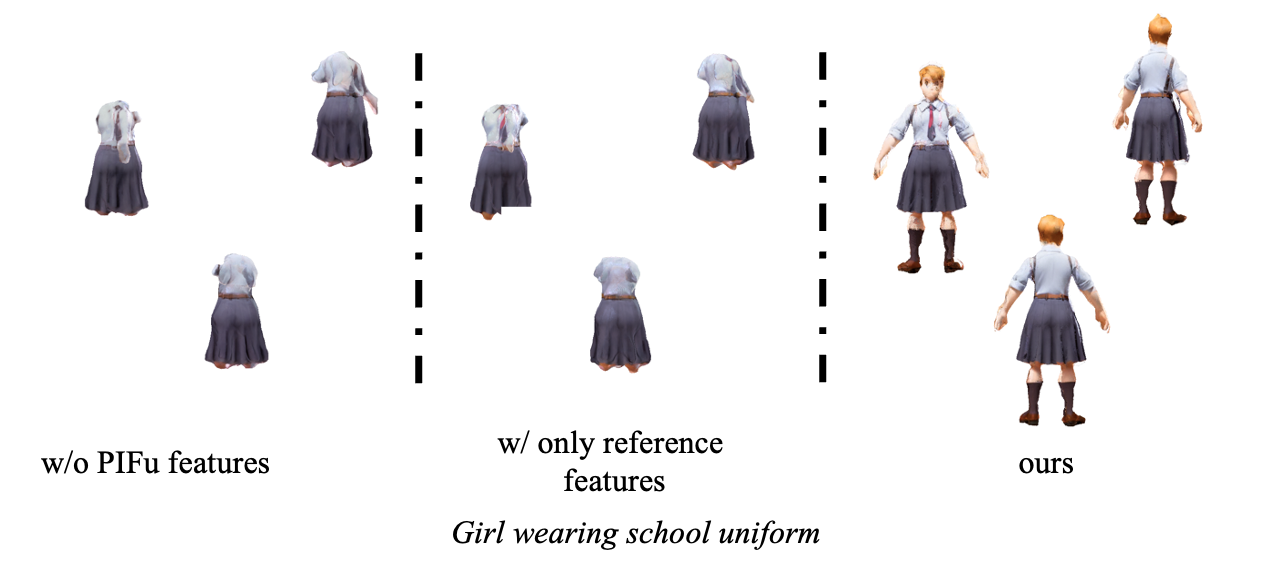}
    \caption{Evaluations on pixel-aligned image features}
    \label{fig:features}
\end{figure}

We performed evaluations to assess the effectiveness of pixel-aligned image features, and present the ablation results in Figure~\ref{fig:features}. 
\vspace{-1em}
\vspace{-1em}

Based on above results, pixel-aligned image features provide the network with stable 3D human priors, as the image encoder is pre-trained and locked during the training. Additionally, multi-view image features provide a more comprehensive and robust understanding of the human in 3D space compared to single reference-view image features that are limited to the frontal view. For these experiments, the diffusion denoising process remains active during the training. The gradient from the diffusion model is very strong, unstable, and lacking alignment with the multi-view images, consequently leading to missed limbs or other structures within the generated outputs.

\clearpage
\section{Additional qualitative comparisons}
\begin{figure}[htbp]
    \centering 
    \setlength{\tabcolsep}{0.195pt}
    \begin{tabular}{ccccc} 
         \includegraphics[align=c,width=0.195\linewidth]{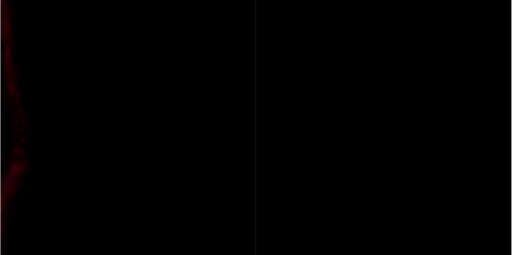}&
        \includegraphics[align=c,width=0.195\linewidth]{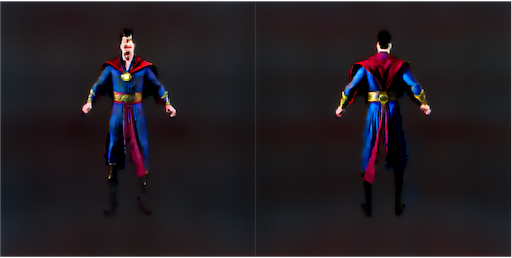} &
        \includegraphics[align=c,width=0.195\linewidth]{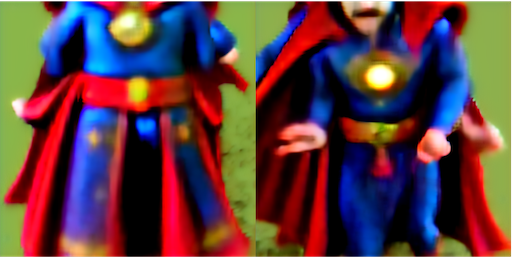}&
        \includegraphics[align=c,width=0.195\linewidth]{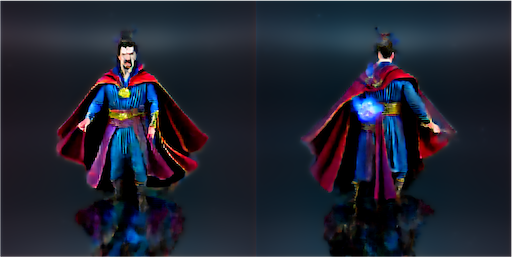}
        &
        \includegraphics[align=c,width=0.195\linewidth]{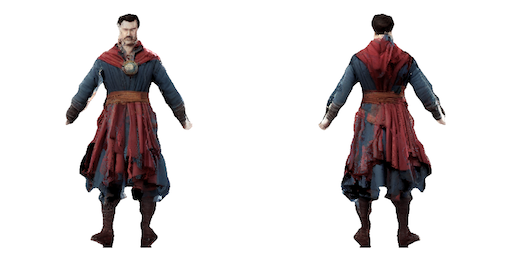}\\
        \multicolumn{5}{c}{\cprompt{Doctor Strange}} 
        \\
        \includegraphics[align=c,width=0.195\linewidth]{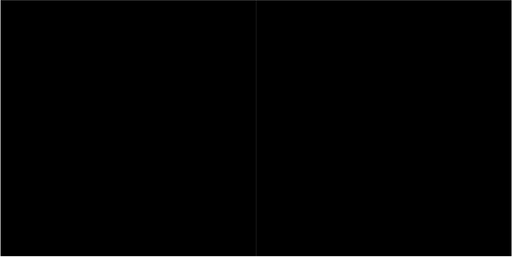}&
        \includegraphics[align=c,width=0.195\linewidth]{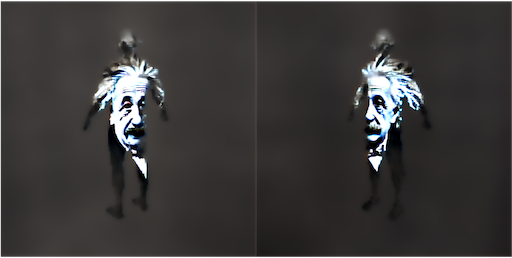} &
        \includegraphics[align=c,width=0.195\linewidth]{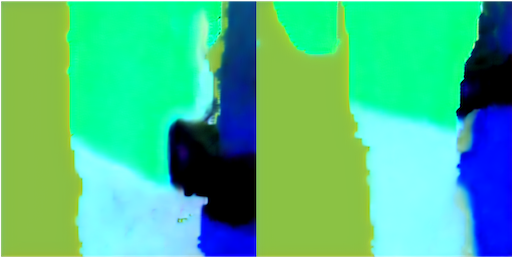}&
        \includegraphics[align=c,width=0.195\linewidth]{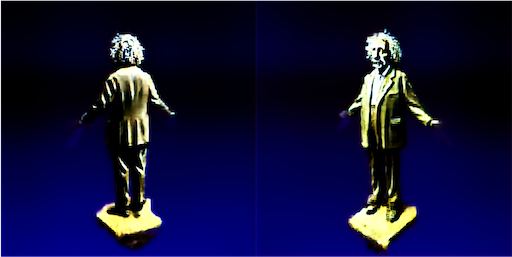}
        &
        \includegraphics[align=c,width=0.195\linewidth]{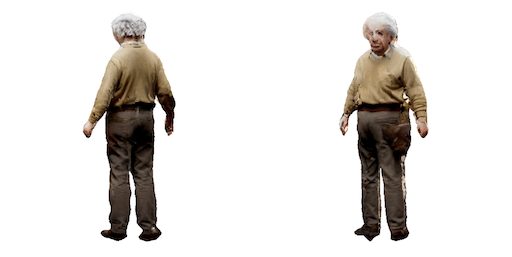}\\
        \multicolumn{5}{c}{\cprompt{Albert Einstein}}
        \\\includegraphics[align=c,width=0.195\linewidth]{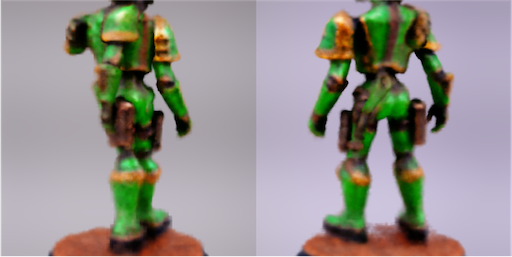}&
        \includegraphics[align=c,width=0.195\linewidth]{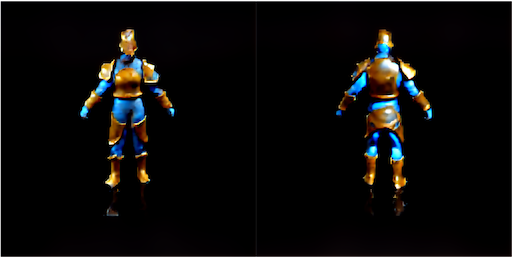} &
        \includegraphics[align=c,width=0.195\linewidth]{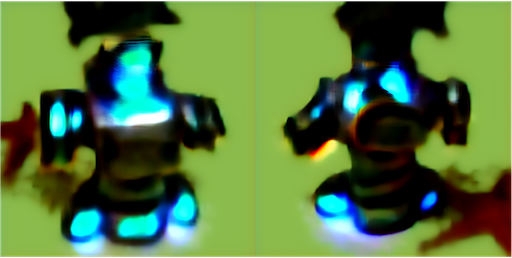}&
        \includegraphics[align=c,width=0.195\linewidth]{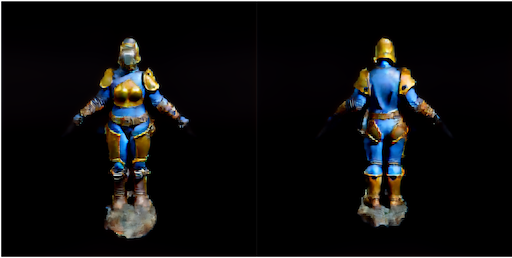}&
        \includegraphics[align=c,width=0.195\linewidth]{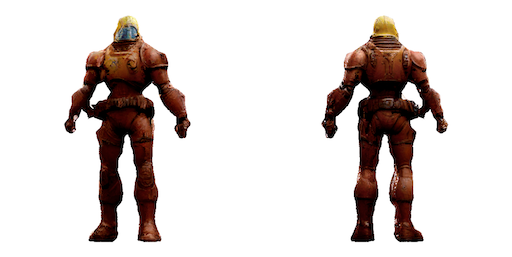}\\
        \multicolumn{5}{c}{\cprompt{Fallout armor figurine}} \\
        \includegraphics[align=c,width=0.195\linewidth]{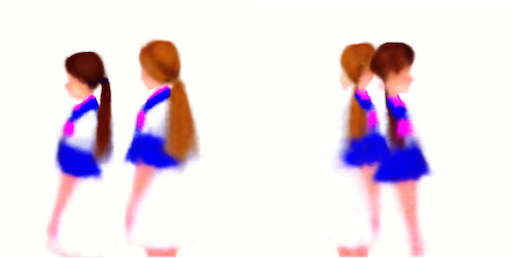}&
        \includegraphics[align=c,width=0.195\linewidth]{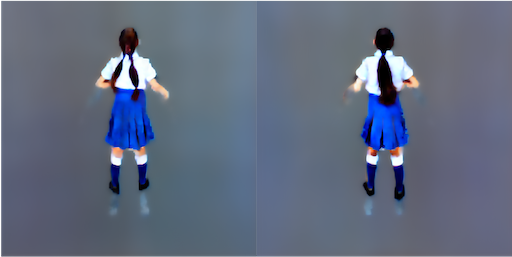} &
        \includegraphics[align=c,width=0.195\linewidth]{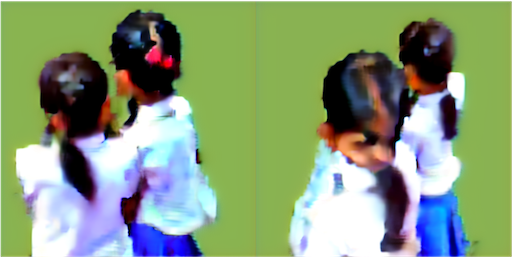}&
        \includegraphics[align=c,width=0.195\linewidth]{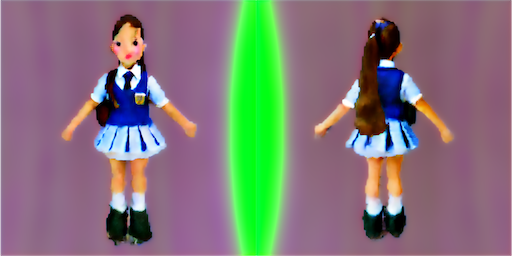}&
        \includegraphics[align=c,width=0.195\linewidth]{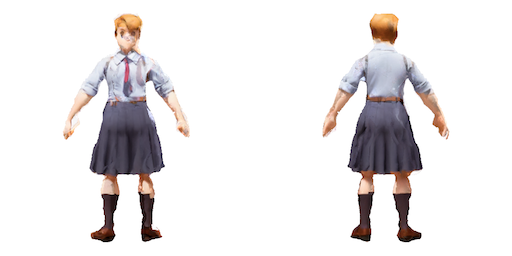}\\
        \multicolumn{5}{c}{\cprompt{Girl wearing school uniform}} \\
        \includegraphics[align=c,width=0.195\linewidth]{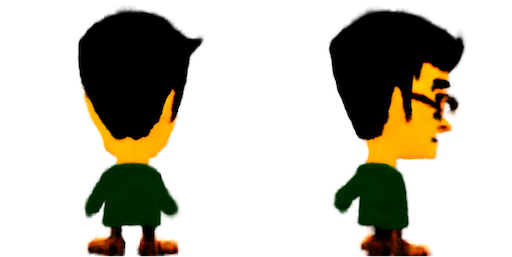}&
        \includegraphics[align=c,width=0.195\linewidth]{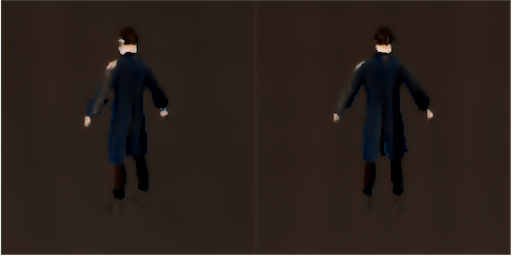} &
        \includegraphics[align=c,width=0.195\linewidth]{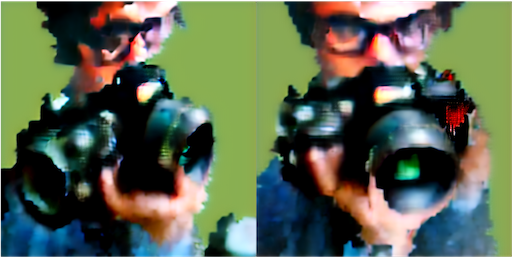}&
        \includegraphics[align=c,width=0.195\linewidth]{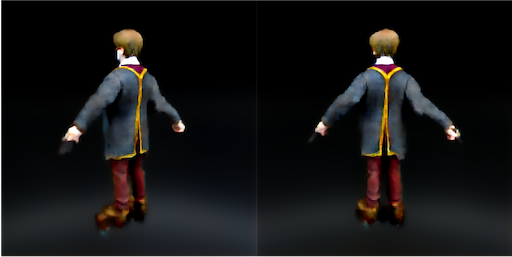}&
        \includegraphics[align=c,width=0.195\linewidth]{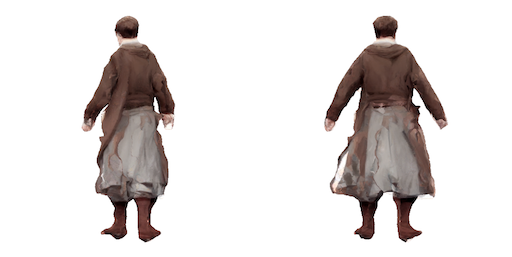}\\
        \multicolumn{5}{c}{\cprompt{Harry Potter}} \\
        \includegraphics[align=c,width=0.195\linewidth]{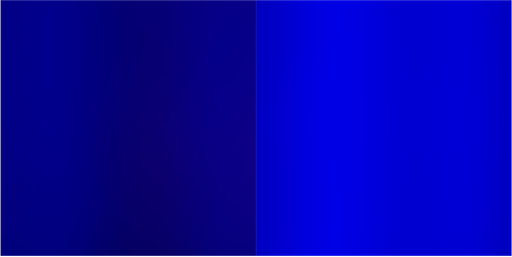}&
        \includegraphics[align=c,width=0.195\linewidth]{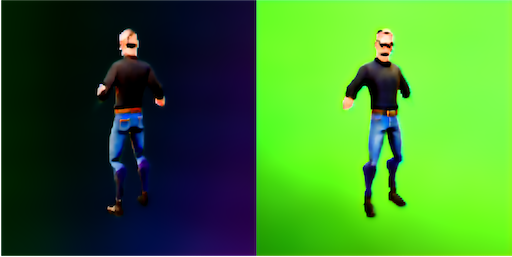} &
        \includegraphics[align=c,width=0.195\linewidth]{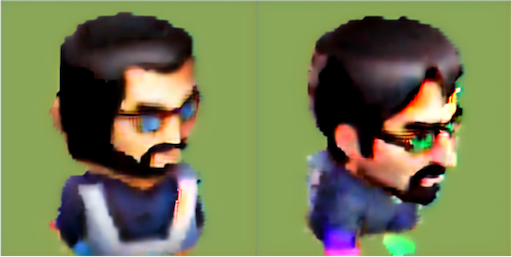}&
        \includegraphics[align=c,width=0.195\linewidth]{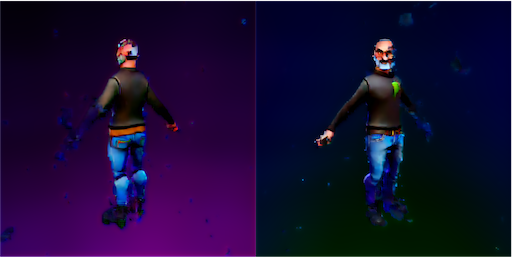}&
        \includegraphics[align=c,width=0.195\linewidth]{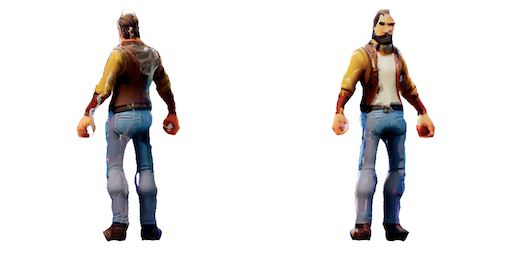}\\
        \multicolumn{5}{c}{\cprompt{Steve Jobs as the Fortnite character}} \\
        \includegraphics[align=c,width=0.195\linewidth]{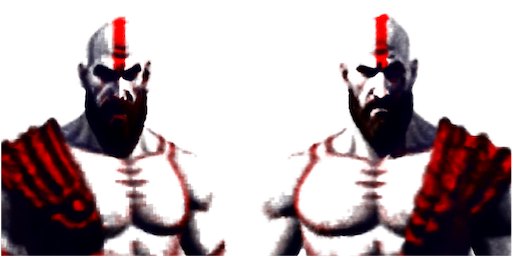}&
        \includegraphics[align=c,width=0.195\linewidth]{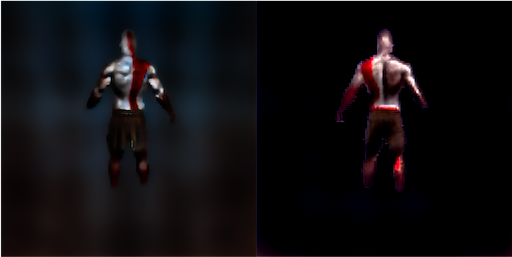} &
        \includegraphics[align=c,width=0.195\linewidth]{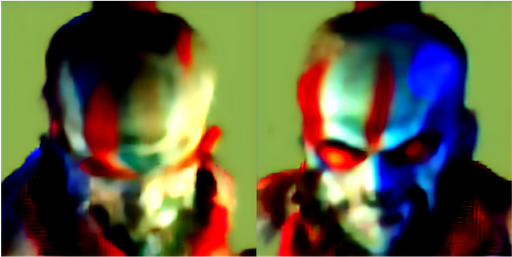}&
        \includegraphics[align=c,width=0.195\linewidth]{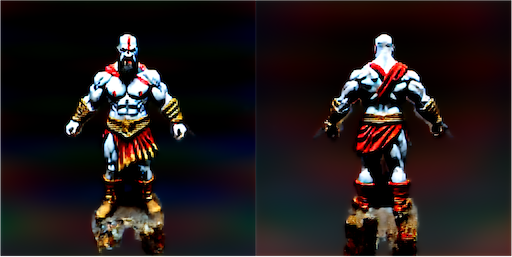}&
        \includegraphics[align=c,width=0.195\linewidth]{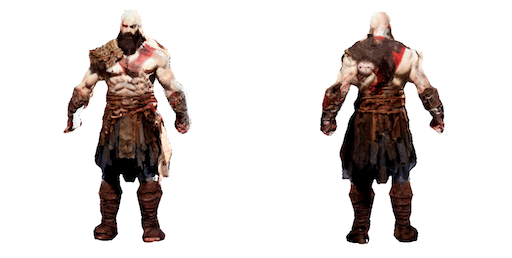}\\
        \multicolumn{5}{c}{\cprompt{Kratos in God of War}} \\
        \includegraphics[align=c,width=0.195\linewidth]{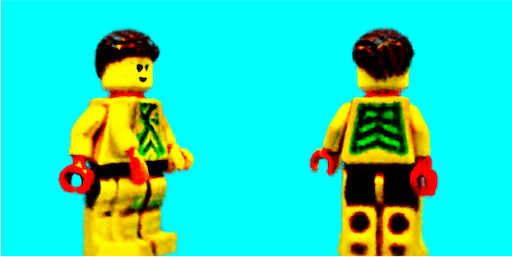}&
        \includegraphics[align=c,width=0.195\linewidth]{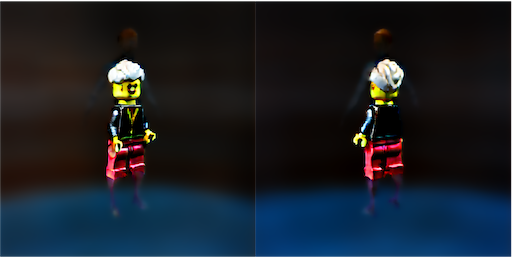} &
        \includegraphics[align=c,width=0.195\linewidth]{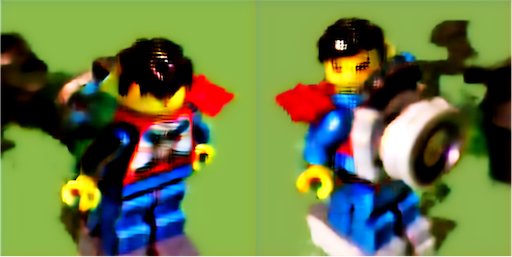}&
        \includegraphics[align=c,width=0.195\linewidth]{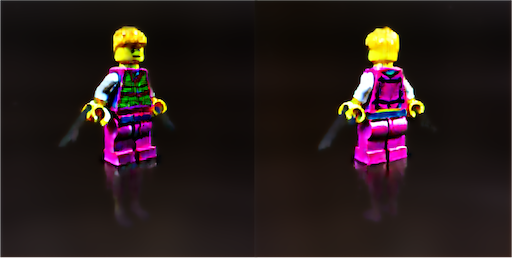}&
        \includegraphics[align=c,width=0.195\linewidth]{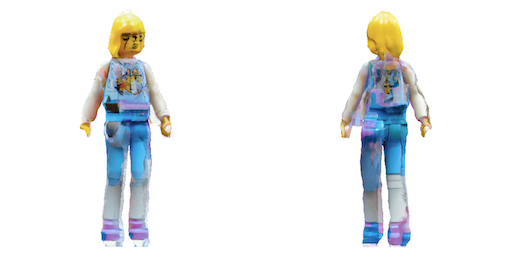}\\
        \multicolumn{5}{c}{\cprompt{Lego friend figurine}} \\
        \includegraphics[align=c,width=0.195\linewidth]{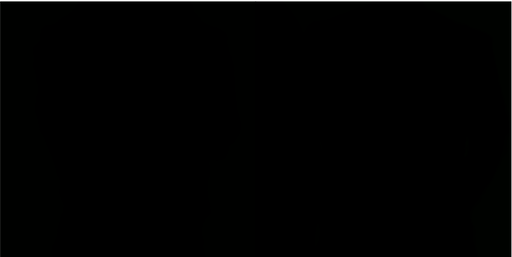}&
        \includegraphics[align=c,width=0.195\linewidth]{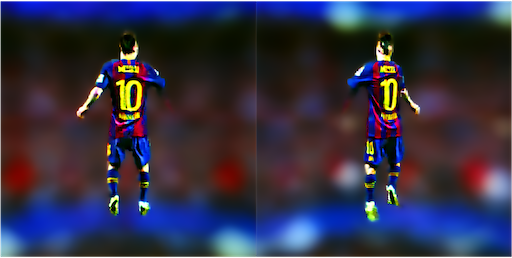} &
        \includegraphics[align=c,width=0.195\linewidth]{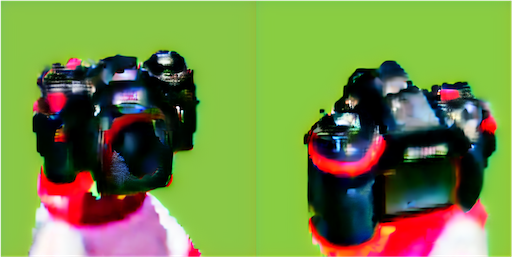}&
        \includegraphics[align=c,width=0.195\linewidth]{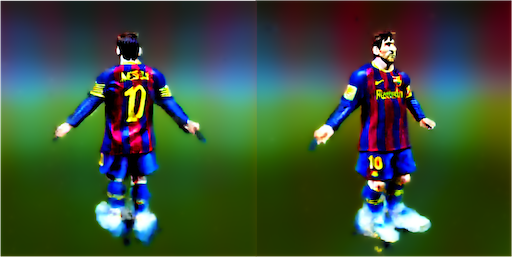}&
        \includegraphics[align=c,width=0.195\linewidth]{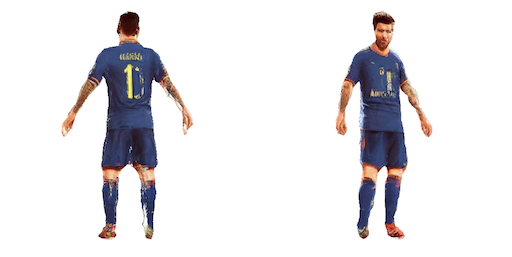}\\
        \multicolumn{5}{c}{\cprompt{Lionel Messi}} \\
        \includegraphics[align=c,width=0.195\linewidth]{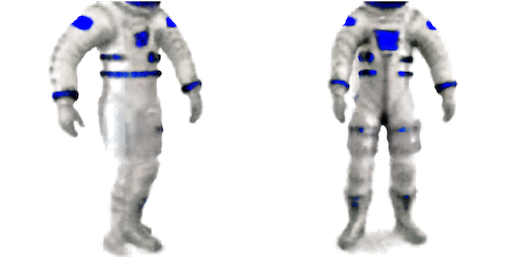}&
        \includegraphics[align=c,width=0.195\linewidth]{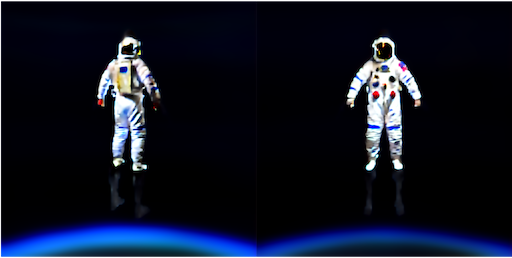} &
        \includegraphics[align=c,width=0.195\linewidth]{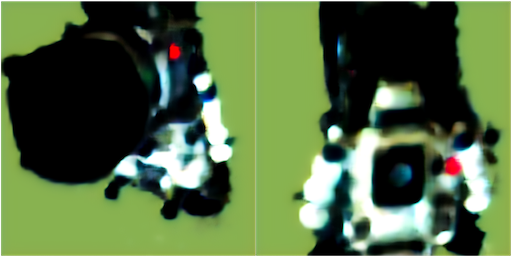}&
        \includegraphics[align=c,width=0.195\linewidth]{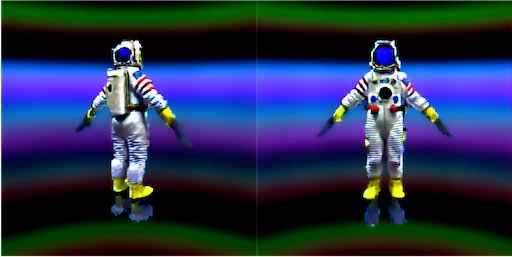}&
        \includegraphics[align=c,width=0.195\linewidth]{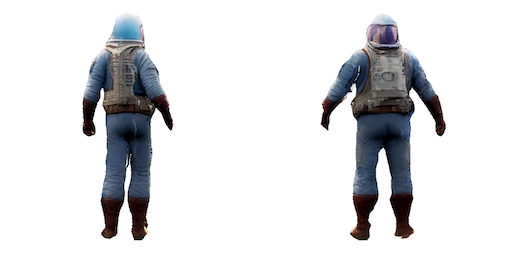}\\
        \multicolumn{5}{c}{\cprompt{Spacesuit}} \\
        \tabletitle{DreamFusion*~\cite{stable-dreamfusion}} & \tabletitle{Latent-NeRF~\cite{metzer2022latent-nerf}} & 
        \tabletitle{3DFuse~\cite{seo20233dfuse}} & \tabletitle{DreamAvatar~\cite{cao2023dreamavatar}} &      
        \tabletitle{Ours}    \\
    \end{tabular}
    \caption{Comparison with existing text-to-3D methods.}
    \label{fig:more_comparison}
\end{figure}

In this section, we provide more comparisons with stable-DreamFusion~\cite{stable-dreamfusion}, the open-source implementation of DreamFusion~\cite{poole2022dreamfusion}, Latent-NeRF~\cite{metzer2022latent-nerf}, 3DFuse~\cite{seo20233dfuse}, and DreamAvatar~\cite{cao2023dreamavatar} in Fig.~\ref{fig:more_comparison}. 

As provided in Fig.~\ref{fig:more_comparison}, accurately capturing the topology and structure of human models proves to be a challenging task for DreamFusion, Latent-NeRF, and 3DFuse. Additionally, geometry generated through SDS always exhibits noisy artifacts, while texture often suffers from high saturation. Furthermore, SDS-based methods lack visualized control over the generated results to meet our expectations. In contrast, Guide3D successfully transfers texture attributes from images to the 3D space, offering more convenient and personalized control of the geometry and texture for researchers, artists, and engineers.

On the other hand, DreamAvatar~\cite{cao2023dreamavatar} and AvatarCLIP~\cite{hong2022avatarclip} are commendable methods for 3D human generation. However, it is important to highlight certain considerations:

(1). DreamAvatar still encounters challenges when it comes to extreme cases such as generating loose clothing and avatars not harmoniously conformed to the SMPL model. For instance, in Fig.~\ref{fig:comparison_dreamavatar}, the legs of "Doctor Strange" generated by DreamAvatar appear to be in disarray. Additionally, as shown in Fig.~\ref{fig:more_comparison}, the "Lego friend figurine" generated by DreamAvatar includes opaque SMPL's legs and arms, which deviates from the overall texture and geometry of the generated Lego man and is thus deemed unreasonable.

(2). AvatarCLIP can also produce good details, yet it may face difficulties specifically in generating loose clothing. Almost all the generated avatars produced by AvatarCLIP adhere to the shape of SMPL, lacking the authentic topology of real-world clothing.

Last but not least, it is crucial to emphasize that current methods can only deal with text prompts. However, in real-world applications, the image is the most common and useful intermediate, making existing methods unsuitable for modern graphic pipelines.
\clearpage
\clearpage
\section{Failure cases}
\begin{figure}[htbp]
    \centering 
    \setlength{\tabcolsep}{0.2pt}
    \includegraphics[align=c,width=0.98\linewidth]{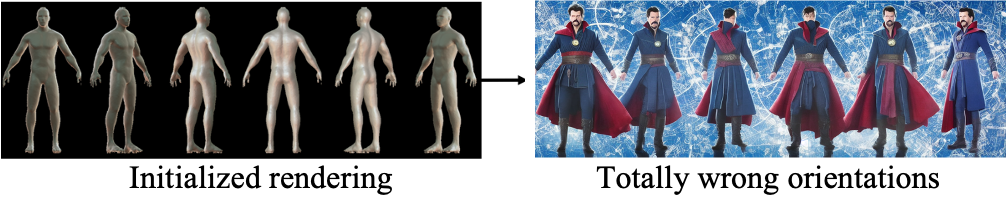}
    \caption{Failure example on multi-view image generations, with wrong orientations, poses and textures}
    \label{fig:failure}
\end{figure}

Our method can robustly create 3D models from multi-view images. However, we have observed some failures related to the generation of multi-view images (see Fig.~\ref{fig:failure} for examples).

\section{Multi-view image generation}
\begin{figure}[htbp]
    \centering 
    \setlength{\tabcolsep}{0.2pt}
    \includegraphics[align=c,width=0.98\linewidth]{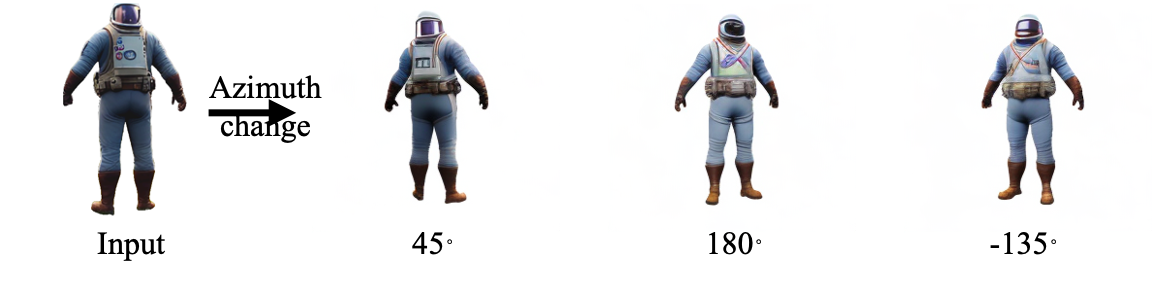}
    \caption{Examples of Zero-1-to-3. It struggles to generate correct orientation, body structure, texture and height.}
    \label{fig:zero123}
\end{figure}

It's intuitive that training and employing a human-oriented diffusion model will help minimize the inconsistency in generated multi-view images. However, datasets like RenderPeople~\footnote{\url{https://renderpeople.com}} or THUman2.0~\cite{tao2021function4d} primarily consist of normal individuals wearing ordinary clothing, rather than well-known characters that are encoded in the diffusion models. Therefore, they can not be well-recognized by the text description to the diffusion model. Hence, it is not feasible to directly employ such datasets to train a human-oriented diffusion model.

we further experimented on human-oriented examples using the officially released Zero-1-to-3~\cite{liu2023zero}, in Fig.~\ref{fig:zero123}, revealing that Zero-1-to-3 falls short in addressing human subjects.

\clearpage

\section{Results on synthetic dataset}
\begin{figure}[htbp]
    \vspace{-0.5em}
    \centering 
    \setlength{\tabcolsep}{0.245pt}
    \begin{tabular}{c} 
        
        \includegraphics[align=c,width=0.95\linewidth]{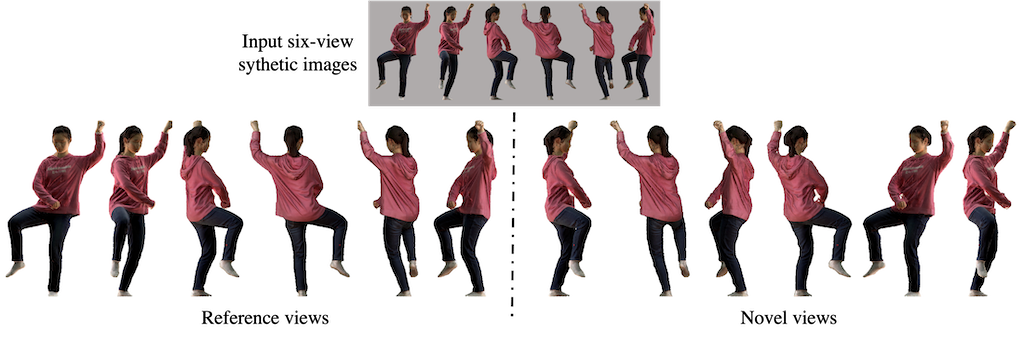}\\
        \vspace{-1.0em}
        \includegraphics[align=c,width=0.95\linewidth]{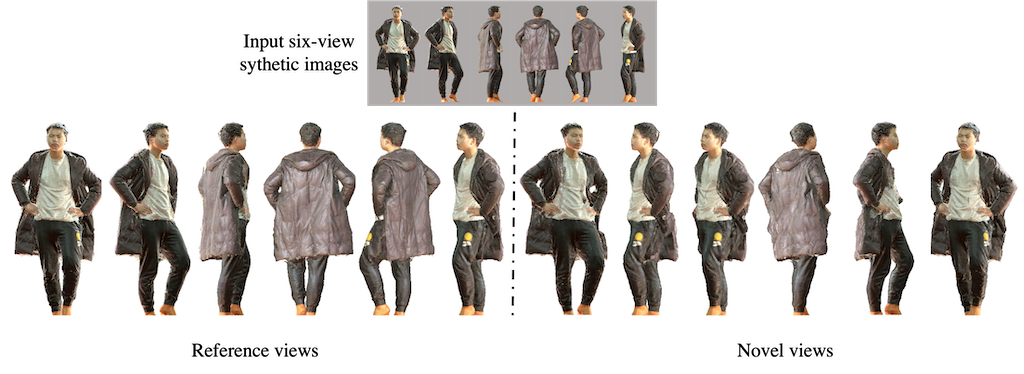}\\
        \vspace{-1.0em}
        \includegraphics[align=c,width=0.95\linewidth]{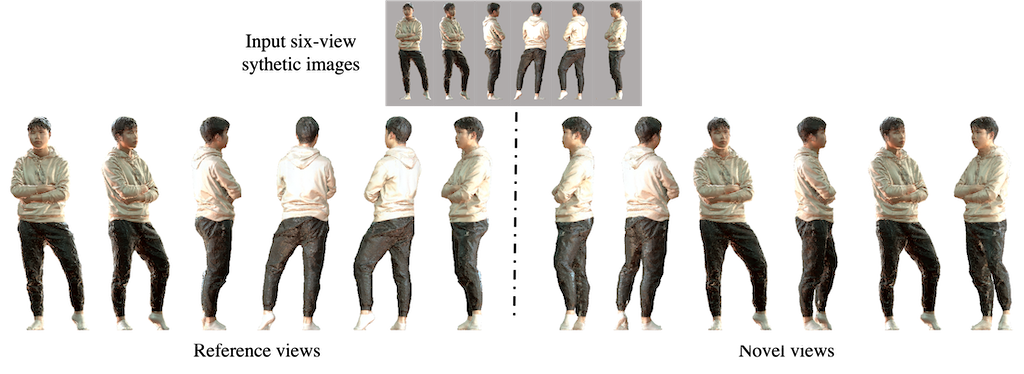}\\
        \vspace{-1.0em}
        \includegraphics[align=c,width=0.95\linewidth]{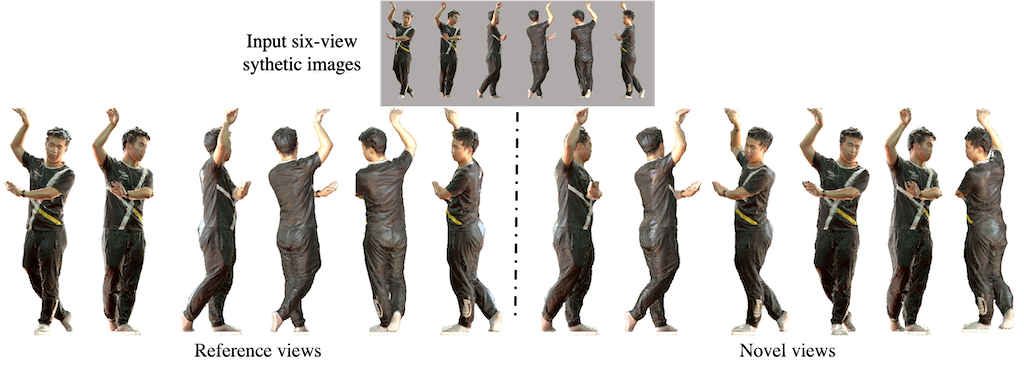}\\
        \vspace{-1em}
    \end{tabular}
    \caption{Generation results of Guide3D on synthetic data. 
    }
    \label{fig:a-synthetic}
\end{figure}

\begin{figure}[htbp]
    \centering 
    \setlength{\tabcolsep}{0.33pt}
    \begin{tabular}{c|c|c} 
        \includegraphics[align=c,width=\showwidth]{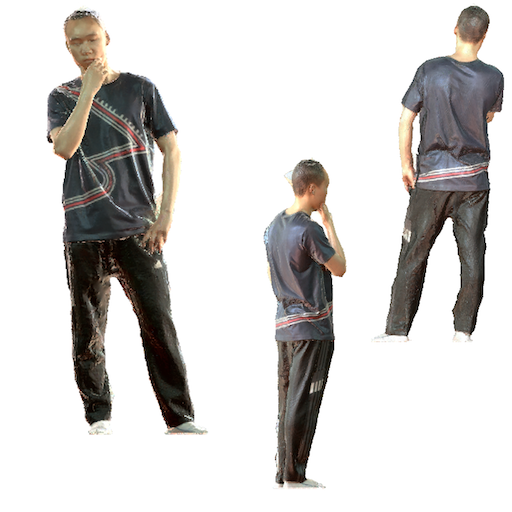}&
        \includegraphics[align=c,width=\showwidth]{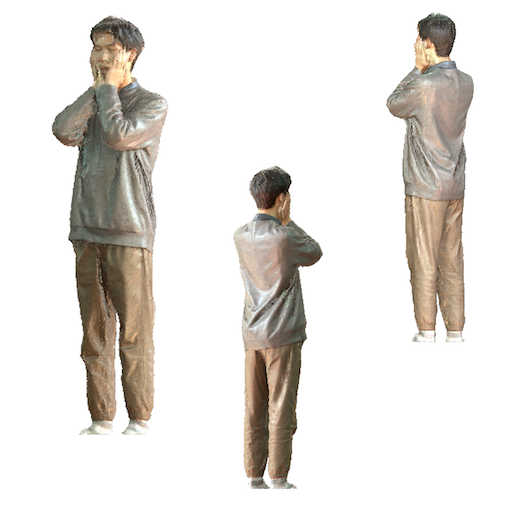}&
        \includegraphics[align=c,width=\showwidth]{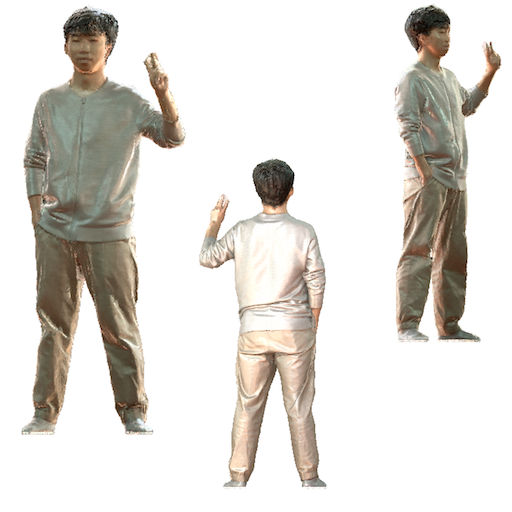} \\
        \cline{1-3}
        \includegraphics[align=c,width=\showwidth]{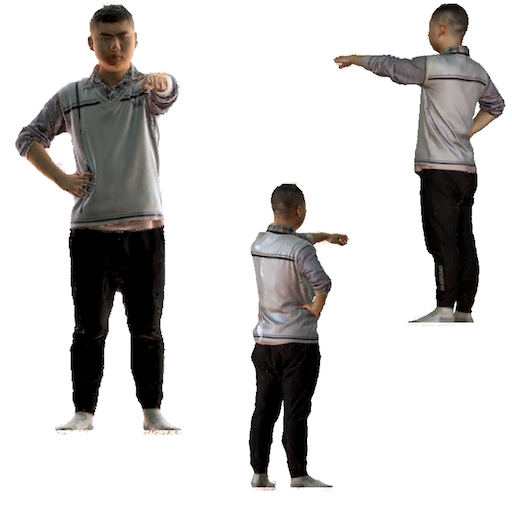}&
        \includegraphics[align=c,width=\showwidth]{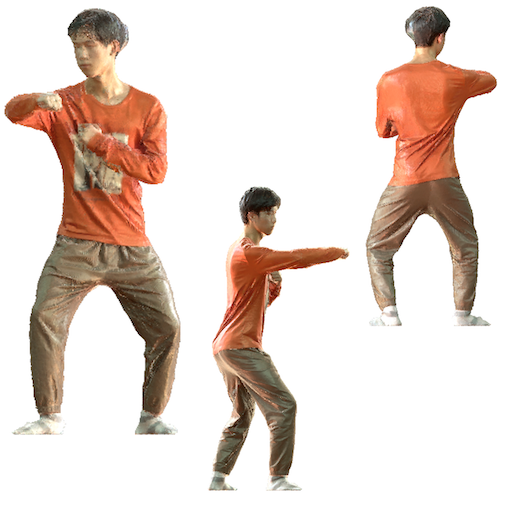}&
        \includegraphics[align=c,width=\showwidth]{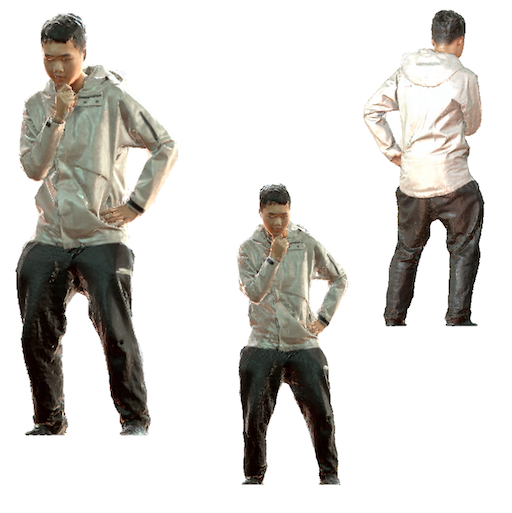} \\
        
        \cline{1-3}
        \includegraphics[align=c,width=\showwidth]{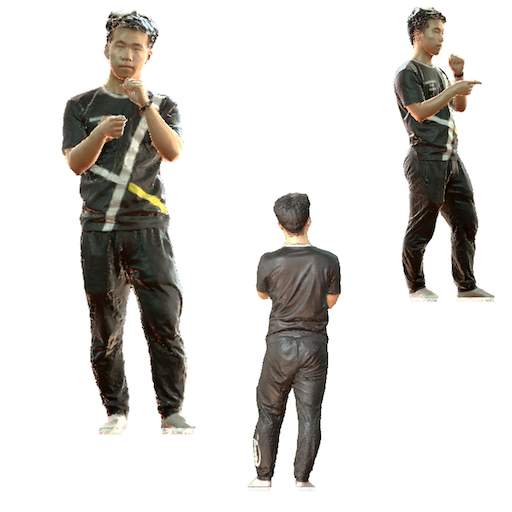}&
        \includegraphics[align=c,width=\showwidth]{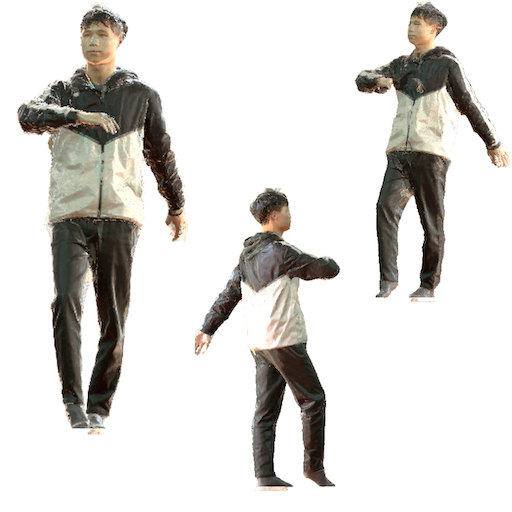}&
        \includegraphics[align=c,width=\showwidth]{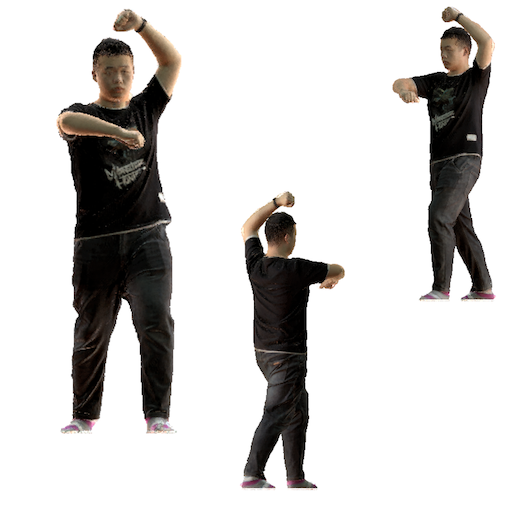} \\
        \cline{1-3}
        
        \includegraphics[align=c,width=\showwidth]{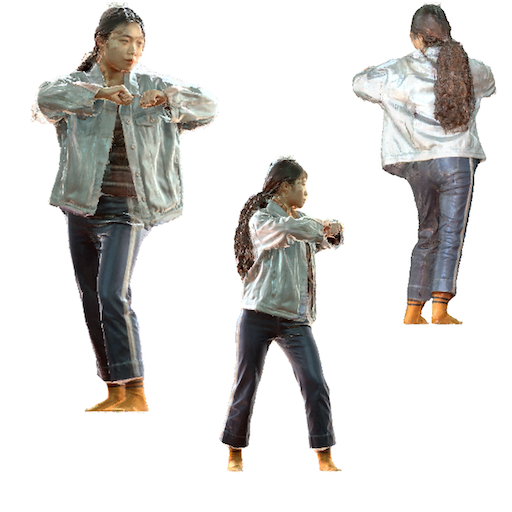}&
        \includegraphics[align=c,width=\showwidth]{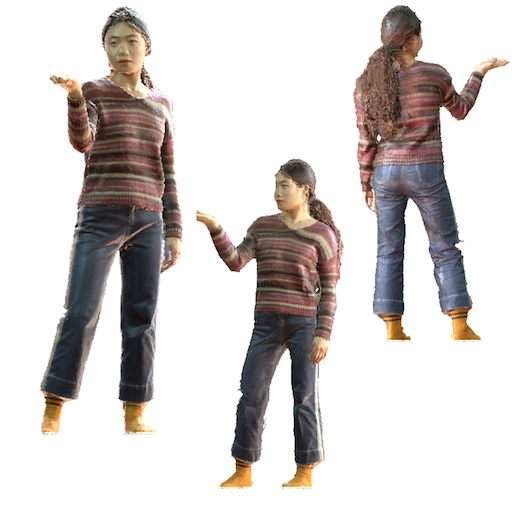}&
        \includegraphics[align=c,width=\showwidth]{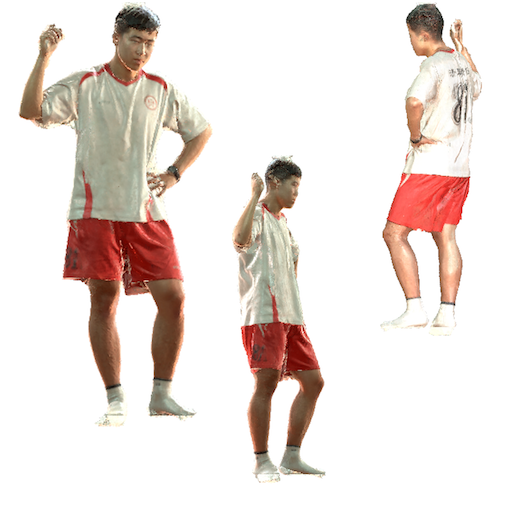} \\
    \end{tabular}
    \caption{Generation results of Guide3D on synthetic data. Our proposed approach can handle complex poses (including self-occlusion), and transfer detailed topologies, \eg ponytails, from 2D images to 3D models.
    }
    \label{fig:b-synthetic}
\end{figure}

Given the current limitations of text-to-image generation techniques~\cite{zhang2023controlnet}, which struggle with producing multi-view images of characters in complex poses, we further conduct analyses of our approach on synthetic THUman2.0 datasets~\cite{tao2021function4d}. In Fig.~\ref{fig:a-synthetic}, we display the outcomes of our approach, including both reference and novel views. As evident from the results, Guide3D capably manages complex poses (including self-occlusion), transfers detailed topologies, such as ponytails, from 2D images to 3D models, and skillfully accommodates loose clothing. Additional examples can be found in Fig.~\ref{fig:b-synthetic}. Note that in these cases, we omit the text prompt and diffusion denoising process since the diffusion model is specifically designed to address the multi-view inconsistency problem.

\clearpage

\end{document}